\documentclass[preprint,12pt,sort&compress]{elsarticle}

\usepackage{graphicx}
\usepackage{subcaption}
\usepackage{algorithm,algpseudocode}
\usepackage{enumitem}
\usepackage{hyperref}       
\usepackage{xcolor}         

\usepackage{amssymb}
\usepackage{amsmath}
\usepackage{amsthm}

\journal{Nuclear Physics B}

\begin{document}

\begin{frontmatter}



\title{Systematic Performance Assessment of Deep Material Networks for Multiscale Material Modeling}


\author[label1]{Xiaolong He\corref{cor1}}
\cortext[cor1]{Corresponding Author}
\ead{xiaolong.he@synopsys.com}
\author[label1]{Haoyan Wei}
\author[label1]{Wei Hu}
\author[label2]{Henan Mao}
\author[label1]{C. T. Wu}

\affiliation[label1]{organization={Computational and Multiscale Mechanics Group, Synopsys Inc.}, 
            city={Livermore},
            postcode={94551}, 
            state={CA},
            country={United States of America}}

\affiliation[label2]{organization={Synopsys Inc.}, 
			city={Redmond},
			postcode={98052}, 
			state={WA},
			country={United States of America}}

\begin{abstract}
Deep Material Networks (DMNs) are structure-preserving,  mechanistic machine learning models that embed micromechanical principles into their architectures, enabling strong extrapolation capabilities and significant potential to accelerate multiscale modeling of complex microstructures.
A key advantage of these models is that they can be trained exclusively on linear elastic data and then generalized to nonlinear inelastic regimes during online prediction.
Despite their growing adoption, systematic evaluations of their performance across the full offline-online pipeline remain limited.
This work presents a comprehensive comparative assessment of DMNs with respect to prediction accuracy, computational efficiency, and training robustness.
We investigate the effects of offline training choices, including initialization, batch size, training data size, and activation regularization on online generalization performance and uncertainty.
The results demonstrate that both prediction error and variance decrease with increasing training data size, while initialization and batch size can significantly influence model performance.
Moreover, activation regularization is shown to play a critical role in controlling network complexity and therefore generalization performance.
Compared with the original DMN, the rotation-free Interaction-based Material Network (IMN) formulation achieves a $3.4 \times$ - $4.7 \times$ speed-up in offline training, while maintaining comparable online prediction accuracy and computational efficiency.
These findings clarify key trade-offs between model expressivity and efficiency in structure-preserving material networks and provide practical guidance for their deployment in multiscale material modeling.
\end{abstract}




\end{frontmatter}


\section{Introduction}\label{sec:introduction}
Modern material systems with carefully engineered microstructures offer new opportunities to tailor materials with enhanced mechanical properties and functionalities across a wide range of engineering applications. 
Across diverse material classes, including composites, polycrystals, architected materials, and multifunctional systems, multiscale modeling has become a central paradigm for linking microscale mechanisms to macroscopic responses, enabling predictive simulation and design across length and time scales \cite{fish2010multiscale,geers2010multi,fish2021mesoscopic,bishara2023state}.
Efficient and accurate computational modeling of these multiscale material systems is essential not only for improving the predictive capabilities of large-scale structural analysis, but also for accelerating the discovery, optimization, and design of advanced materials.

Owing to the multiscale nature of heterogeneous materials, macroscopic properties are strongly influenced by microscale morphology, constituent properties, and phase interactions.
Although direct numerical simulations (DNS) of large-scale heterogeneous systems can provide high-fidelity predictions, they are often computationally prohibitive.
As a result, multiscale modeling methods have been developed to bridge microscale behavior and macroscale responses. 
Homogenization based on the representative volume element (RVE) concept \cite{hill1963elastic} has become a cornerstone of multiscale material modeling.
RVEs can be solved numerically using high-fidelity DNS methods, such as the Finite Element Method (FEM) \cite{feyel2000fe2,geers2010multi} and Fast Fourier Transform methods \cite{moulinec1998numerical,de2017finite,naili2020short}. 
Numerical RVE models can be coupled with structure-level FE models to model heterogeneous structures \cite{feyel2000fe2,feyel2003multilevel,kouznetsova2004multi,geers2010multi,spahn2014multiscale,kochmann2018efficient,tan2020direct,he2022multiscale}.
However, the high computational cost of RVE-based DNS remains a major bottleneck, limiting its applications to large-scale industrial problems \cite{xu2020data}.

Recent advances in machine learning (ML) and data science have created new opportunities for
developing data-driven material modeling and multiscale simulation methods \cite{ghaboussi1991knowledge,lecun2015deep,goodfellow2016deep,wang2018multiscale,ghavamian2019accelerating,liu2021review,bishara2023state,fuhg2024review}. 
Model-free data-driven computational mechanics approaches aim to bypass traditional constitutive modeling by solving physics-constrained optimization problems directly over material databases characterizing constitutive behaviors \cite{kirchdoerfer2016data,ibanez2018manifold,eggersmann2019model,he2020physics,he2020physicsAniso,he2021deep,he2021manifold,eggersmann2021model,gorgogianni2023adaptive,xu2024quantum}.
However, these approaches face challenges when modeling the nonlinear inelastic material behaviors.
In parallel, ML-based surrogate models have been proposed to approximate nonlinear constitutive laws directly \cite{vlassis2020geometric,masi2021thermodynamics,liu2021review,he2022thermodynamically,fuhg2024review,maia2023physically,he2024incremental,jones2022neural,guo2025history}.
However, such data-driven surrogates typically require extensive training data and often exhibit limited extrapolation capabilities to unseen material systems and loading paths due to the lack of explicit microscale physics.

To overcome these challenges and accelerate multiscale material modeling, several reduced-order modeling techniques have been developed as surrogates for high-fidelity simulations, including proper-orthogonal decomposition \cite{yvonnet2007reduced,oliver2017reduced,fritzen2018two,kaneko2021hyper}, self-consistent clustering analysis \cite{liu2016self,liu2018microstructural}, and 
a mechanistic, structure-preserving ML model known as the Deep Material Network (DMN) \cite{liu2019deep,liu2019exploring}.
\textit{Structure-preserving} ML refers to models that explicitly embed known mathematical, physical, or geometric structures, such as conservation laws, symmetries, constraints, invariants, and stability properties, into the model architecture, loss function, or training procedure, rather than relying purely on statistical fitting of data. By introducing such inductive biases, structure-preserving models achieve improved physical consistency, generalization, and stability, while reducing data requirements by ensuring that learned surrogates respect the underlying structure of the target system \cite{hernandez2021structure,celledoni2021structure,zhang2022gfinns,gruber2023energetically,gruber2024efficiently,park2024tlasdi,he2025thermodynamically}.
DMN can be regarded as a structure-preserving ML model in this sense, as its architecture and components explicitly encode mechanistic structure derived from micromechanics and classical homogenization theory.
Specifically, DMNs are constructed from a hierarchy of mechanistic building blocks with analytical homogenization solutions and physically interpretable parameters. 
As a result, DMNs trained solely on linear elastic homogenization data can effectively extrapolate to nonlinear inelastic composite responses, while preserving thermodynamic consistency and inheriting stress-strain monotonicity from their microscale constituents \cite{liu2019deep,liu2019exploring, gajek2020micromechanics}.

Inspired by the original work by Liu et al. \cite{liu2019deep,liu2019exploring}, compact variants of DMN have been proposed, including the rotation-free DMN \cite{gajek2020micromechanics} and the Interaction-based Material Network (IMN) \cite{noels2022interaction}. 
These models employ more efficient parameterizations of material orientations, thereby reducing the number of independent parameters while retaining the structure-preserving micromechanics-based homogenization philosophy.
In this study, we collectively refer to these compact formulations as IMN.
DMN and IMN have been successfully applied to a wide range of materials and structures, including multi-phase composites \cite{gajek2020micromechanics,gajek2021fe,noels2022micromechanics}, woven structures \cite{wu2021micro}, short-fiber-reinforced composites \cite{gajek2022fe,dey2022training,wei2023ls}, polycrystalline materials \cite{wei2025orientation}, micropolar composite constitutive models \cite{francis2025micropolar}, and fiber suspensions in incompressible non-Newtonian fluids \cite{sterr2025deep}.
In parallel, these material networks have been extended to capture cohesive interfacial failure \cite{liu2020deep}, strain localization \cite{liu2021cell}, thermal responses of woven composites \cite{shin2024deep}, coupled thermo-mechanical behavior \cite{shin2024deepb}, and internal microstructural stress and strain partitioning \cite{shin2026u}.
To efficiently model composites with varying microstructures, e.g., different fiber orientations or volume fractions, without retraining, DMN has been enhanced with transfer-learning strategies based on physically informed descriptors \cite{liu2019transfer,liu2020intelligent,huang2022microstructure,wei2023ls}, integrated with graph-based \cite{jean2025graph} and transformer-based feature representations \cite{wei2025foundation}, as well as micro-mechanics-informed parametric network architectures \cite{li2024micromechanics}.
In addition, strategies such as initialization \cite{shin2023deep} and sampling techniques \cite{noels2022interaction,noels2022micromechanics,dey2022training,srinivas2026rapid} have been explored to improve offline training efficiency and online prediction accuracy.
These structure-preserving material networks, as efficient surrogate models of material microstructures, have been coupled with FEM for multiscale structural simulations, achieving orders-of-magnitude speedups over DNS-based approaches \cite{liu2020intelligent,gajek2021fe,gajek2022fe,wei2023ls}.
DMN has been integrated into the multiphysics simulation software LS-DYNA for nonlinear multiscale modeling \cite{wei2023ls}.

Despite these advances, systematic, head-to-head evaluations of DMN and IMN remain limited across the full offline-online pipeline, including training robustness, computational cost, and online inference efficiency.
This paper addresses this gap by conducting a comprehensive performance assessment of DMN and IMN across three practical dimensions: (i) sensitivity to offline training choices, including initialization, batch size, training data size, and activation regularization; (ii) offline training efficiency; and (iii) online prediction accuracy and the computational cost of different iterative solvers at the macroscopic material point level.  

The remainder of this paper is organized as follows. 
Section \ref{sec:method} reviews the theoretical foundations of DMN and IMN, including their structure-preserving mechanistic building blocks, offline training procedures, and online prediction algorithms.
Section \ref{sec:experiment} investigates the effects of initialization, batch size, training data size, and activation regularization on training and generalization performance. 
Two online prediction schemes for IMN are evaluated, including the fixed-point iteration scheme and the Newton iteration, followed by a systematic comparison of DMN and IMN in terms of offline training cost and online prediction performance. 
Finally, concluding remarks and discussions are presented in Section \ref{sec:conclusion}.

\section{Methodology}\label{sec:method}
This section reviews structure-preserving mechanistic building blocks of DMN and IMN, and summarizes formulations and procedures for offline training and online prediction. 

\subsection{Deep Material Network (DMN)}\label{sec:dmn}
\begin{figure}[htp]
	\centering
	\includegraphics[width=0.6\linewidth]{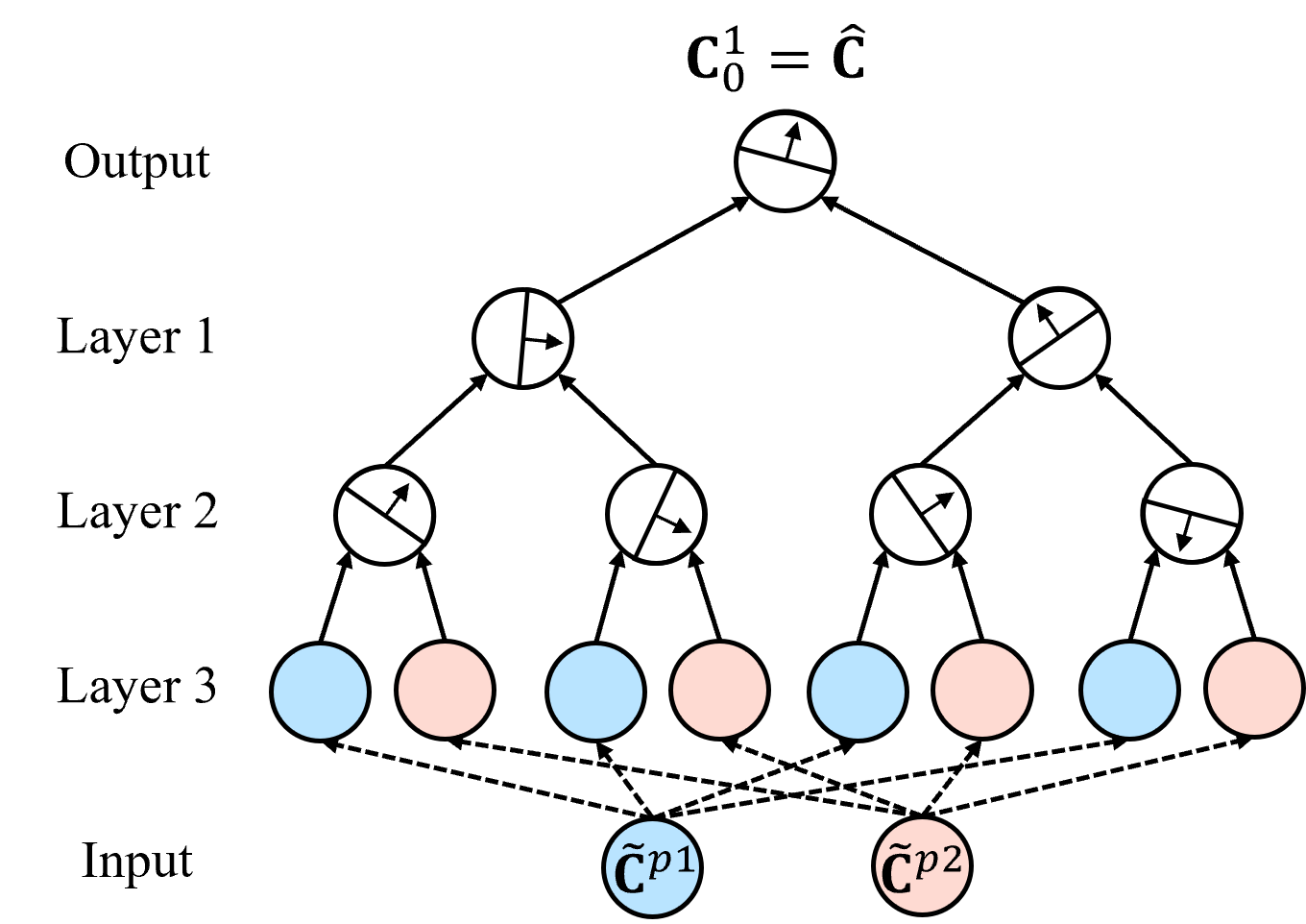}
	\caption{Architecture of a 3-layer deep material network.}\label{fig.architecture}
\end{figure}

The Deep Material Network (DMN) \cite{liu2019deep,liu2019exploring} is a mechanistic, structure-preserving ML model that embeds micromechanics principles for data-driven  multiscale modeling.
Fig. \ref{fig.architecture} illustrates a three-layer DMN constructed using a binary-tree network structure.
An $N$-layer binary-tree network contains a total of $2^{N+1} - 1$ nodes, among which $2^N$ nodes are located in the bottom layer $N$.
These bottom-layer nodes are called the \textit{base nodes}.

For the $n$-th node in layer $i$, four parameters are defined: a nodal weight $w_i^n$ and three independent Euler angles $\{\alpha_i^n, \beta_i^n, \gamma_i^n\}$.
Here, $0 \le i \le N$ denotes the layer index and $1 \le n \le 2^{i}$ denotes the node index within layer $i$, and $i=0$ corresponds to the output layer of the network.
The weights of base nodes are activated using the rectified linear unit (ReLU),
\begin{equation}\label{eq.dmn_activation}
    w_N^n = ReLU(z^n) = max(z^n,0), \quad 1 \le n \le 2^N,
\end{equation}
where $z^n$ denotes the activation of the $n$-th base node. The nodal weight $w_i^n$ of a parent node is computed as the sum of the weights of its two child nodes,
\begin{equation}\label{eq.dmn_nodal_weight}
    w_i^n = w_{i+1}^{2n-1} + w_{i+1}^{2n}, \quad 0 \le i \le N-1, \quad 1 \le n \le 2^i.
\end{equation}
Therefore, all nodal weights in DMN can be calculated from the activation of base nodes, $\{z^n\}_{n=1}^{2^N}$.
As a result, the independent parameters of DMN consists of $2^N$ activations and $3 \times (2^{N+1}-1)$ Euler angles, summing to $7 \times 2^N-3$ parameters.
Unlike purely data-driven surrogate models, these parameters have clear physical interpretations derived from mechanistic building blocks that preserve the physics of homogenization.

The averaged stress $\bar{\boldsymbol{\sigma}}_i^n$ of a parent node can be obtained by the rule of mixtures using the nodal weights and stresses of its two child nodes,
\begin{equation}\label{eq.dmn_avg_stress}
    \bar{\boldsymbol{\sigma}}_i^n 
    = f_{i+1}^{2n-1} \boldsymbol{\sigma}_{i+1}^{2n-1} 
    + f_{i+1}^{2n} \boldsymbol{\sigma}_{i+1}^{2n},
\end{equation}
where the volume fractions are defined as
\begin{equation}\label{eq.dmn_volume_fraction}
	f_{i+1}^{2n-1} = \frac{w_{i+1}^{2n-1}}{w_{i+1}^{2n-1}+w_{i+1}^{2n}}, \quad
	f_{i+1}^{2n} = \frac{w_{i+1}^{2n}}{w_{i+1}^{2n-1}+w_{i+1}^{2n}}.
\end{equation}

Accordingly, the averaged material stiffness $\bar{\mathbf{C}}_i^n$ of a parent node is expressed as a function of the nodal weights and material stiffness of its two child nodes,
\begin{equation}\label{eq.dmn_avg_stiffness}
    \bar{\mathbf{C}}_i^n 
    = \mathbf{C}_{i+1}^{2n} 
    - f_{i+1}^{2n-1} (\mathbf{C}_{i+1}^{2n}-\mathbf{C}_{i+1}^{2n-1}) \boldsymbol{\mathcal{A}},
\end{equation}
where $\boldsymbol{\mathcal{A}}$ is the strain concentration matrix within the DMN building block, derived by enforcing the interfacial equilibrium and kinematic compatibility of a two-phase composite material \cite{liu2019exploring}.
Detailed derivations and expressions for $\boldsymbol{\mathcal{A}}$ are provided in \cite{liu2019exploring}.

To capture directional material behaviors of complex microstructures, the Euler angles $\{\alpha_i^n, \beta_i^n, \gamma_i^n\}$ are defined at each node of the network.
Applying three-dimensional (3D) rotations to the averaged stiffness and stress yields rotated stiffness and stress,
\begin{equation}\label{eq.dmn_rotated_stiffness}
    \mathbf{C}_i^n 
    = \mathbf{R}^T(\alpha_i^n, \beta_i^n, \gamma_i^n) \bar{\mathbf{C}}_i^n  \mathbf{R}(\alpha_i^n, \beta_i^n, \gamma_i^n),
\end{equation}
\begin{equation}\label{eq.dmn_rotated_stress}
    \boldsymbol{\sigma}_i^n 
    = \mathbf{R}(\alpha_i^n, \beta_i^n, \gamma_i^n) \bar{\boldsymbol{\sigma}}_i^n,
\end{equation}
where $\mathbf{R}(\alpha_i^n, \beta_i^n, \gamma_i^n)$ is the rotation matrix based on the Euler angles.
Note that the base node stiffness is rotated prior to homogenization into the parent node stiffness.

\begin{figure}[htp]
\centering
    \begin{subfigure}{0.38\textwidth}
        \centering
        \includegraphics[width=1\linewidth]{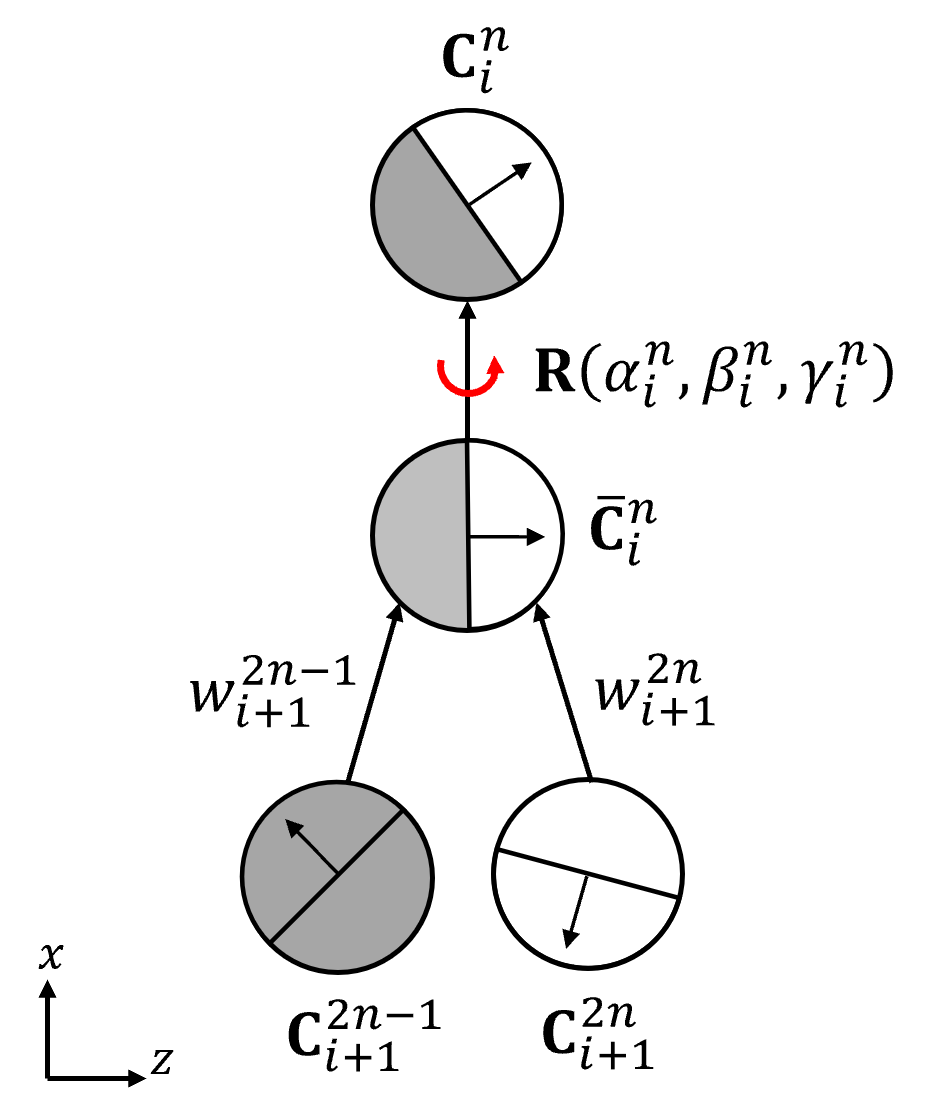}
        \caption{DMN}
    \end{subfigure}
    \begin{subfigure}{0.35\textwidth}
        \centering
        \includegraphics[width=1\linewidth]{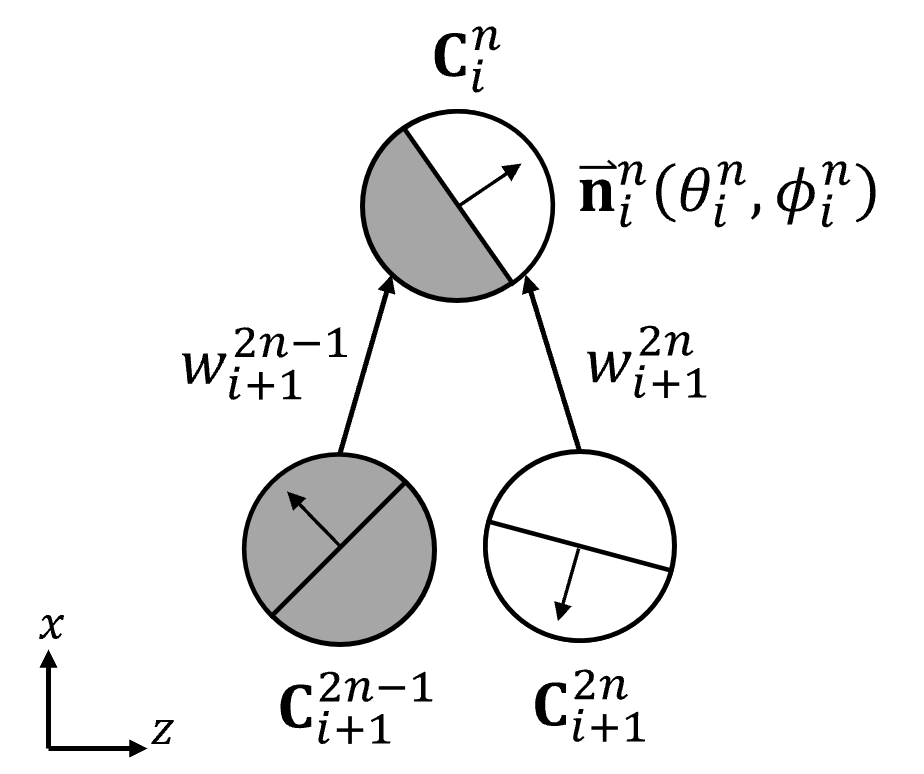}
        \caption{IMN}
    \end{subfigure}
\caption{Mechanistic building block of (a) DMN; (b) IMN.}\label{fig.building_block}
\end{figure}

In summary, each mechanistic DMN building block involves two fundamental operations: averaging and rotation, which preserve the essential micromechanical homogenization structure, as illustrated in Fig. \ref{fig.building_block}(a). 
By recursively applying these operations to physical quantities, such as stress, strain, and stiffness, throughout the network, DMN effectively homogenizes the microscopic material behaviors of base nodes to predict macroscopic composite responses at the top layer.

\subsubsection{Offline Training}\label{sec:dmn_offline}
Let $\boldsymbol{\psi} = \{\mathbf{z}, \boldsymbol{\alpha}, \boldsymbol{\beta}, \boldsymbol{\gamma} \}$ denote the set of independent trainable parameters of the DMN, where $\mathbf{z} = \{z^n\}_{n=1}^{2^N}$, $\boldsymbol{\alpha} = \{\alpha_i^n\}_{n=1}^{2^i}$, $\boldsymbol{\beta} = \{\beta_i^n\}_{n=1}^{2^i}$, $\boldsymbol{\gamma} = \{\gamma_i^n\}_{n=1}^{2^i}$, $0 \le i \le N$.
The homogenized material stiffness of a two-phase composite at the top node ($i=0$) can be expressed as
\begin{equation}\label{eq.dmn_top_stiffness}
    \hat{\mathbf{C}} 
    = \mathbf{C}_0^1 
    = \mathcal{H}(\tilde{\mathbf{C}}^{p1}, \tilde{\mathbf{C}}^{p2},\boldsymbol{\psi}),
\end{equation}
where $\tilde{\mathbf{C}}^{p1}$ and $\tilde{\mathbf{C}}^{p2}$ represent the stiffness matrices of the two base materials, which are assigned to DMN's base nodes, as illustrated in Fig. \ref{fig.architecture}.
Specifically, the base nodes with odd indices are assigned $\tilde{\mathbf{C}}^{p1}$, while the base nodes with even indices are assigned $\tilde{\mathbf{C}}^{p2}$.
The operator $\mathcal{H}$ represents the nonlinear, structure-preserving mapping induced by the sequence of averaging and rotation operations applied across all DMN layers from the base nodes to the top node.

The DMN parameters are determined by minimizing the following loss function,
\begin{equation}\label{eq.dmn_loss}
    \mathcal{L}(\boldsymbol{\psi}) 
    = \frac{1}{2N_s}\sum_{j=1}^{N_s}
    \frac{\big\lVert \mathbf{C}_{j} - \hat{\mathbf{C}}_{j} \big\rVert^2}{\big\lVert\mathbf{C}_j \big\rVert^2}
    + \eta \Big(\sum_{n=1}^{2^N} ReLU(z^n) - \xi \Big)^2,
\end{equation}
where $\lVert \cdot \rVert$ denotes the Frobenius norm and $\hat{\mathbf{C}}_{j}=\mathcal{H}(\tilde{\mathbf{C}}_j^{p1}, \tilde{\mathbf{C}}_j^{p2},\boldsymbol{\psi})$ is the prediction from the material network for the $j$-th sample.
The first term measures the relative error between the predicted and reference homogenized stiffness matrices, while the second term regularizes the activation of the base nodes. 
This regularization controls the effective network complexity and, consequently, influences both online prediction accuracy and computational efficiency.
The effects of the regularization parameters $\eta$ and $\xi$ on online performance are examined in Section \ref{sec:regularization_effect}.
The training dataset consists of $\{\tilde{\mathbf{C}}_j^{p1}, \tilde{\mathbf{C}}_j^{p2}, \mathbf{C}_j\}_j^{N_s}$, where $\tilde{\mathbf{C}}_j^{p1}$ and $\tilde{\mathbf{C}}_j^{p2}$ represent linear elastic stiffness matrices of the two base materials for the $j$-th sample, whereas $\mathbf{C}_j$ is the corresponding homogenized stiffness matrix of the RVE, obtained from DNS or experimental measurements. 
Here, $N_s$ denotes the total number of material samples.

\subsubsection{Online Prediction}\label{sec:dmn_online}
Although the DMN is trained on linear elastic data, it learns the essential microstructural interactions and can be applied for online prediction of nonlinear inelastic composite behavior under general loading conditions.
Given a prescribed macroscale strain increment, forward homogenization and backward de-homogenization of material information are performed in the DMN to predict multiscale material responses.
The incremental stress-strain relationship of the $n$-th node in layer $i$ is defined as
\begin{equation}\label{eq.dmn_online_stress_strain_relation}
    \Delta \bar{\boldsymbol{\sigma}}_i^n = \bar{\mathbf{C}}_i^n \Delta \bar{\boldsymbol{\varepsilon}}_i^n + \delta \bar{\boldsymbol{\sigma}}_i^n,
\end{equation}
where $\Delta \bar{\boldsymbol{\sigma}}_i^n$, $\Delta \bar{\boldsymbol{\varepsilon}}_i^n$, and $\bar{\mathbf{C}}_i^n$, denote the averaged incremental stress, incremental strain, and stiffness of the $n$-th node in layer $i$, respectively.
The variable $\delta \bar{\boldsymbol{\sigma}}_i^n$ denotes the averaged residual stress at this node, which arises only in the presence of material nonlinearities.
Similar to other material quantities, the forward homogenization of the residual stress can be derived analytically as
\begin{equation}\label{eq.dmn_online_residual_stress_homo}
    \delta \bar{\boldsymbol{\sigma}}_i^n
    = f_{i+1}^{2n-1} \delta \boldsymbol{\sigma}_{i+1}^{2n-1} 
    + f_{i+1}^{2n} \delta \boldsymbol{\sigma}_{i+1}^{2n}
    + \boldsymbol{\mathcal{X}},
\end{equation}
where $\boldsymbol{\mathcal{X}}$ is a correction term that depends on the stiffness matrices, nodal weights, and residual stresses of the two child nodes.
Detailed derivations and expressions of $\boldsymbol{\mathcal{X}}$ are provided in \cite{liu2019exploring}.
The rotated residual stress of child nodes, $\delta \boldsymbol{\sigma}_{i+1}^{2n-1}$ and $\delta \boldsymbol{\sigma}_{i+1}^{2n}$, are obtained by
\begin{equation}\label{eq.dmn_rotated_residual_stress}
	\delta\boldsymbol{\sigma}_i^n 
	= \mathbf{R}(\alpha_i^n, \beta_i^n, \gamma_i^n) \delta\bar{\boldsymbol{\sigma}}_i^n.
\end{equation}

\begin{figure}[htp]
\centering
\includegraphics[width=0.5\linewidth]{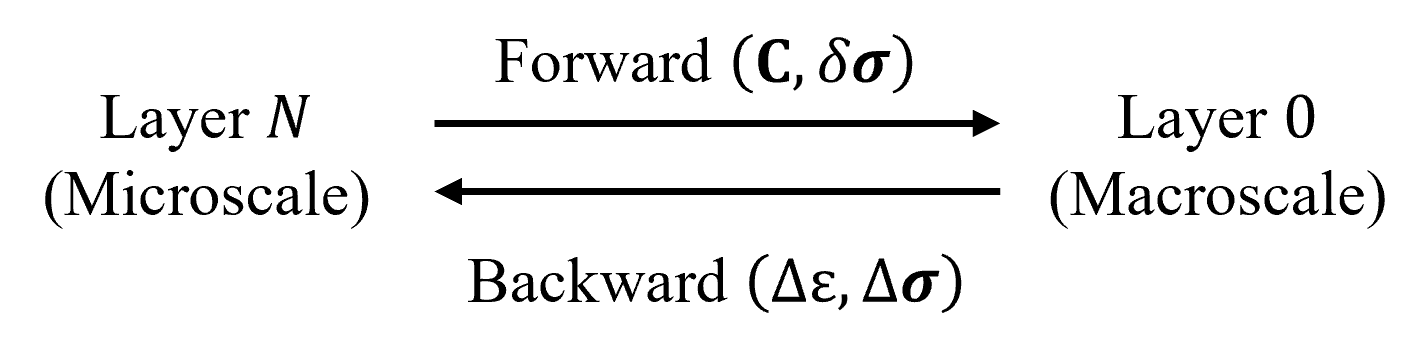}
\caption{Data flow in the online stage of DMN.}\label{fig.dmn_online_data_flow}
\end{figure}

The DMN online iterations start from the bottom layer with nodal incremental strains $\Delta \bar{\boldsymbol{\varepsilon}}_N^n$.
At each base node, the corresponding stiffness $\bar{\mathbf{C}}_N^n$ and residual stress $\delta \bar{\sigma}_N^n$ are obtained by evaluating constitutive laws of microscale base materials together with Eq. \eqref{eq.dmn_online_stress_strain_relation}.
Then, the stiffness matrices and residual stresses are propagated forward to the top layer by forward homogenization (Eqs. \eqref{eq.dmn_avg_stiffness} and \eqref{eq.dmn_online_residual_stress_homo}) and rotation (Eqs. \eqref{eq.dmn_rotated_stiffness} and \eqref{eq.dmn_rotated_residual_stress}), yielding $\mathbf{C}_0^1$ and $\delta\boldsymbol{\sigma}_0^1$ at the top node.
Given the macroscale loading condition at the top node, $\Delta \boldsymbol{\varepsilon}_0^1$ and $\Delta \boldsymbol{\sigma}_0^1$ are computed using Eq. \eqref{eq.dmn_online_stress_strain_relation}, which are subsequently propagated backward to the bottom layer through inverse rotation and backward de-homogenization based on the stress-strain relations within each building block, yielding updated nodal strain increments $\Delta \bar{\boldsymbol{\varepsilon}}_N^n$ at base nodes.
This process completes one DMN online iteration of forward homogenization and backward de-homogenization, as illustrated in Fig. \ref{fig.dmn_online_data_flow}.

The DMN online iterations continue until the relative difference between $\Delta \bar{\boldsymbol{\varepsilon}}_N^{n(k+1)}$ and $\Delta \bar{\boldsymbol{\varepsilon}}_N^{n(k)}$ is sufficiently small, with $k$ denoting the online iteration counter.
Upon convergence, the microscale stresses and internal state variables (e.g., equivalent plastic strain) at base nodes are updated, and the stress increment at the top node, $\Delta\boldsymbol{\sigma}_0^1$, is used to update the macroscale stress state.
It is noted that the total number of operations in one online iteration is proportional to the number of active base nodes.
Algorithm \ref{algorithm.dmn} summarizes the procedures for DMN online prediction.
The effects of residual stress on DMN online prediction performance are investigated and discussed in Section \ref{sec:dmn_residual}.
\begin{algorithm}
	\caption{DMN online prediction}\label{algorithm.dmn}
	\begin{algorithmic}[1]
		\Statex Input: macroscale strain increment $\Delta\boldsymbol{\varepsilon}$
		\Statex Output: macroscale stress increment $\Delta\boldsymbol{\sigma}$
		\State Initialize $\Delta \bar{\boldsymbol{\varepsilon}}_N^{n(1)} = \mathbf{0}$
		\For{$k \gets 1$ to $N_{iter}$}
			\State Compute stiffness $\bar{\mathbf{C}}_N^n$ and residual stress $\delta \bar{\boldsymbol{\sigma}}_N^n$ by evaluating constitutive laws of microscale base materials and Eq. \eqref{eq.dmn_online_stress_strain_relation}
			\State Forward homogenization of $\bar{\mathbf{C}}_N^n$ and $\delta \bar{\boldsymbol{\sigma}}_N^n$ to obtain $\mathbf{C}_0^1$ and $\delta\boldsymbol{\sigma}_0^1$
			\State Compute strain increment $\Delta \boldsymbol{\varepsilon}_0^1$ and stress increment $\Delta \boldsymbol{\sigma}_0^1$ by $\Delta \boldsymbol{\sigma}_0^1 = \mathbf{C}_0^1 \Delta \boldsymbol{\varepsilon}_0^1 + \delta\boldsymbol{\sigma}_0^1$
			\State Backward de-homogenization of strain and stress increments to obtain $\Delta \bar{\boldsymbol{\varepsilon}}_N^{n(k+1)}$
			\State Check if $\lVert \Delta \bar{\boldsymbol{\varepsilon}}_N^{n(k+1)} - \Delta \bar{\boldsymbol{\varepsilon}}_N^{n(k)} \rVert / \lVert \Delta \bar{\boldsymbol{\varepsilon}}_N^{n(k)} \rVert \le tol.$ If not, go to 3.
		\EndFor
		\State macroscale stress increment $\Delta\boldsymbol{\sigma} = \Delta \boldsymbol{\sigma}_0^1$
	\end{algorithmic}
\end{algorithm}

\subsection{Interaction-based Material Network (IMN)}\label{sec:imn}
The Interaction-based Material Network (IMN) \cite{gajek2020micromechanics,noels2022interaction} uses the same binary-tree network structure as the DMN.
The fundamental distinction between the two models lies in how their mechanistic building blocks are constructed.

In the original DMN formulation \cite{liu2019deep,liu2019exploring}, each building block is defined based on an interaction interface between two child nodes, where the unit normal to the interface, initially aligned with the material coordinate system's z-axis direction, is rotated to an arbitrary orientation by a 3D rotation operation parameterized by three Euler angles $\{\alpha, \beta, \gamma\}$, as illustrated in \ref{fig.building_block}(a). 
On the other hand, the IMN defines each building block with an interaction interface whose unit normal vector $\vec{\mathbf{n}}$ is parameterized by only two angles $\{\theta, \phi\}$, $\vec{\mathbf{n}}=[\cos(2\pi\theta) \sin(\pi\phi), \sin(2\pi\theta) \sin(\pi\phi), \cos(\pi\phi)]$, as illustrated in Fig. \ref{fig.building_block}(b).
As a result, while both DMN and IMN can orient homogenized material nodes arbitrarily to capture directional material behavior, IMN requires fewer trainable parameters due to its more compact orientation parameterization.
Specifically, in an $N$-layer DMN, each node is associated with three independent Euler angles, resulting in $3 \times (2^{N+1}-1)$ trainable orientation parameters, while an $N$-layer IMN assigns only two angles $\{\theta, \phi\}$ to each parent node, yielding $2 \times (2^{N}-1)$ trainable orientation parameters.
Together with the $2^N$ activation parameters in the bottom layer, an $N$-layer IMN has a total of $3 \times 2^{N}-2$ independent trainable parameters, compared with $7 \times 2^{N}-3$ for an $N$-layer DMN. 
For the same network depth, the number of trainable parameters in DMN is therefore approximately 2.3 times that of IMN.
Although IMN is more parameter-efficient, the effect of this reduction in model complexity on learning capability, offline training efficiency, and online prediction accuracy remains unclear. 
In this study, we conduct a quantitative comparison between DMN and IMN to systematically assess the trade-offs.

In the following, we briefly review the mechanistic building block of IMN. 
For clarity, the superscript "$h$" denotes quantities associated with a parent node, while superscripts "1" and "2" denote quantities associated with its two child nodes.
Based on the rule of mixtures, the strain and stress relations between a parent node and its two child nodes are given by
\begin{equation}\label{eq.imn_avg_strain}
    \boldsymbol{\varepsilon}^h
    = f^{1} \boldsymbol{\varepsilon}^{1}
    + f^{2} \boldsymbol{\varepsilon}^{2}.
\end{equation}
\begin{equation}\label{eq.imn_avg_stress}
    \boldsymbol{\sigma}^h 
    = f^{1} \boldsymbol{\sigma}^{1} 
    + f^{2} \boldsymbol{\sigma}^{2},
\end{equation}
with $f^1+f^2=1$.
The Hill-Mandel condition enforcing energy conservation between the parent and child nodes reads
\begin{equation}\label{eq.imn_energy_conservation}
    (\boldsymbol{\varepsilon}^h)^T \boldsymbol{\sigma}^h
    = (\boldsymbol{\varepsilon}^{1})^T \boldsymbol{\sigma}^{1} f^{1}
    + (\boldsymbol{\varepsilon}^{2})^T \boldsymbol{\sigma}^{2} f^{2}.
\end{equation}
Combining Eqs. \eqref{eq.imn_avg_strain}-\eqref{eq.imn_avg_stress} with Eq. \eqref{eq.imn_energy_conservation} gives
\begin{equation}\label{eq.imn_firstNsecond}
    (\boldsymbol{\varepsilon}^{1} - \boldsymbol{\varepsilon}^{2})^T(\boldsymbol{\sigma}^{1} - \boldsymbol{\sigma}^{2}) = 0.
\end{equation}
Traction continuity across the interaction interface is enforced as
\begin{equation}\label{eq.imn_traction_equilibrium}
    \mathbf{H}^T (\boldsymbol{\sigma}^{2} - \boldsymbol{\sigma}^{1}) = \mathbf{0},
\end{equation}
where $\mathbf{H}$ is an orientation matrix associated with the interface unit normal $\vec{\mathbf{n}}$, defined as
\begin{equation}\label{eq.imn_H}
    \mathbf{H} (\vec{\mathbf{n}}) = 
    \begin{bmatrix}
        \vec{n}_1 & 0 & 0 \\
        0 & \vec{n}_2 & 0 \\
        0 & 0 & \vec{n}_3 \\
        0 & \vec{n}_3 & \vec{n}_2 \\
        \vec{n}_3 & 0 & \vec{n}_1 \\
        \vec{n}_2 & \vec{n}_1 & 0
    \end{bmatrix}.
\end{equation}
Considering Eqs. \eqref{eq.imn_firstNsecond} and \eqref{eq.imn_traction_equilibrium}, one obtains
\begin{equation}\label{eq.imn_strain_normal_vector}
    (\boldsymbol{\varepsilon}^{1} - \boldsymbol{\varepsilon}^{2})^T
    = \frac{1}{f^{1} f^{2}} (\mathbf{H} \mathbf{b})^T, \quad \forall \mathbf{b}
\end{equation}
Combining Eqs. \eqref{eq.imn_avg_strain}-\eqref{eq.imn_avg_stress} with Eq. \eqref{eq.imn_strain_normal_vector}, the strains of child nodes can be expressed in terms of the homogenized strain as
\begin{equation}\label{eq.imn_micro_macrostrain}
    \boldsymbol{\varepsilon}^{1} = \boldsymbol{\varepsilon}^h + \frac{1}{f^{1}} \mathbf{H} \mathbf{b}, \quad
    \boldsymbol{\varepsilon}^{2} = \boldsymbol{\varepsilon}^h - \frac{1}{f^{2}} \mathbf{H} \mathbf{b}
\end{equation}
Eqs. \eqref{eq.imn_avg_strain}-\eqref{eq.imn_micro_macrostrain} constitute the mechanistic foundation of the IMN building block.
By considering linear constitutive relationships of the parent node and child nodes, $\boldsymbol{\sigma}^h = \mathbf{C}^h \boldsymbol{\varepsilon}^h$, $\boldsymbol{\sigma}^{1} = \mathbf{C}^{1} \boldsymbol{\varepsilon}^{1}$, $\boldsymbol{\sigma}^{2} = \mathbf{C}^{2} \boldsymbol{\varepsilon}^{2}$, together with the traction continuity condition in Eq. \eqref{eq.imn_traction_equilibrium} and strain relationship in Eq. \eqref{eq.imn_micro_macrostrain}, the $\mathbf{b}$ vector can be expressed as 
\begin{equation}\label{eq.imn_b_vector}
    \mathbf{b} 
    = \mathbf{B} \boldsymbol{\varepsilon}^h, \quad
    \text{with } \mathbf{B} = f^{1} f^{2} \big[ \mathbf{H}^T (f^{2} \mathbf{C}^{1} + f^{1} \mathbf{C}^{2})\mathbf{H} \big]^{-1} \mathbf{H}^T(\mathbf{C}^{2} - \mathbf{C}^{1}),
\end{equation}
and the homogenized stiffness $\mathbf{C}^h$ of the parent node is given by
\begin{equation}\label{eq.imn_stiffness_homo}
    \mathbf{C}^h 
    = f^{1} \mathbf{C}^{1} + f^{2} \mathbf{C}^{2} 
    - (\mathbf{C}^{2} - \mathbf{C}^{1}) \mathbf{H} \mathbf{B}.
\end{equation}

\subsubsection{Offline Training}\label{sec:imn_offline}
Let $\Tilde{\boldsymbol{\psi}} = \{\mathbf{z}, \boldsymbol{\theta}, \boldsymbol{\phi} \}$ denote the set of independent trainable parameters of the IMN, where $\mathbf{z} = \{z^n\}_{n=1}^{2^N}$, $\boldsymbol{\theta} = \{\theta_i^n\}_{n=1}^{2^i}$, $\boldsymbol{\phi} = \{\phi_i^n\}_{n=1}^{2^i}$, $0 \le i \le N-1$.
The homogenized material stiffness of a two-phase composite at the top node ($i=0$) can be expressed as
\begin{equation}\label{eq.dmn_top_stiffness}
    \hat{\mathbf{C} }
    = \mathbf{C}_0^{1} 
    = \Tilde{\mathcal{H}}(\tilde{\mathbf{C}}^{p1}, \tilde{\mathbf{C}}^{p2},\Tilde{\boldsymbol{\psi}}),
\end{equation}
where the operator $\Tilde{\mathcal{H}}$ represents the nonlinear structure-preserving mapping induced by the sequence of homogenization operations applied across all IMN layers from the base nodes to the top node.
Here, $\tilde{\mathbf{C}}^{p1}$ and $\tilde{\mathbf{C}}^{p2}$ denote the linear elastic stiffness matrices of the two base materials. 
The ReLU activation is employed to obtain the nodal weights in the bottom layer, as shown in Eq. \eqref{eq.dmn_activation}.
The IMN parameters are determined by minimizing the loss function defined in Eq. \eqref{eq.dmn_loss} with $\hat{\mathbf{C}}_j = \Tilde{\mathcal{H}}(\tilde{\mathbf{C}}^{p1}, \tilde{\mathbf{C}}^{p2}, \Tilde{\boldsymbol{\psi}})$.

\subsubsection{Online Prediction}\label{sec:imn_online}
\begin{figure}[htp]
\centering
\includegraphics[width=0.5\linewidth]{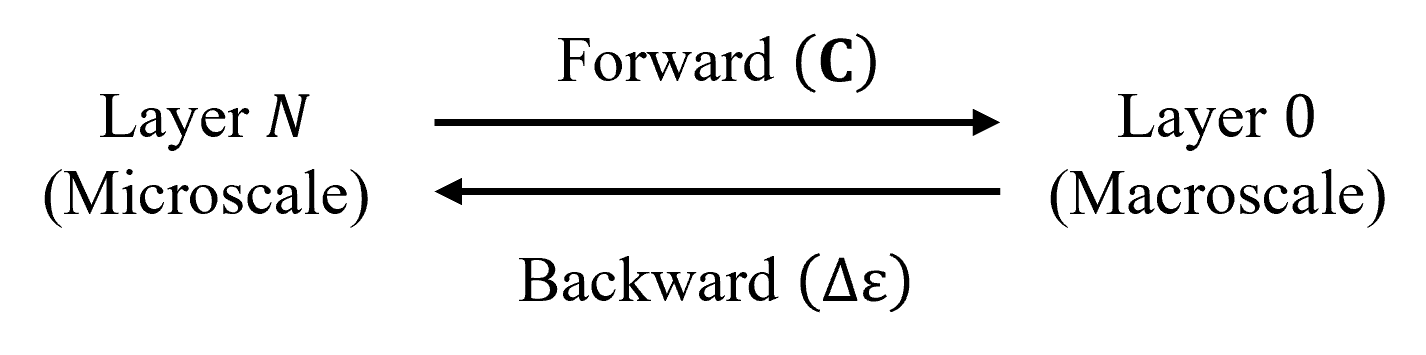}
\caption{Data flow in the fixed-point online stage of IMN.}\label{fig.imn_online_data_flow}
\end{figure}

\paragraph{Fixed-Point Iteration}
Similar to the DMN online iteration described in Section \ref{sec:dmn_online}, a straightforward IMN online iteration scheme, termed the fixed-point iteration, can be formulated. 
This scheme involves forward homogenization and backward de-homogenization, as illustrated in Fig. \ref{fig.imn_online_data_flow}.
The key distinction from the DMN online algorithm is that the IMN fixed-point iteration does not explicitly involve stress variables. 

Specifically, the IMN fixed-point online iterations begin at the bottom layer with nodal incremental strains $\Delta\boldsymbol{\varepsilon}_N^n$.
At each base node, the corresponding stiffness matrix $\mathbf{C}_N^n$ is obtained by evaluating the constitutive laws of the microscale base materials.
The stiffness matrices are then propagated forward to the top layer via forward homogenization (Eqs. \eqref{eq.imn_stiffness_homo}, yielding the homogenized stiffness $\mathbf{C}_0^1$ at the top node.
During this forward propagation, the matrix $\mathbf{B}$ defined in Eq. \eqref{eq.imn_b_vector} is simultaneously determined for all parent nodes.

Given the prescribed macroscale loading condition at the top node $\Delta \boldsymbol{\varepsilon}_0^1$, one can back propagate it to the bottom layer through de-homogenization using Eqs. \eqref{eq.imn_micro_macrostrain} and \eqref{eq.imn_b_vector}, yielding updated nodal strain increments $\Delta \boldsymbol{\varepsilon}_N^n$ in the bottom layer.
This procedure completes one IMN fixed-point online iteration.

The IMN fixed-point iterations are repeated until the relative difference between $\Delta \boldsymbol{\varepsilon}_N^{n(k+1)}$ and $\Delta \boldsymbol{\varepsilon}_N^{n(k)}$ is below a prescribed tolerance, where $k$ denotes the iteration counter.
Upon convergence, the microscale stresses at the base nodes are updated according to their constitutive laws, and the resulting stress increment at the top node, $\Delta \boldsymbol{\sigma}_0^1$, is used to update the macroscale stress state. 
Algorithm \ref{algorithm.imn_fixed_point} summarizes the complete procedure for IMN online prediction using fixed-point iterations.

\begin{algorithm}
	\caption{IMN online prediction with fixed-point iterations}\label{algorithm.imn_fixed_point}
	\begin{algorithmic}[1]
		\Statex Input: macroscale strain increment $\Delta\boldsymbol{\varepsilon}$
		\Statex Output: macroscale stress increment $\Delta\boldsymbol{\sigma}$
		\State Initialize $\Delta\boldsymbol{\varepsilon}_N^{n(1)} = \mathbf{0}$
		\For{$k \gets 1$ to $N_{iter}$}
		\State Compute stiffness $\mathbf{C}_N^n$ by evaluating constitutive laws of microscale base materials
		\State Forward homogenization of stiffness $\mathbf{C}_N^n$ to obtain $\mathbf{C}_0^1$ for the top node and $\mathbf{B}_i^n$ for all parent nodes
		\State Backward de-homogenization of macroscale strain increment $\Delta\boldsymbol{\varepsilon}$ to obtain $\Delta \boldsymbol{\varepsilon}_N^{n(k+1)}$
		\State Check if $\lVert \Delta \boldsymbol{\varepsilon}_N^{n(k+1)} - \Delta \boldsymbol{\varepsilon}_N^{n(k)} \rVert / \lVert \Delta \boldsymbol{\varepsilon}_N^{n(k)} \rVert \le tol.$ If not, go to 3.
		\EndFor
		\State Compute stress increments $\Delta\boldsymbol{\sigma}_N^n$ by evaluating constitutive laws of microscale base materials 
		\State Compute macroscale stress increment: $\Delta\boldsymbol{\sigma} = \sum_{n=1}^{2^N} w_N^n \Delta\boldsymbol{\sigma}_N^n$
	\end{algorithmic}
\end{algorithm}

\paragraph{Newton Iteration}
Based on the relationship between the strain of the parent node and the strains of its child nodes, as given in Eq. \eqref{eq.imn_micro_macrostrain}, the nodal strains in the bottom layer can be expressed in terms of the macroscale strain at the top node, $\boldsymbol{\varepsilon}_0^1$, as
\begin{equation}\label{eq.imn_bottom_top_strain}
    \boldsymbol{\varepsilon}_N^n 
    = \boldsymbol{\varepsilon}_0^1
    + \sum_{i=1}^N \frac{\mathbf{H}\Big(\vec{\mathbf{n}}_{i-1}^{1+r_{N-i+1}^n}\Big)}{w_i^{1+r_{N-i}^n}}
    (-1)^{r_{N-i}^n} \Big( w_{i-1}^{1+r_{N-i+1}^n} \mathbf{b}_{i-1}^{1+r_{N-i+1}^n} \Big),
    \quad n = 1, ..., 2^N,
\end{equation}
where $r_i^n = \mod(n-1, 2^i)$. 
Defining the vectors, $\boldsymbol{\varepsilon}^{\text{nodes}} = [\boldsymbol{\varepsilon}_N^1,\boldsymbol{\varepsilon}_N^2,..., \boldsymbol{\varepsilon}_N^{2^N}]^T$ and $\boldsymbol{\varepsilon}^{\text{macro}} = [\boldsymbol{\varepsilon}_0^1,\boldsymbol{\varepsilon}_0^1,...,\boldsymbol{\varepsilon}_0^1]^T$, the system of equations in Eq. \eqref{eq.imn_bottom_top_strain} can be written as
\begin{equation}\label{eq.imn_local_strain}
    \boldsymbol{\varepsilon}^{\text{local}}
    = 
    \boldsymbol{\varepsilon}^{\text{nodes}} - \boldsymbol{\varepsilon}^{\text{macro}}
    = 
    \begin{bmatrix}
        \mathbf{A}_{1,1} & \mathbf{A}_{1,2} & \hdots & \mathbf{A}_{1,2^N-1} \\
        \mathbf{A}_{2,1} & \mathbf{A}_{2,2} & \hdots & \mathbf{A}_{2,2^N-1} \\
        \vdots & \vdots & \vdots & \vdots \\
        \mathbf{A}_{2^N,1} & \mathbf{A}_{2^N,2} & \hdots & \mathbf{A}_{2^N,2^N-1} \\
    \end{bmatrix}
        \begin{bmatrix}
        \mathbf{a}_1 \\ \mathbf{a}_2 \\ \vdots \\ \mathbf{a}_{2^N-1}
    \end{bmatrix}
    = \mathbf{A} \mathbf{a},
\end{equation}
where $i=1,...,N$, $n=1,...,2^N$, and
\begin{equation}
    \begin{split}
        \mathbf{a}_{2^{i-1}+j-1} & = w_{i-1}^j \mathbf{b}_{i-1}^j, \quad \text{for } j = 1,..., 2^{i-1}, \\
        \mathbf{A}_{n, 2^{i-1}+j-1} & = \frac{\mathbf{H}\Big(\vec{\mathbf{n}}_{i-1}^{1+r_{N-i+1}^n}\Big)}{w_i^{1+r_{N-i}^n}} (-1)^{r_{N-i}^n}, \quad \text{for } j = 1 + r_{N-i+1}^n, \\
        \mathbf{A}_{n, 2^{i-1}+j-1} & = \mathbf{0}_{6\times3}, \quad \text{for } j \ne 1 + r_{N-i+1}^n.
    \end{split}
\end{equation}
The vector $\mathbf{a}$ is referred to as the jumping vector, and $\mathbf{A}$ denotes a weighted orientation matrix representing the IMN architecture.
For a prescribed macroscale strain $\boldsymbol{\varepsilon}_0^1$, the jumping vector $\mathbf{a}$ is determined by minimizing a residual vector that enforces interfacial traction equilibrium at all parent nodes of the IMN. 
The residual vector is defined as
\begin{equation}\label{eq.imn_residual}
    \mathbf{R}
    = 
    \begin{bmatrix}
        \mathbf{R}_1 \\ \mathbf{R}_2 \\ \vdots \\ \mathbf{R}_{2^N-1}
    \end{bmatrix}
    = 
    \begin{bmatrix}
        \mathbf{H}(\vec{\mathbf{n}}_0^1)^T (\boldsymbol{\sigma}_1^1 - \boldsymbol{\sigma}_1^2) \\
        \mathbf{H}(\vec{\mathbf{n}}_1^1)^T (\boldsymbol{\sigma}_2^1 - \boldsymbol{\sigma}_2^2) \\
        \vdots\\
        \mathbf{H}(\vec{\mathbf{n}}_{N-1}^{2^{N-1}})^T (\boldsymbol{\sigma}_N^{2^N-1} - \boldsymbol{\sigma}_N^{2^N})
    \end{bmatrix}.
\end{equation}
Using the stress homogenization relation in Eq. \eqref{eq.imn_avg_stress}, the residual vector $\mathbf{R}$ can be expressed in terms of the stresses at the base nodes as
\begin{equation}\label{eq.imn_residual2}
    \mathbf{R}
    = \mathbf{A}^T \mathbf{W}
    \begin{bmatrix}
        \boldsymbol{\sigma}_N^1 \\ \boldsymbol{\sigma}_N^2 \\ \vdots \\ \boldsymbol{\sigma}_N^{2^N}
    \end{bmatrix}
    = \mathbf{A}^T \mathbf{W} \boldsymbol{\sigma}^{\text{nodes}}
    = \mathbf{A}^T \mathbf{W} \mathbf{K} \boldsymbol{\varepsilon}^{\text{nodes}}
\end{equation}
where the block-diagonal matrices $\mathbf{W}$ and $\mathbf{K}$ are defined as
\begin{equation}\label{eq.imn_W_K}
    \mathbf{W}
    =
    \begin{bmatrix}
        \mathbf{W}^1 & \mathbf{0}_{6\times6} & \hdots & \mathbf{0}_{6\times6} \\
        \mathbf{0}_{6\times6} & \mathbf{W}^2 & \hdots & \mathbf{0}_{6\times6} \\
        \vdots & \vdots & \vdots & \vdots \\
        \mathbf{0}_{6\times6} & \mathbf{0}_{6\times6} & \hdots & \mathbf{W}^{2^N}
    \end{bmatrix},
    \mathbf{K}
    =
    \begin{bmatrix}
        \mathbf{C}^1 & \mathbf{0}_{6\times6} & \hdots & \mathbf{0}_{6\times6} \\
        \mathbf{0}_{6\times6} & \mathbf{C}^2 & \hdots & \mathbf{0}_{6\times6} \\
        \vdots & \vdots & \vdots & \vdots \\
        \mathbf{0}_{6\times6} & \mathbf{0}_{6\times6} & \hdots & \mathbf{C}^{2^N}
    \end{bmatrix}.
\end{equation}
Here, $\mathbf{W}^n = w_N^n \mathbf{I}_{6\times6}$ and $\mathbf{C}^n$ denote the weight matrix and the tangent stiffness matrix of the $n$-th base node, respectively.

If a base node has a null weight, the associated residual is undefined and must be removed from the residual vector $\mathbf{R}$.
Correspondingly, the rows associated with these base nodes and the columns corresponding to their parent nodes are removed from the matrix $\mathbf{A}$ to ensure a consistent reduced system.

Combining Eqs. \eqref{eq.imn_local_strain} and \eqref{eq.imn_residual2} gives
\begin{equation}\label{eq.imn_residual3}
    \mathbf{R}
    = \mathbf{A}^T \mathbf{W} \mathbf{K} (\mathbf{A}\mathbf{a} + \boldsymbol{\varepsilon}^{\text{macro}}).
\end{equation}
The Jacobian matrix of the residual with respect to the jumping vector is therefore given by
\begin{equation}\label{eq.imn_jacobian}
	\mathbf{J}=\frac{d\mathbf{R}}{d\mathbf{a}} = \mathbf{A}^T \mathbf{W} \mathbf{K} \mathbf{A},
\end{equation}
which is used within a Newton-type iterative scheme to update $\mathbf{a}$ until the residual norm $\lVert \mathbf{R} \rVert$ falls below a prescribed tolerance.
The complete IMN online prediction procedure based on Newton iterations is summarized in Algorithm \ref{algorithm.imn_newton}.
\begin{algorithm}
\caption{IMN online prediction with Newton iterations}\label{algorithm.imn_newton}
\begin{algorithmic}[1]
    \Statex Input: macroscale strain increment $\Delta\boldsymbol{\varepsilon}$
    \Statex Output: macroscale stress increment $\Delta\boldsymbol{\sigma}$
    \State Initialize $\mathbf{a} = \mathbf{0}$, $\mathbf{R} = \mathbf{1}$
    \State Construct IMN's $\mathbf{A}$ and $\mathbf{W}$ matrices
    \While{$\lVert \mathbf{R} \rVert > tol.$ }
        \State Compute local strain increments by Eq. \eqref{eq.imn_local_strain}: $\Delta\boldsymbol{\varepsilon}^{\text{local}} = \mathbf{A}\mathbf{a}$
        \State Update strain increments at base nodes: $\Delta\boldsymbol{\varepsilon}^{\text{nodes}} = \Delta\boldsymbol{\varepsilon}^{\text{macro}} + \Delta\boldsymbol{\varepsilon}^{\text{local}}$
        \State Compute stress increments and tangent stiffness at base nodes by constitutive laws of microscale base materials: $\Delta\boldsymbol{\sigma}^{\text{nodes}}$, $\mathbf{K}$
        \State Evaluate the residual by Eq. \eqref{eq.imn_residual2}: $\mathbf{R} = \mathbf{A}^T \mathbf{W} \Delta\boldsymbol{\sigma}^{\text{nodes}}$
        \State Evaluate the Jacobian: $\mathbf{J} = \mathbf{A}^T \mathbf{W} \mathbf{K} \mathbf{A}$
        \State Compute increment in jumping vector: $\delta \mathbf{a} = - \mathbf{J}^{-1} \mathbf{R}$
        \State Update jumping vector: $\mathbf{a} = \mathbf{a} + \delta \mathbf{a}$
    \EndWhile
    \State Compute macroscale stress increment: $\Delta\boldsymbol{\sigma} = \sum_{n=1}^{2^N} w_N^n \Delta\boldsymbol{\sigma}_N^n$
\end{algorithmic}
\end{algorithm}
\section{Experiments}\label{sec:experiment}
In this section, we investigate how initialization, batch size, training data size, and regularization affect model training and generalization performance. 
Two IMN online prediction algorithms are evaluated, namely the fixed-point iteration scheme and the Newton iteration scheme described in Section \ref{sec:imn_online}.
In addition, DMN and IMN are compared in terms of offline training cost and online prediction performance. 

\begin{figure}[htp]
	\centering
	\includegraphics[width=0.3\linewidth]{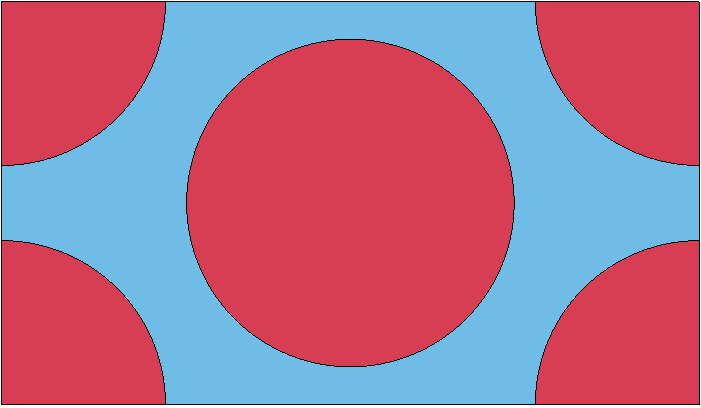}
	\caption{UD fiber-reinforced composite RVE with a fiber volume fraction of 60$\%$.}
	\label{fig.rve}
\end{figure}

\begin{figure}[htp]
	\centering
	\begin{subfigure}{0.325\textwidth}
		\centering
		\includegraphics[width=1\linewidth]{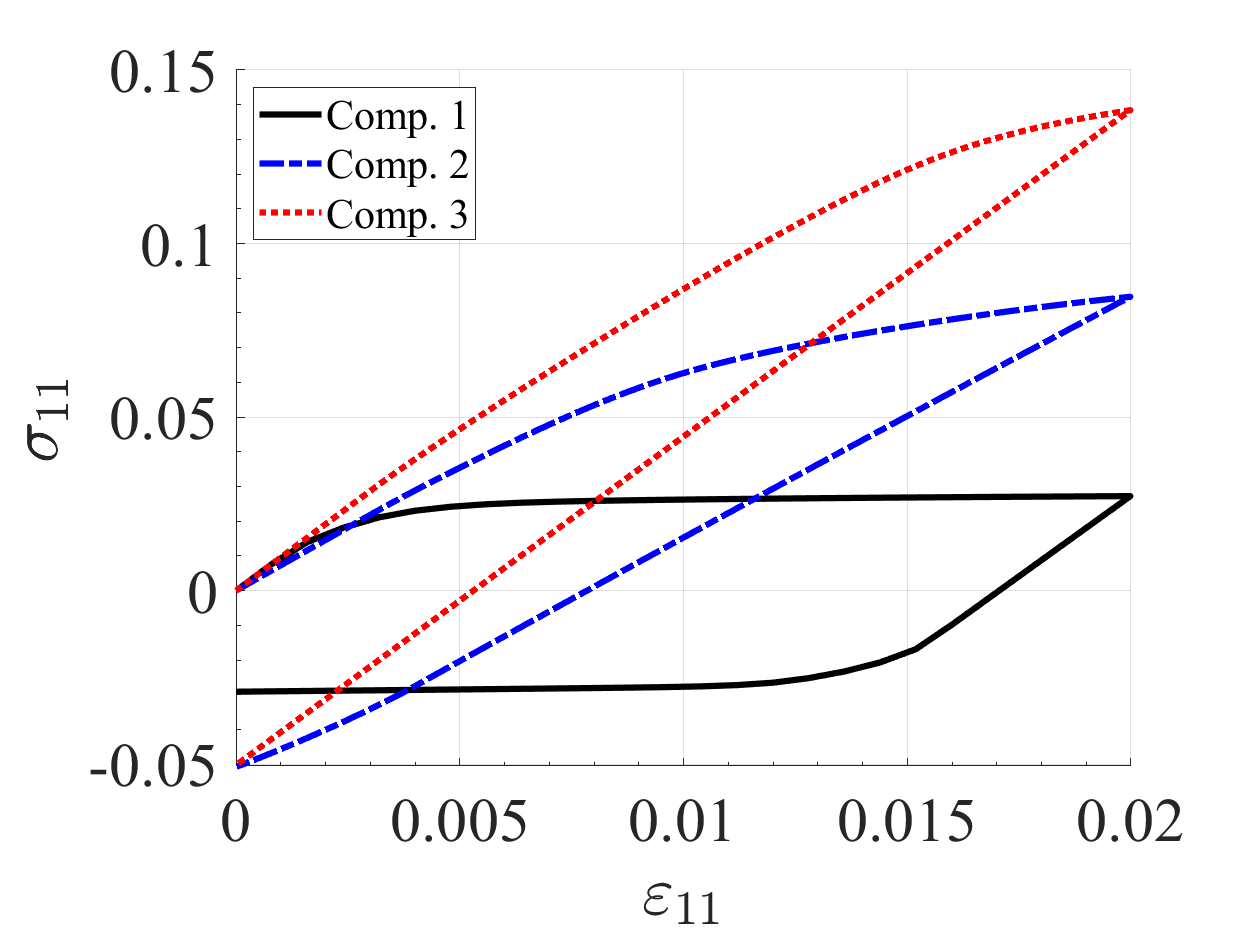}
		\caption{}
	\end{subfigure}
	\begin{subfigure}{0.325\textwidth}
		\centering
		\includegraphics[width=1\linewidth]{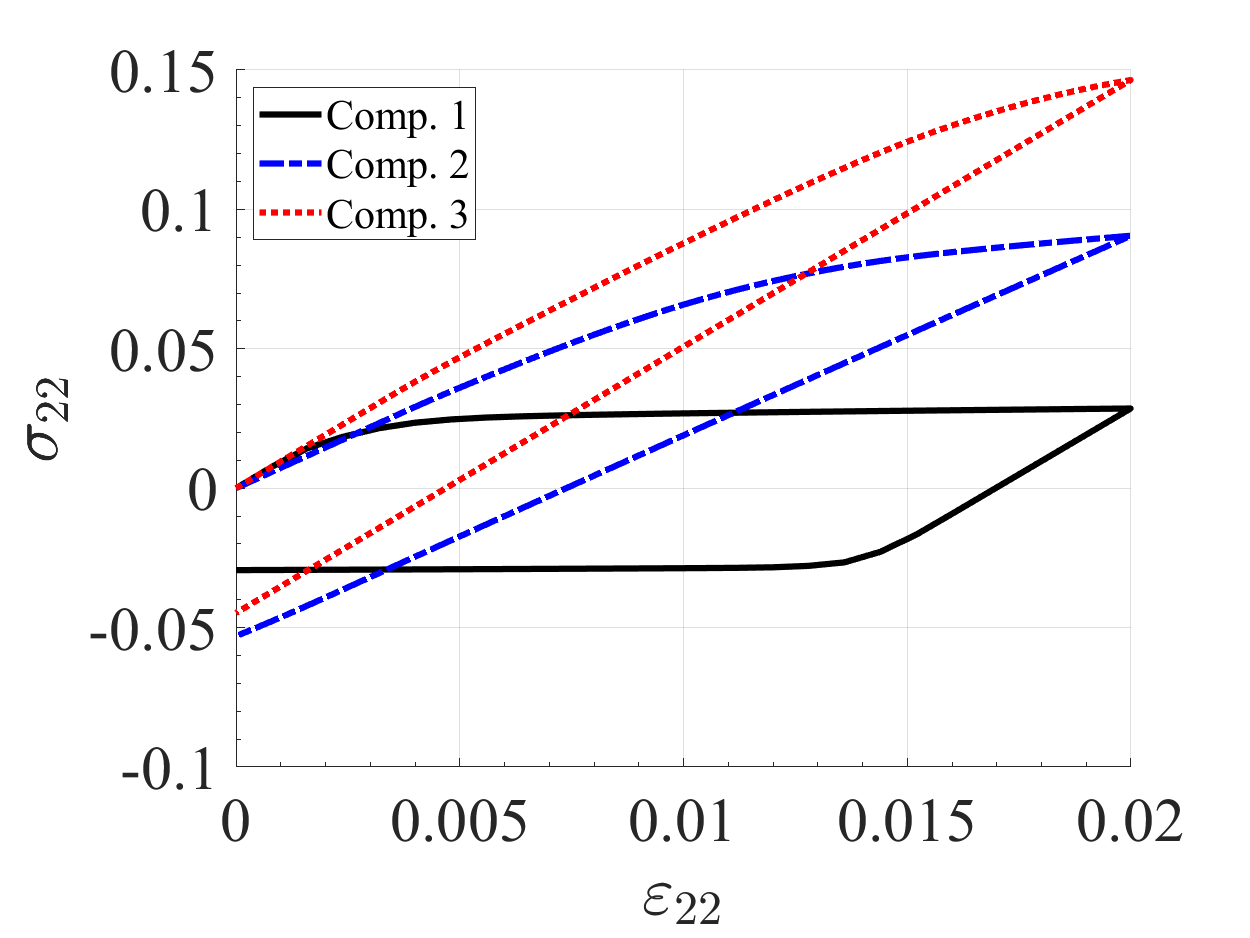}
		\caption{}
	\end{subfigure}
	\begin{subfigure}{0.325\textwidth}
		\centering
		\includegraphics[width=1\linewidth]{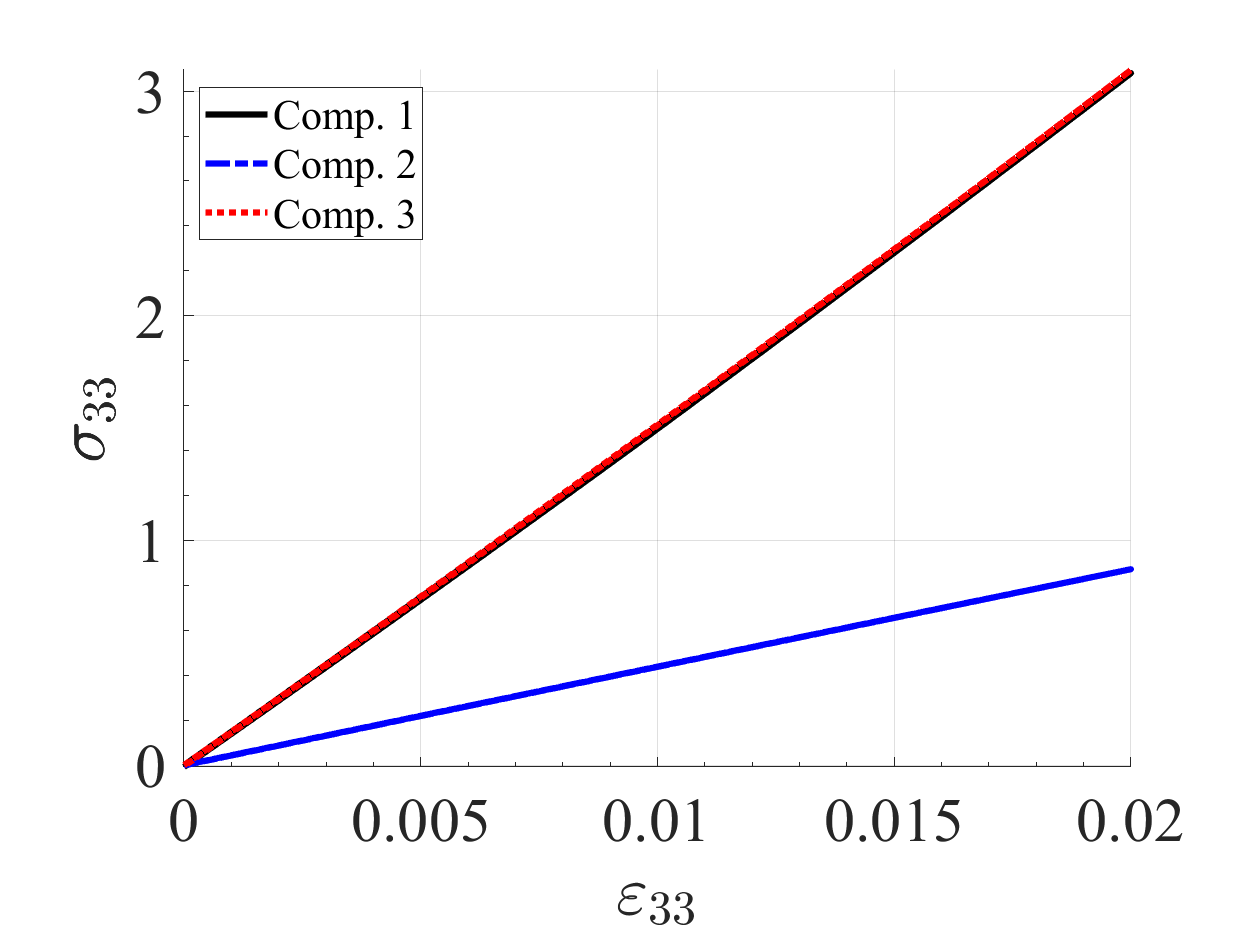}
		\caption{}
	\end{subfigure}
	\begin{subfigure}{0.325\textwidth}
		\centering
		\includegraphics[width=1\linewidth]{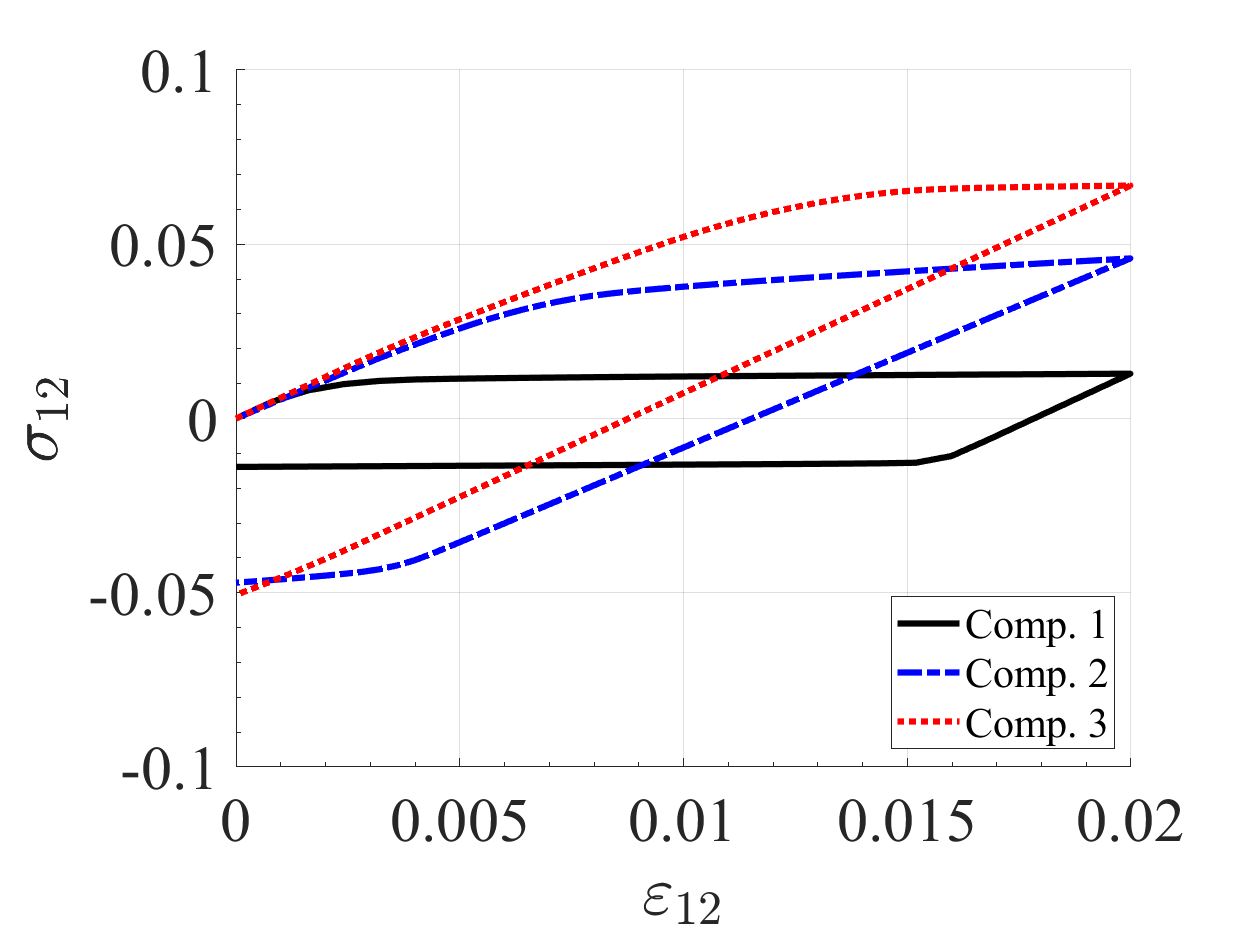}
		\caption{}
	\end{subfigure}
	\begin{subfigure}{0.325\textwidth}
		\centering
		\includegraphics[width=1\linewidth]{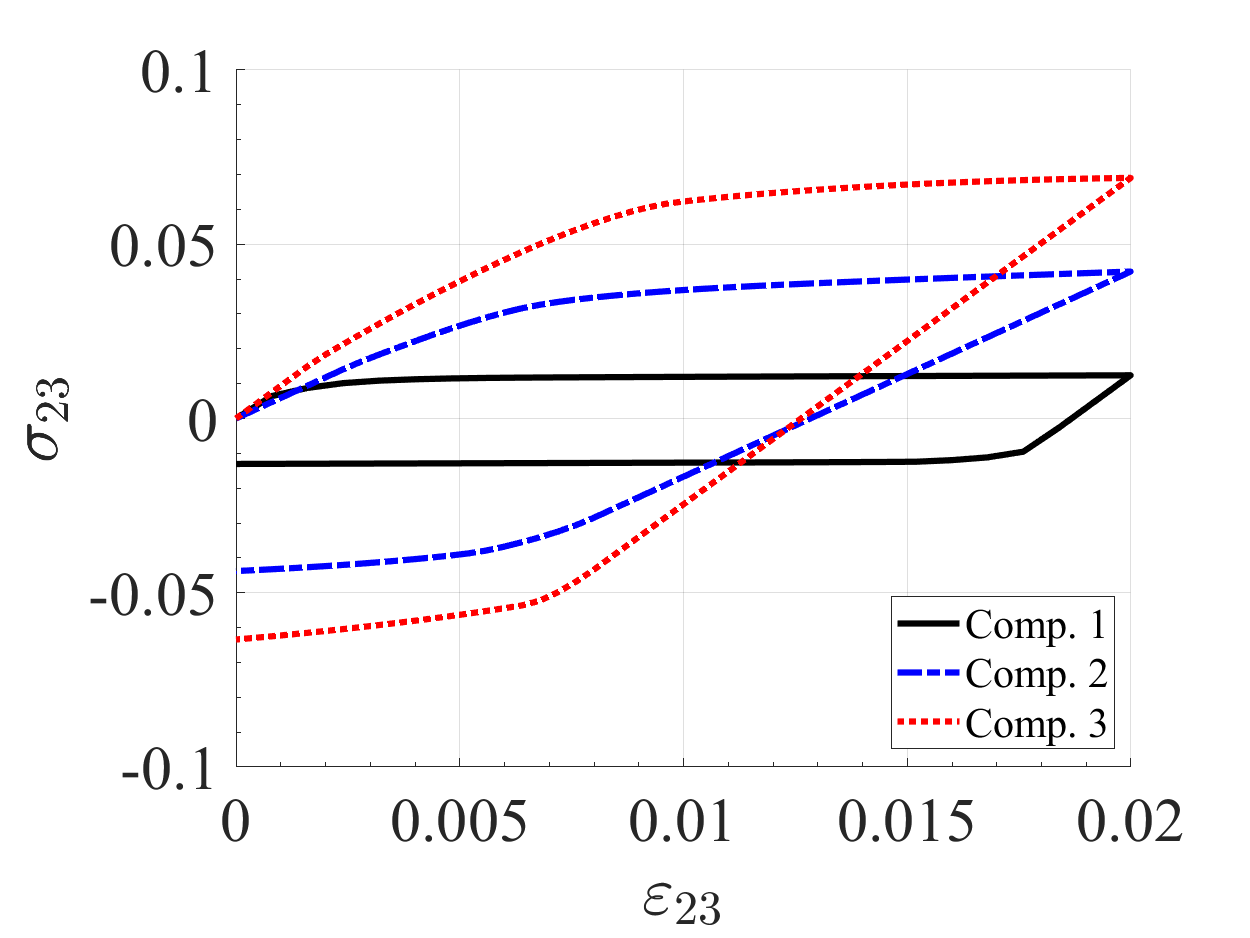}
		\caption{}
	\end{subfigure}
	\begin{subfigure}{0.325\textwidth}
		\centering
		\includegraphics[width=1\linewidth]{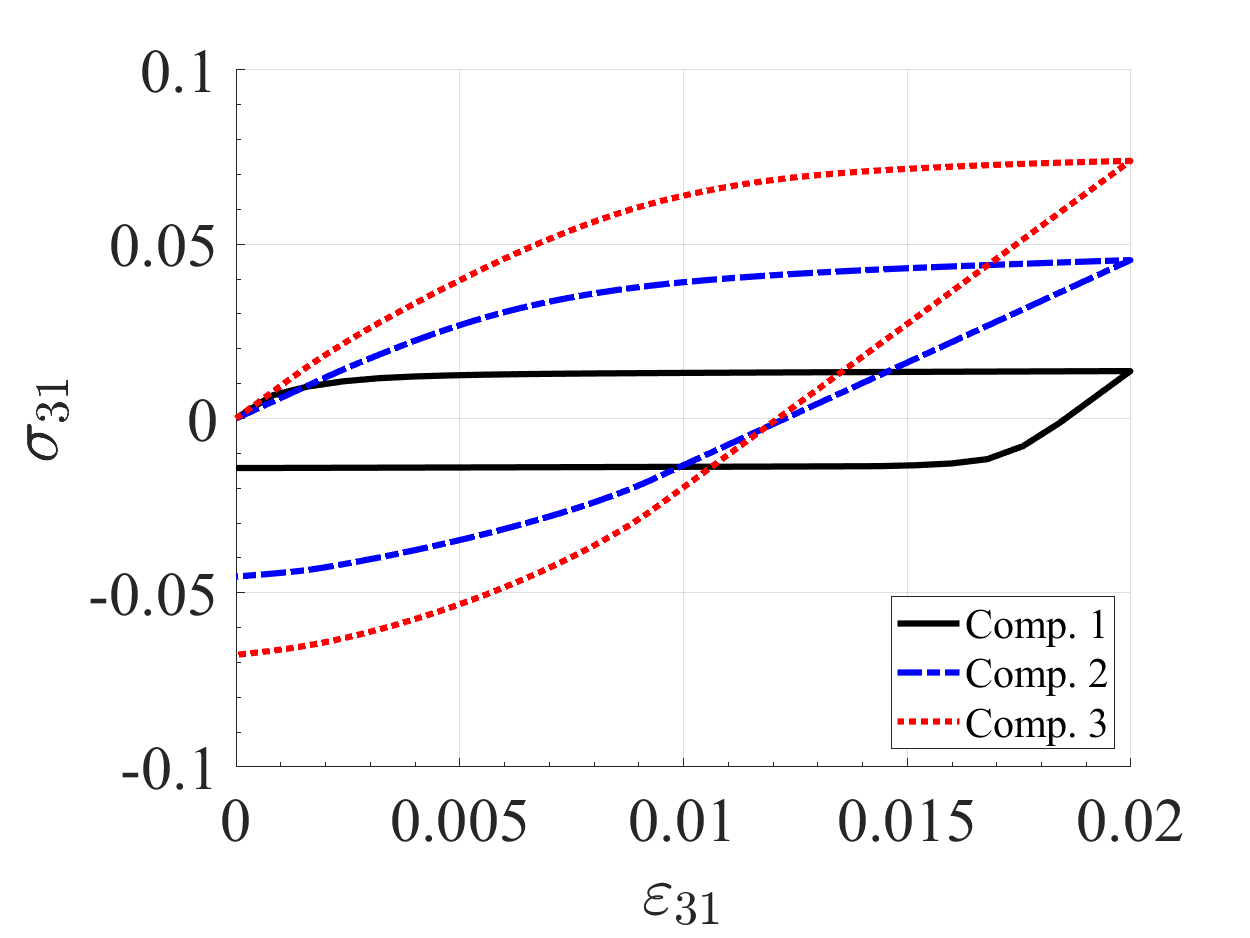}
		\caption{}
	\end{subfigure}
	\caption{Material responses of three composites for online testing: (a) $\varepsilon_{11} \sim \sigma_{11}$; (b) $\varepsilon_{22} \sim \sigma_{22}$; (c) $\varepsilon_{33} \sim \sigma_{33}$; (d) $\varepsilon_{12} \sim \sigma_{12}$; (e) $\varepsilon_{23} \sim \sigma_{23}$; (f) $\varepsilon_{31} \sim \sigma_{31}$.}\label{fig.mat_data}
\end{figure}

To assess model generalization and robustness, three UD fiber-reinforced composites with a fiber volume fraction of 60$\%$ (as shown in Fig. \ref{fig.rve}) are considered for testing.
These composites differ in phase contrast and inelastic material behavior.
The material parameters for the matrix and fiber phases are summarized in Tables \ref{tab:mat_param_matrix}-\ref{tab:mat_param_fiber}.
In the present study, all DNS are performed in LS-DYNA using the specialized RVE analysis module. Using the $\ast$RVE\_ANALYSIS\_FEM keyword \cite{wei2022rve}, periodic boundary conditions are automatically imposed and homogenized stresses are computed.
The homogenized material responses under six independent loading conditions are shown in Fig. \ref{fig.mat_data}, including uniaxial loading in the three normal directions (Fig. \ref{fig.mat_data}(a)-(c)) and three simple shear loading cases (Fig. \ref{fig.mat_data}(d)-(f)). 
The three composites considered for testing are defined as follows.
\begin{itemize}
	\item \textbf{Composite 1}: Matrix: isotropic elastoplastic material with combined linear and exponential hardening; Fiber: orthotropic linear elastic material.
	\item \textbf{Composite 2}: Matrix: isotropic elastoplastic material with combined linear and exponential hardening; Fiber: isotropic linear elastic material.
	\item \textbf{Composite 3}: Matrix: isotropic elastoplastic material with exponential hardening; Fiber: orthotropic linear elastic material.
\end{itemize}

The online prediction accuracy is quantified using the mean relative error of the homogenized stress across the six loading cases, defined as
\begin{equation}\label{eq.metric_stress}
	e_{\boldsymbol{\sigma}} = \frac{1}{6} \sum_{i=1}^6 \frac{\lVert \boldsymbol{\sigma}_i - \hat{\boldsymbol{\sigma}}_i \rVert_{L_2}}{\lVert \boldsymbol{\sigma}_i\rVert_{L_2}},
\end{equation}
where $\boldsymbol{\sigma}_i$ denotes the reference homogenized stress responses under the $i$-th loading condition and $\hat{\boldsymbol{\sigma}}_i$ is the corresponding model prediction.

\begin{table}[htp]
	\centering
	\caption{Material parameters of the matrix phase of three composites for online testing.}
	\begin{tabular}{cccc}
		\hline
		& Composite 1 & Composite 2 & Composite 3 \\
		\hline
		Young's modulus (GPa) & 3.8 & 2.1 & 3.8 \\
		\hline
		Poisson ratio & 0.387 & 0.3 & 0.387 \\
		\hline
		Initial yield strength (GPa) & 0.01 & 0.029 & 0.03 \\
		\hline
	\end{tabular}
	\label{tab:mat_param_matrix}
\end{table}

\begin{table}[htp]
	\centering
	\caption{Material parameters of the fiber phase of three composites for online testing.}
	\begin{tabular}{cccc}
		\hline
		& Composite 1 & Composite 2 & Composite 3 \\
		\hline
		$E_{11}$ (GPa) & 19.8 & 72 & 19.8 \\
		\hline
		$E_{22}$ (GPa) & 19.8 & 72 & 19.8 \\
		\hline
		$E_{33}$ (GPa) & 245 & 72 & 245 \\
		\hline
		$G_{12}$ (GPa) & 5.9 & 29.5 & 5.9 \\
		\hline
		$G_{13}$ (GPa) & 29.2 & 29.5 & 29.2 \\
		\hline
		$G_{23}$ (GPa) & 29.2 & 29.5 & 29.2 \\
		\hline
		$\nu_{12}$ & 0.67 & 0.22 & 0.67 \\
		\hline
		$\nu_{13}$ & 0.02 & 0.22 & 0.02 \\
		\hline
		$\nu_{23}$ & 0.02 & 0.22 & 0.02 \\
		\hline
	\end{tabular}
	\label{tab:mat_param_fiber}
\end{table}

Both DMN and IMN models are implemented in PyTorch \cite{paszke2019pytorch} and trained with the same loss function defined in Eq. \eqref{eq.dmn_loss}.
All models are trained for 10,000 epochs using the Adam optimizer with an initial learning rate of $10^{-2}$.
A learning scheduler is adopted to reduce the learning rate by a factor of 0.8 if the validation loss does not improve for 50 consecutive epochs.
All offline training is conducted on 20 cores of an Intel(R) Xeon(R) W7-3465X CPU, while all online predictions are performed on an 11th Gen Intel(R) i7-11850H CPU.

\subsection{Effects of Initializations, Batch Size, and Data Size}\label{sec:dmn_ns400}
To investigate the effects of initialization, batch size, and training data size on model training and generalization performance, DMNs are trained using three datasets with varying sample sizes and batch configurations: 
\begin{itemize}
	\item \textbf{Dataset 1}: 400 samples for training; 100 samples for validation; batch sizes: 20, 40, 128
	\item \textbf{Dataset 2}: 1024 samples for training; 100 samples for validation; batch sizes: 40, 64, 128
	\item \textbf{Dataset 3}: 2048 samples for training; 100 samples for validation; batch sizes: 128, 256
\end{itemize}
The training data consists of $\{\tilde{\mathbf{C}}_j^{p1}, \tilde{\mathbf{C}}_j^{p2}, \mathbf{C}_j\}_{j=1}^{N_s}$, where $\tilde{\mathbf{C}}_j^{p1}$ and $\tilde{\mathbf{C}}_j^{p2}$ are the linear elastic stiffness matrices of the two constituent base materials for the $j$-th sample, and $\mathbf{C}_j$ is the corresponding homogenized stiffness matrix of the RVE.
In this study, the base material stiffness matrices $\{\tilde{\mathbf{C}}^{p1}, \tilde{\mathbf{C}}^{p2}\}$ are sampled following the procedure proposed in \cite{liu2019exploring}, which is based on orthotropic elasticity tensors. 
The homogenized RVE stiffness matrix $\mathbf{C}$ is obtained by DNS of the RVE.

For each combination of training data size and batch size, the DMN is trained with 10 random initializations.
In total, 80 DMN models are trained and then evaluated through online testing on three composite materials whose base material properties are not included in the training set.
A key advantage of the structure-preserving DMN is that, despite being trained exclusively on linear elastic data, it can accurately predict nonlinear inelastic responses of composites composed of new base materials.

\begin{figure}[htp]
\centering
    \begin{subfigure}{0.325\textwidth}
        \centering
        \includegraphics[width=1\linewidth]{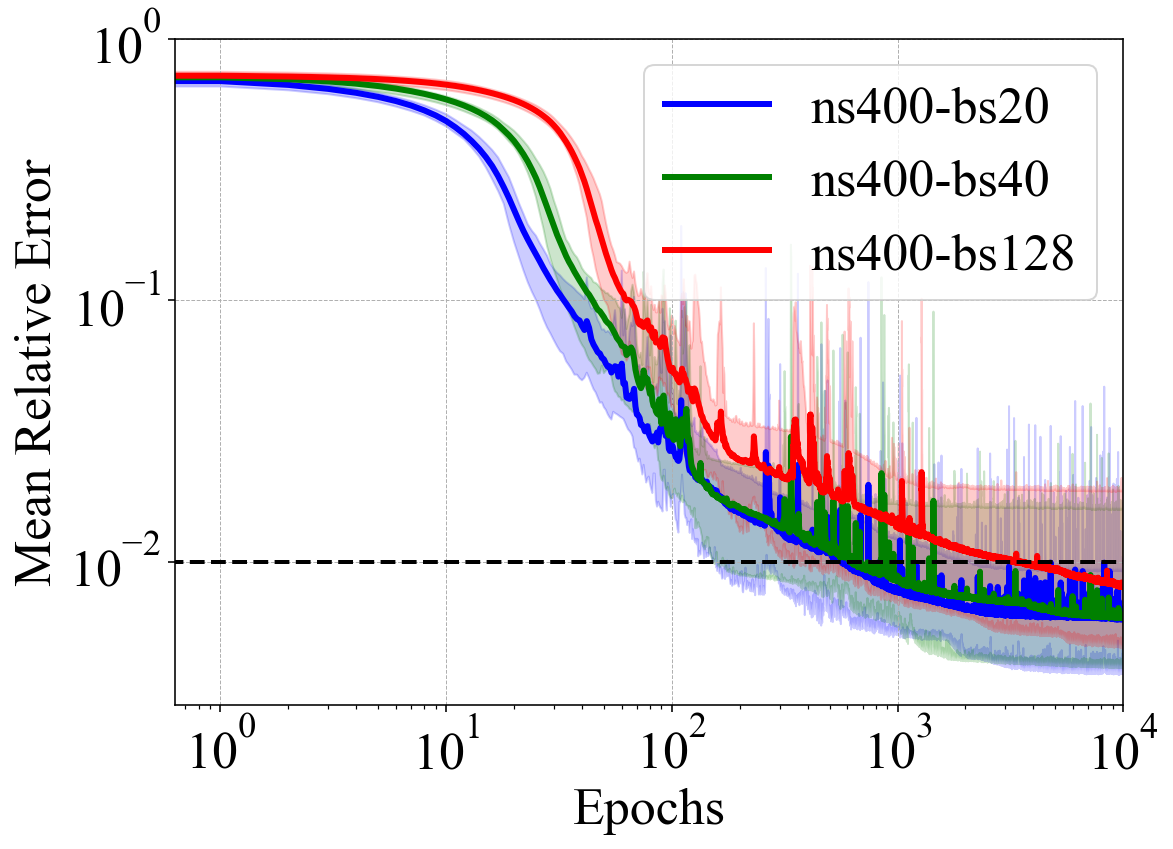}
        \caption{Dataset 1}
    \end{subfigure}
    \begin{subfigure}{0.325\textwidth}
        \centering
        \includegraphics[width=1\linewidth]{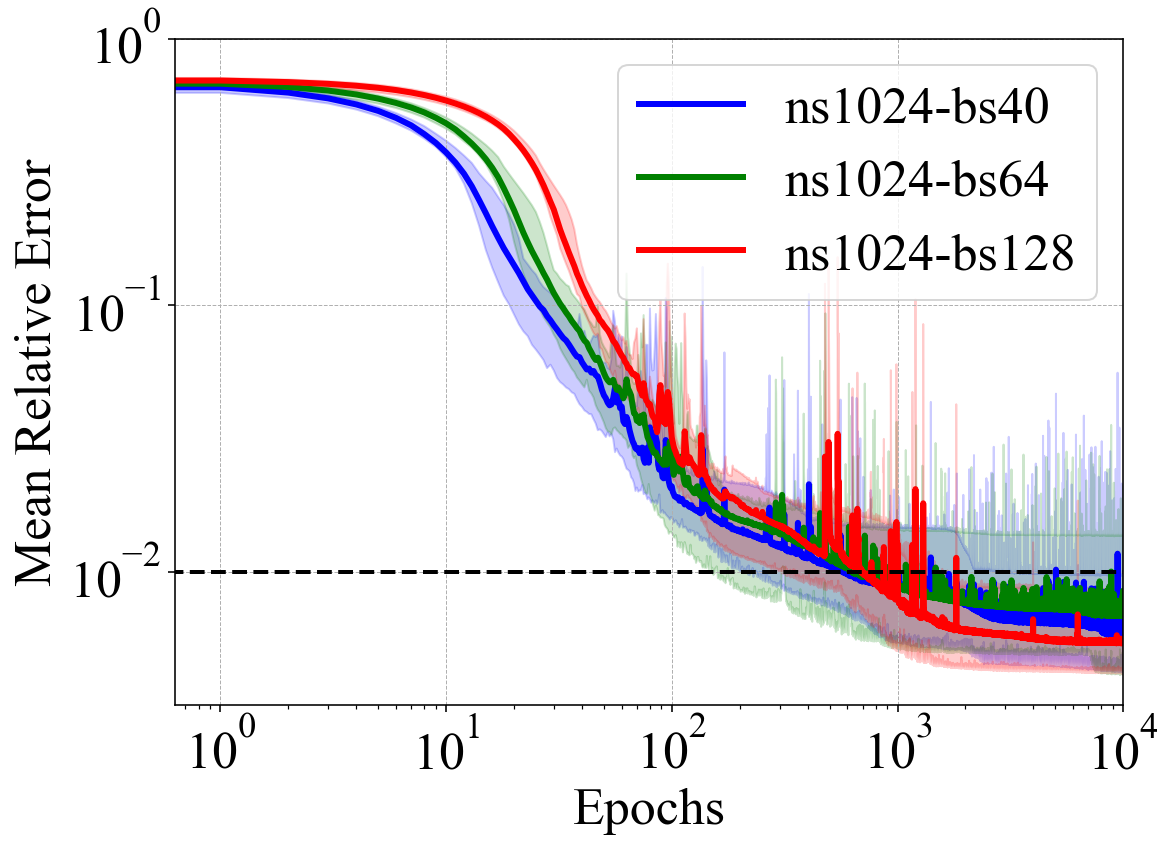}
        \caption{Dataset 2}
    \end{subfigure}
    \begin{subfigure}{0.325\textwidth}
        \centering
        \includegraphics[width=1\linewidth]{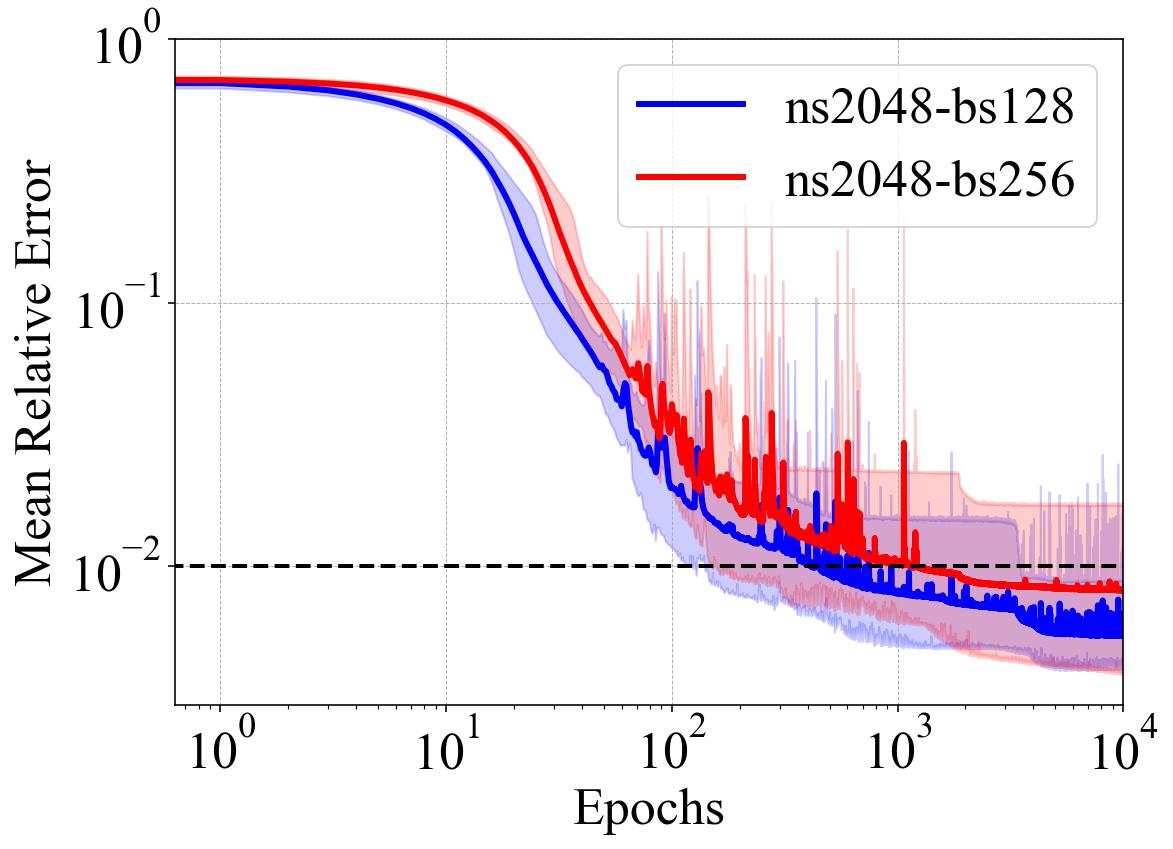}
        \caption{Dataset 3}
    \end{subfigure}
\caption{Training histories of mean relative errors of DMNs trained with: (a) 400 samples using batch sizes of 20, 40, and 128; (b) 1024 samples using batch sizes of 40, 64, and 128; (c) 2048 samples using batch sizes of 128 and 256. Solid lines show the mean over 10 runs with random initializations, and the shaded regions indicate the corresponding standard deviations.}
\label{fig.dmn_loss_data_size}
\end{figure}

Fig. \ref{fig.dmn_loss_data_size} shows the training histories of mean relative errors defined in Eq. \eqref{eq.error_metric}. Solid lines show the mean over 10 runs with random initializations, and the shaded regions indicate the corresponding standard deviations, representing the uncertainty arising from initialization. 
\begin{equation}\label{eq.error_metric}
    e_{\mathbf{C}} = \frac{1}{N_s}\sum_{j=1}^{N_s}
    \sqrt{\frac{||\mathbf{C}_{j} - \hat{\mathbf{C}}_{j}||^2}{||\mathbf{C}_{j}||^2}},
\end{equation}
Since the validation errors closely track the training errors, only the latter are reported here.
The results indicate that batch sizes of 20 or 40 yield the lowest errors for Dataset 1, whereas a batch size of 128 performs best for both Dataset 2 and Dataset 3.
Although no monotonic relationship between data size and batch size is observed, these results suggest that smaller datasets tend to benefit from smaller batch sizes, whereas larger datasets may favor larger batch sizes.

\begin{figure}[htp]
\centering
    \begin{subfigure}{0.325\textwidth}
        \centering
        \includegraphics[width=1\linewidth]{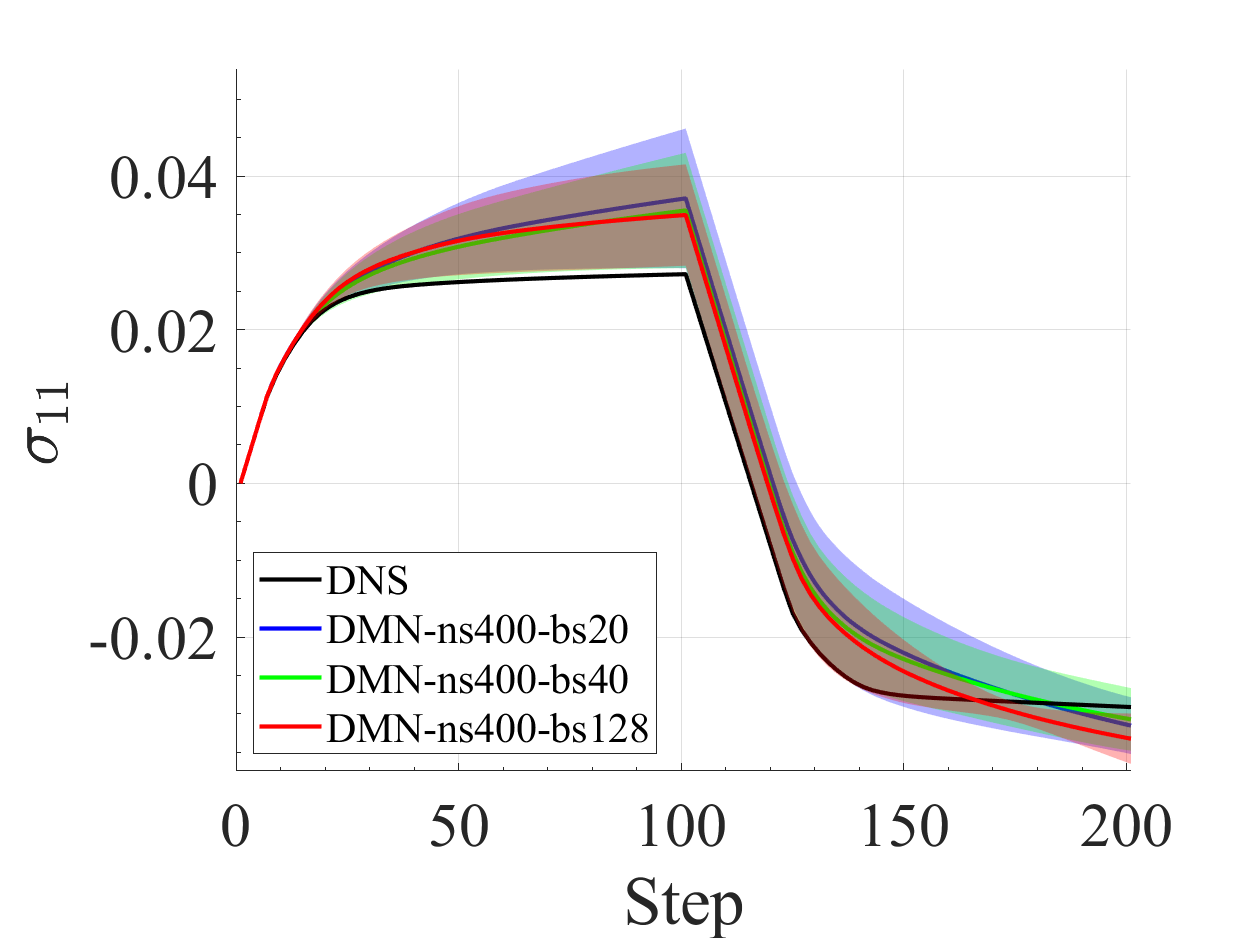}
        \caption{$\sigma_{11}$}
    \end{subfigure}
    \begin{subfigure}{0.325\textwidth}
        \centering
        \includegraphics[width=1\linewidth]{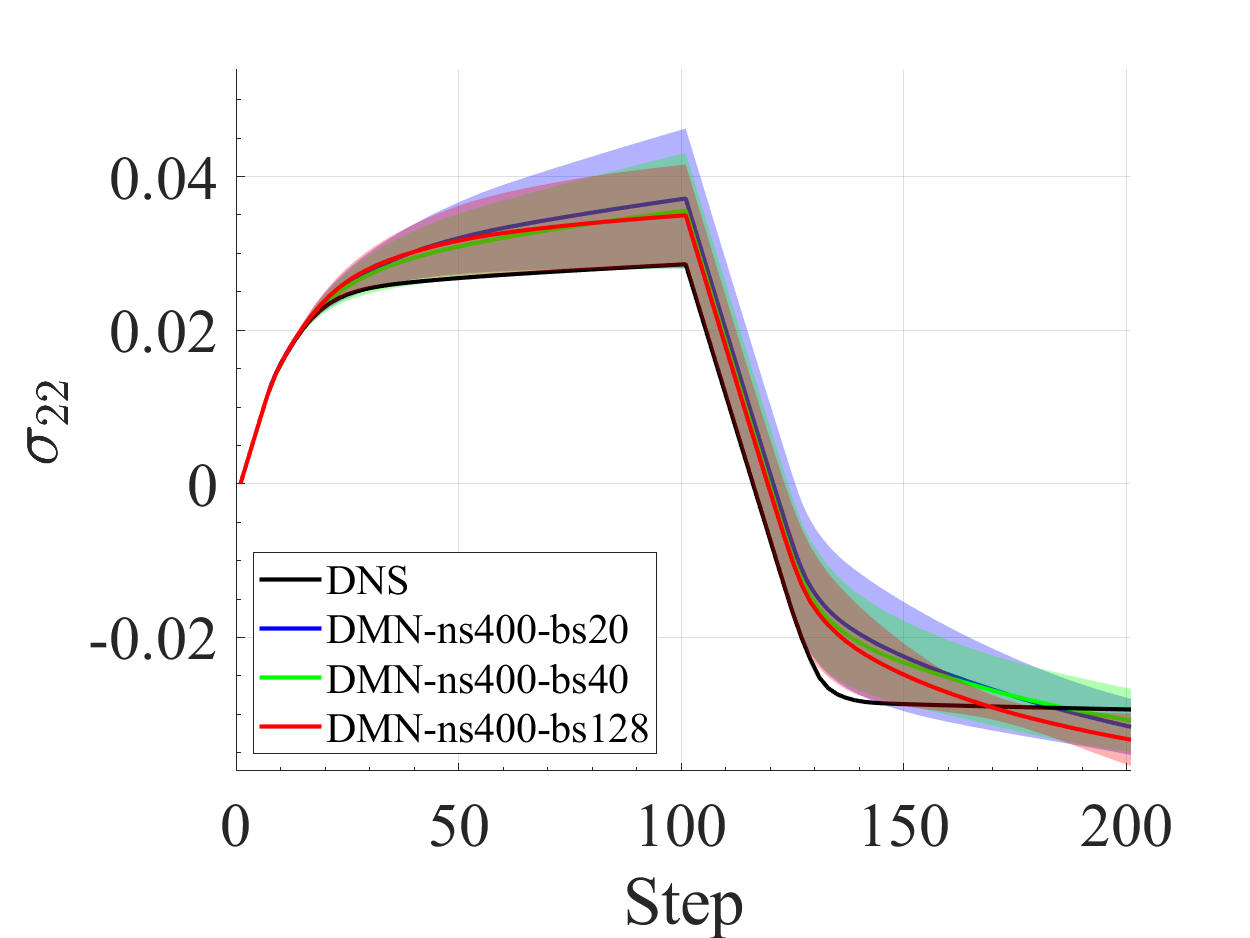}
        \caption{$\sigma_{22}$}
    \end{subfigure}
    \begin{subfigure}{0.325\textwidth}
        \centering
        \includegraphics[width=1\linewidth]{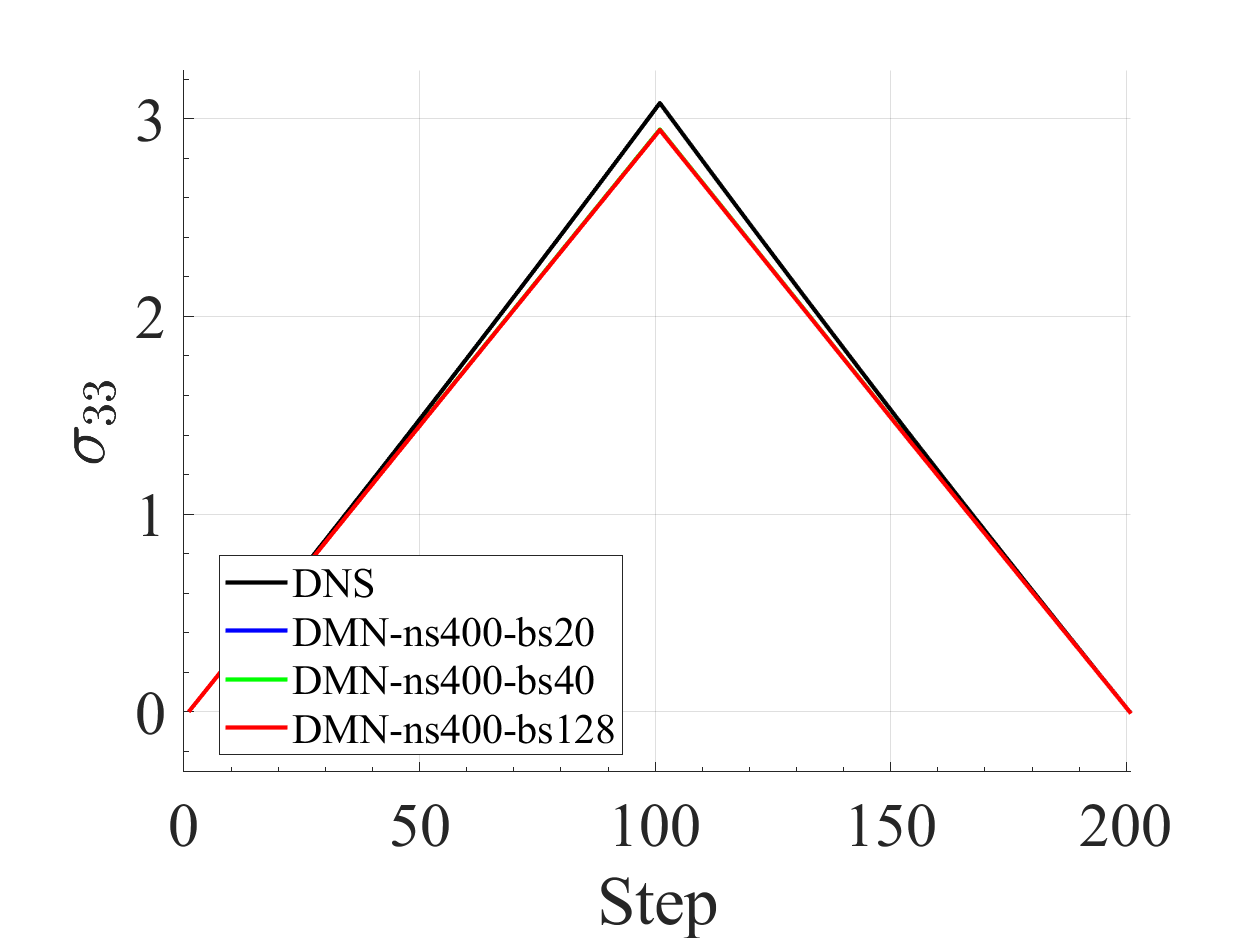}
        \caption{$\sigma_{33}$}
    \end{subfigure}
    \begin{subfigure}{0.325\textwidth}
        \centering
        \includegraphics[width=1\linewidth]{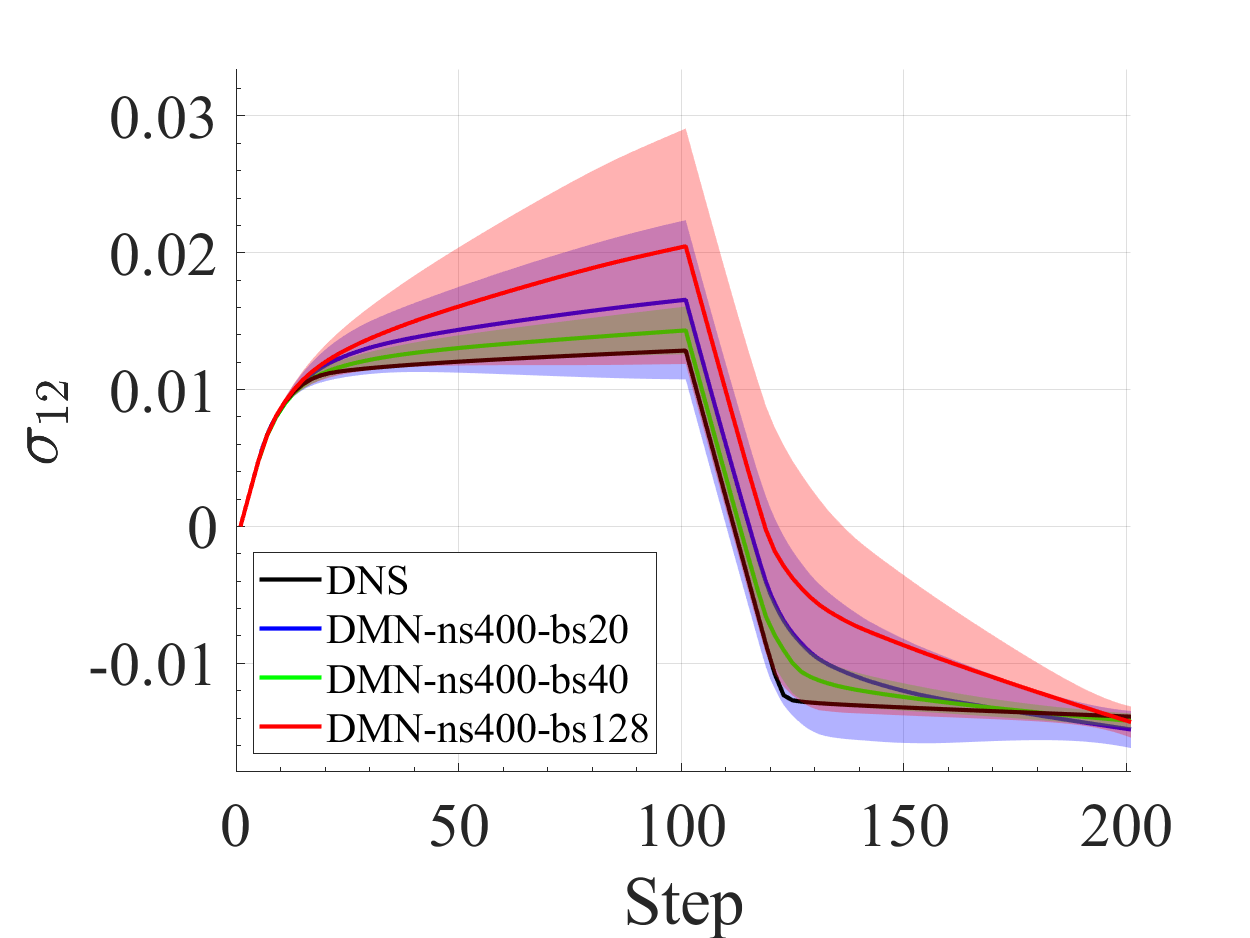}
        \caption{$\sigma_{12}$}
    \end{subfigure}
    \begin{subfigure}{0.325\textwidth}
        \centering
        \includegraphics[width=1\linewidth]{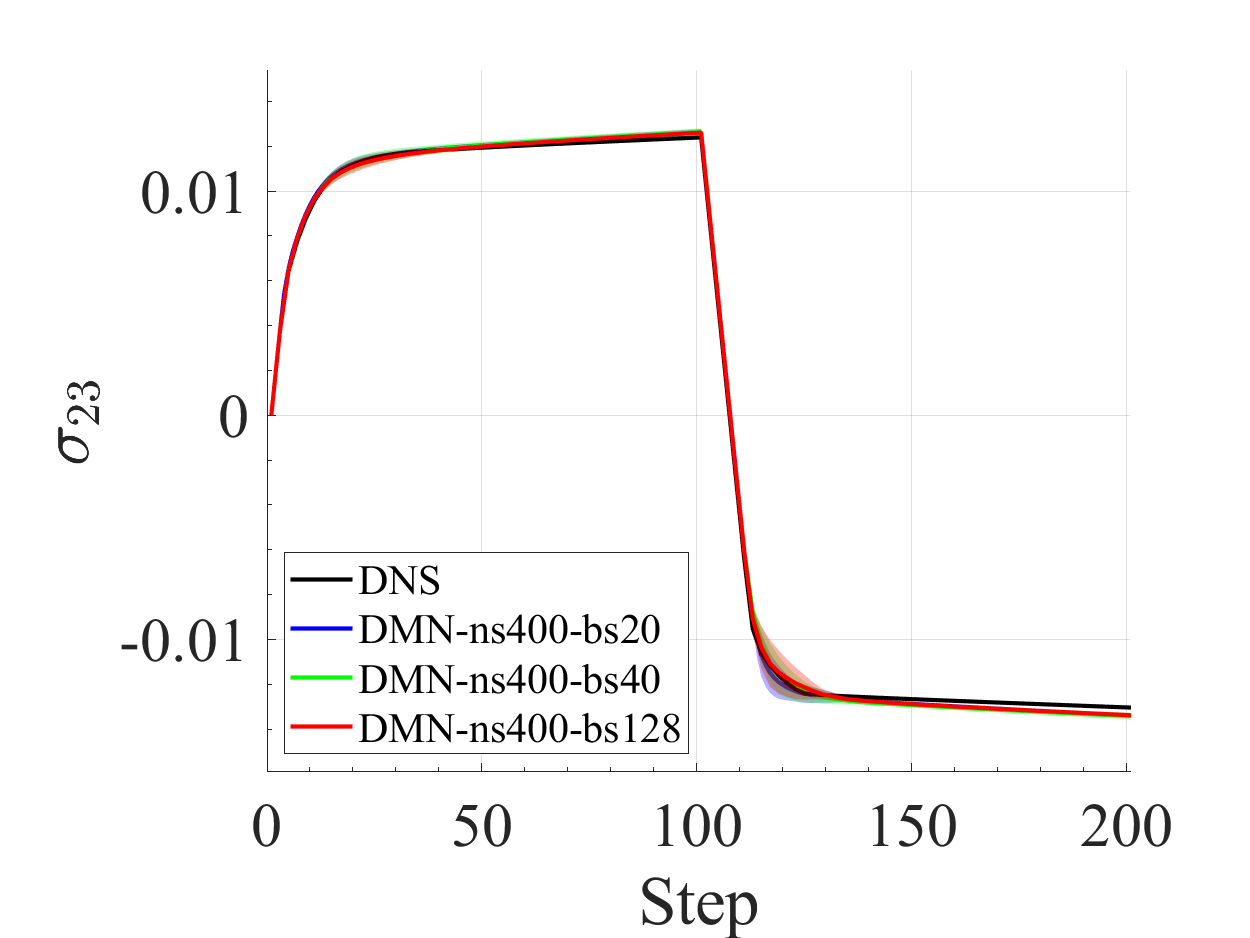}
        \caption{$\sigma_{23}$}
    \end{subfigure}
    \begin{subfigure}{0.325\textwidth}
        \centering
        \includegraphics[width=1\linewidth]{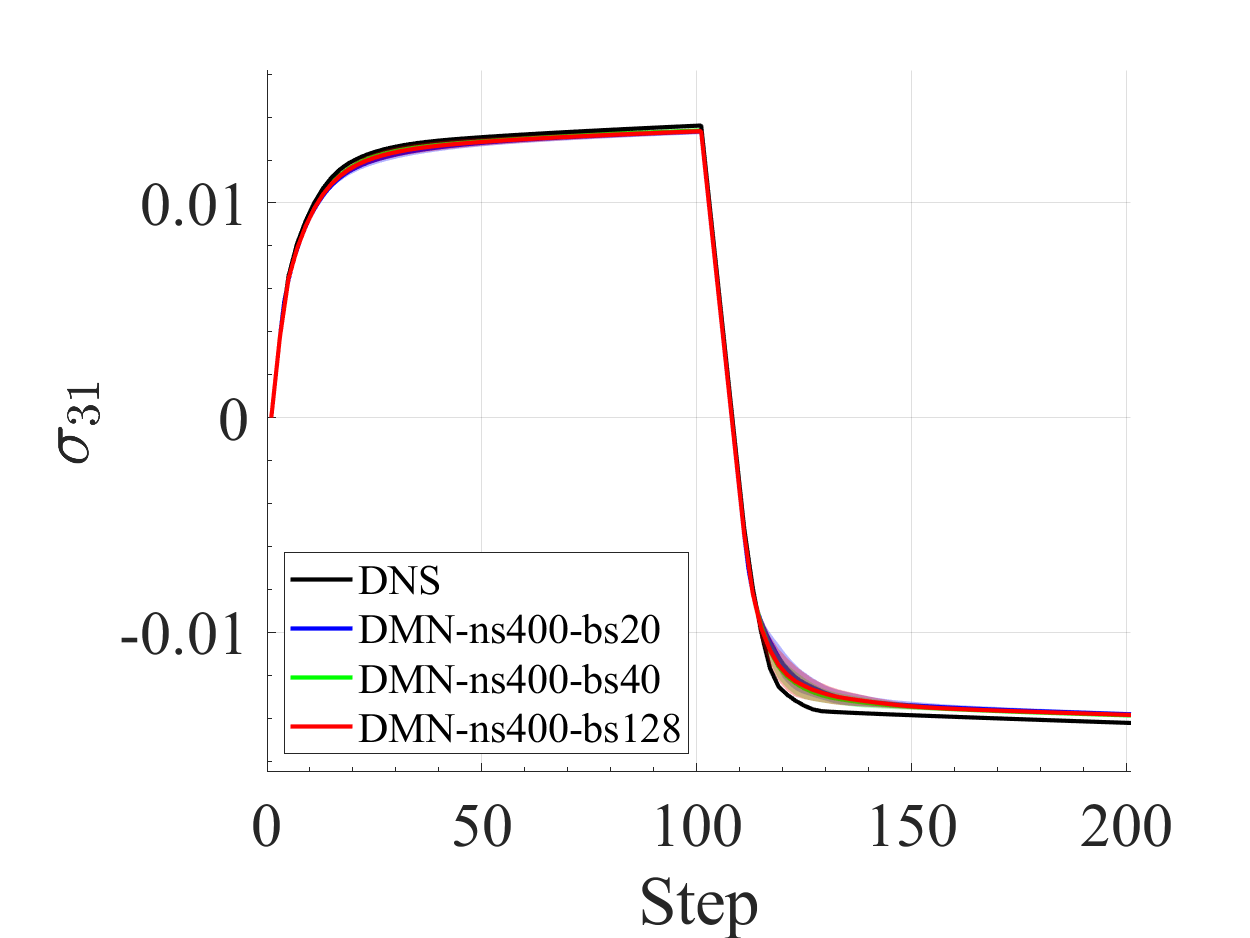}
        \caption{$\sigma_{31}$}
    \end{subfigure}
\caption{Composite 1 - Stress predictions from DMNs trained with 400 samples using batch sizes of 20, 40, and 128: (a) $\sigma_{11}$; (b) $\sigma_{22}$; (c) $\sigma_{33}$; (d) $\sigma_{12}$; (e) $\sigma_{23}$; (f) $\sigma_{31}$; Solid lines show the mean over 10 runs with random initializations, and the shaded regions indicate the corresponding standard deviations.}\label{fig.mat1_ns400}
\end{figure}

\begin{figure}[htp]
\centering
    \begin{subfigure}{0.325\textwidth}
        \centering
        \includegraphics[width=1\linewidth]{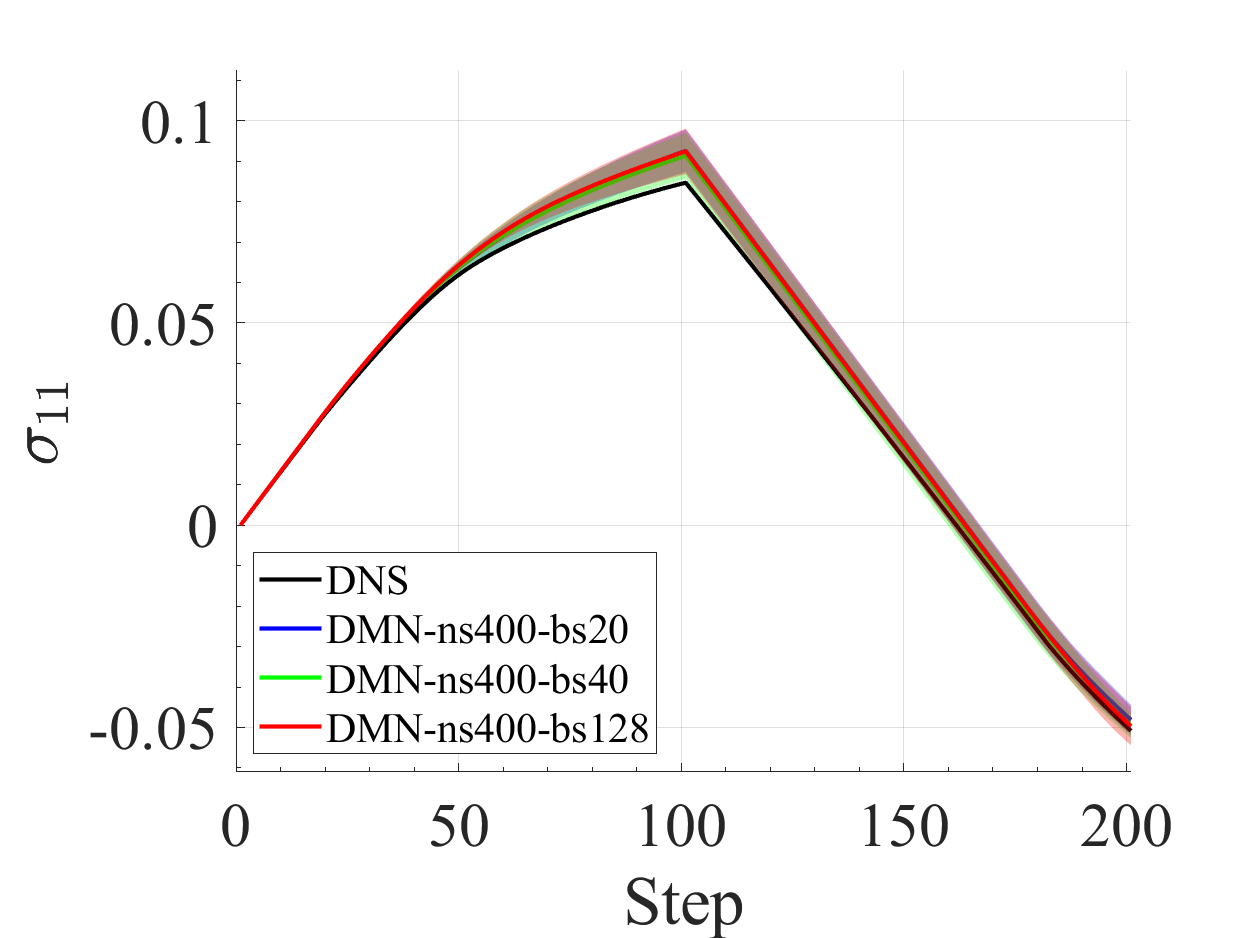}
        \caption{$\sigma_{11}$}
    \end{subfigure}
    \begin{subfigure}{0.325\textwidth}
        \centering
        \includegraphics[width=1\linewidth]{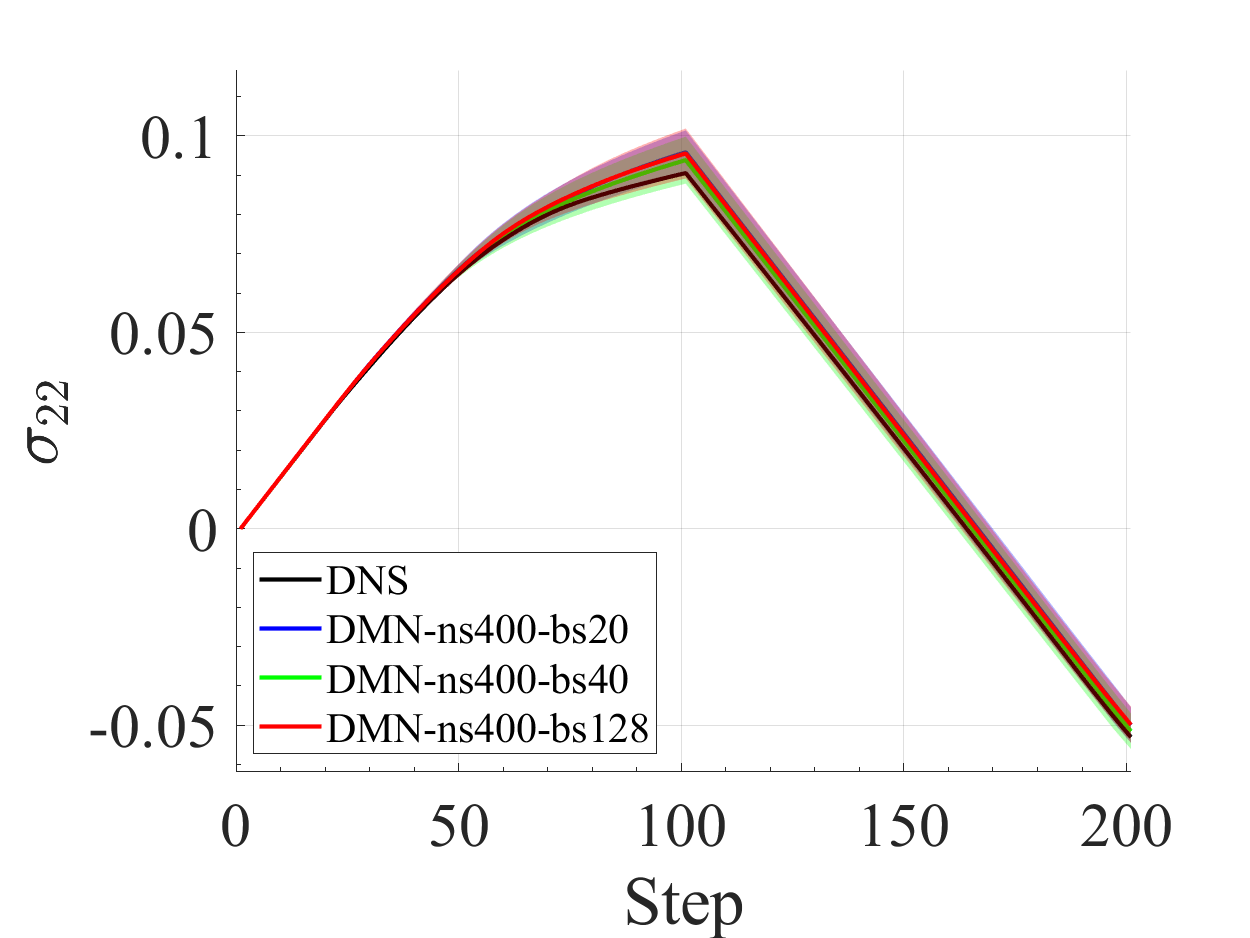}
        \caption{$\sigma_{22}$}
    \end{subfigure}
    \begin{subfigure}{0.325\textwidth}
        \centering
        \includegraphics[width=1\linewidth]{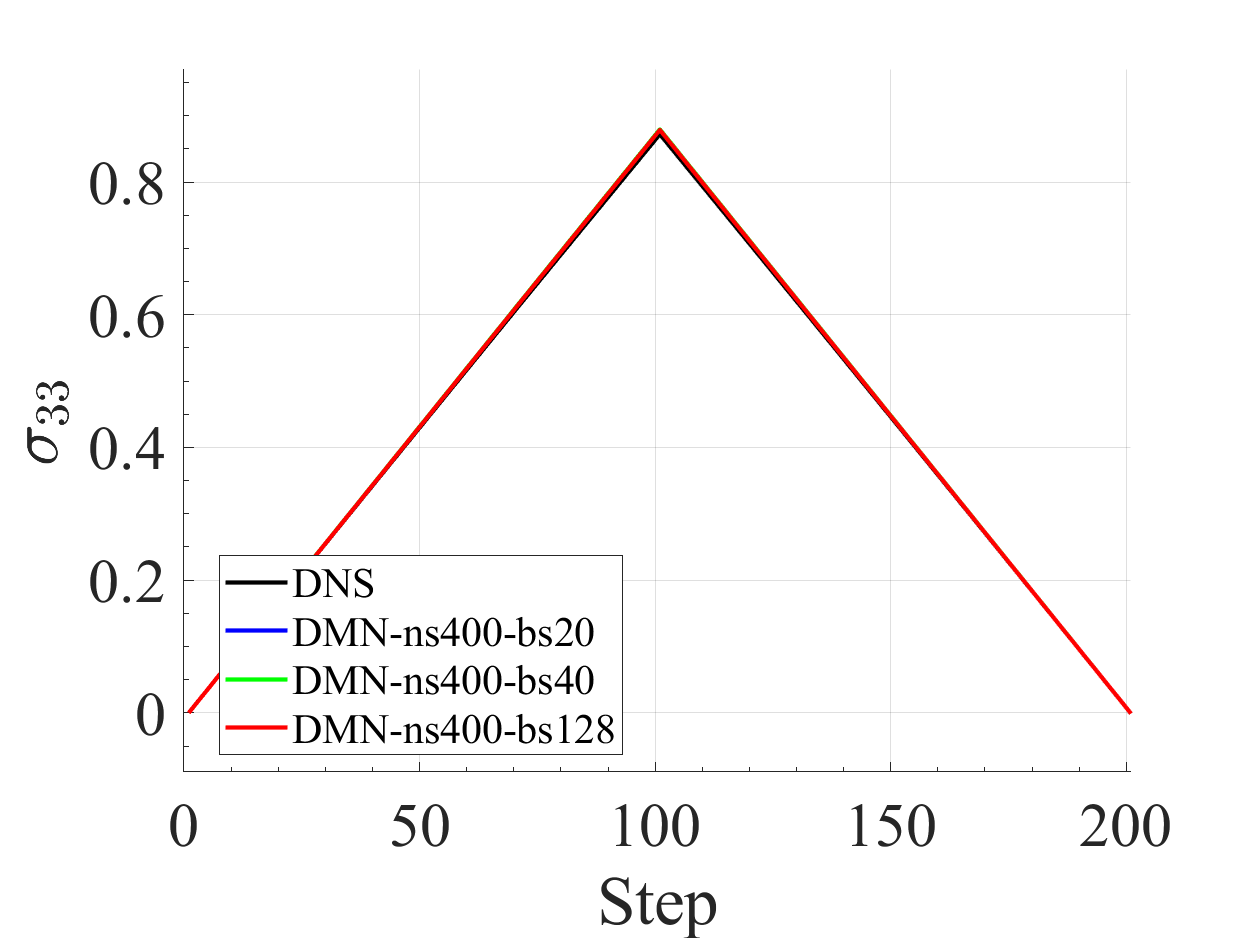}
        \caption{$\sigma_{33}$}
    \end{subfigure}
    \begin{subfigure}{0.325\textwidth}
        \centering
        \includegraphics[width=1\linewidth]{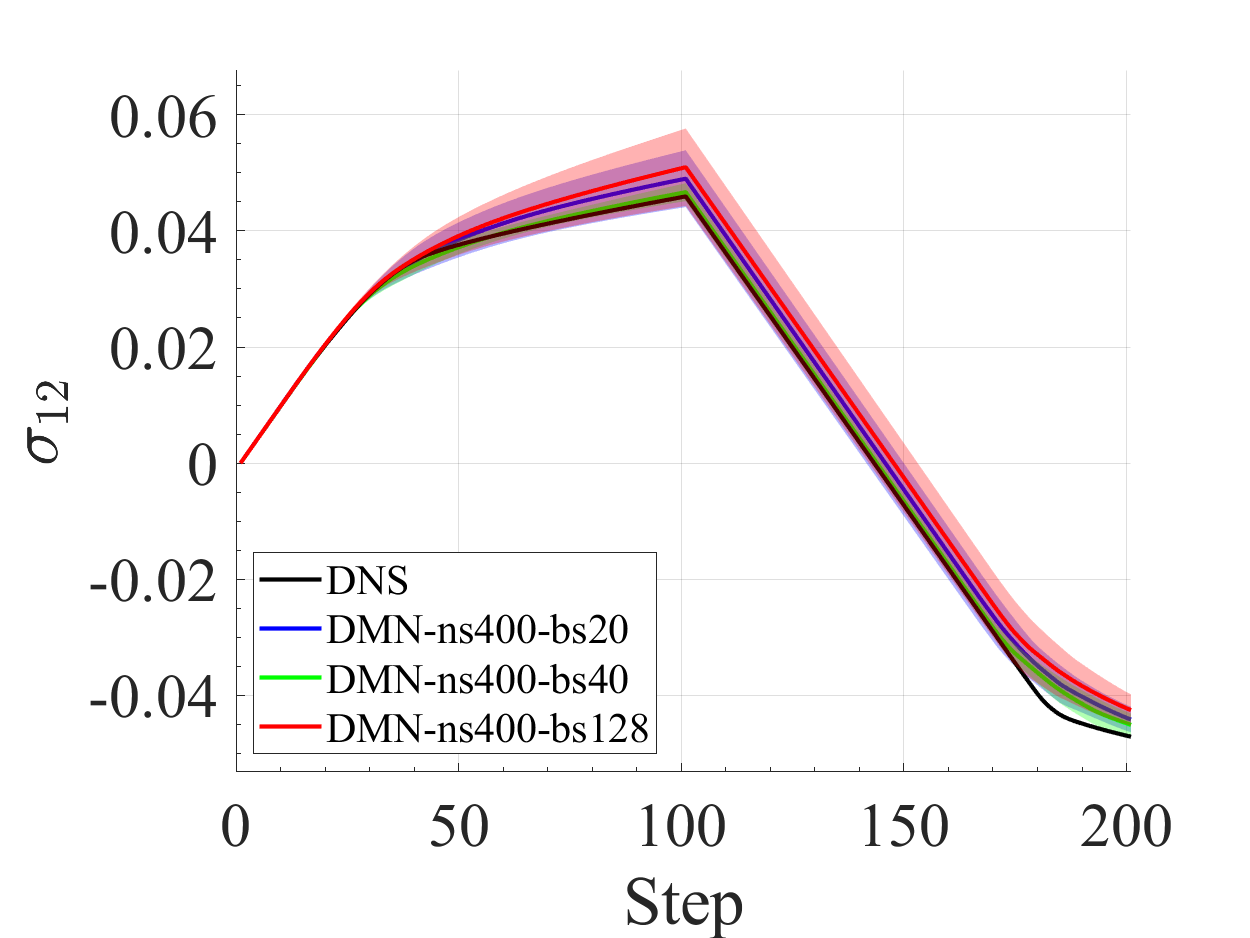}
        \caption{$\sigma_{12}$}
    \end{subfigure}
    \begin{subfigure}{0.325\textwidth}
        \centering
        \includegraphics[width=1\linewidth]{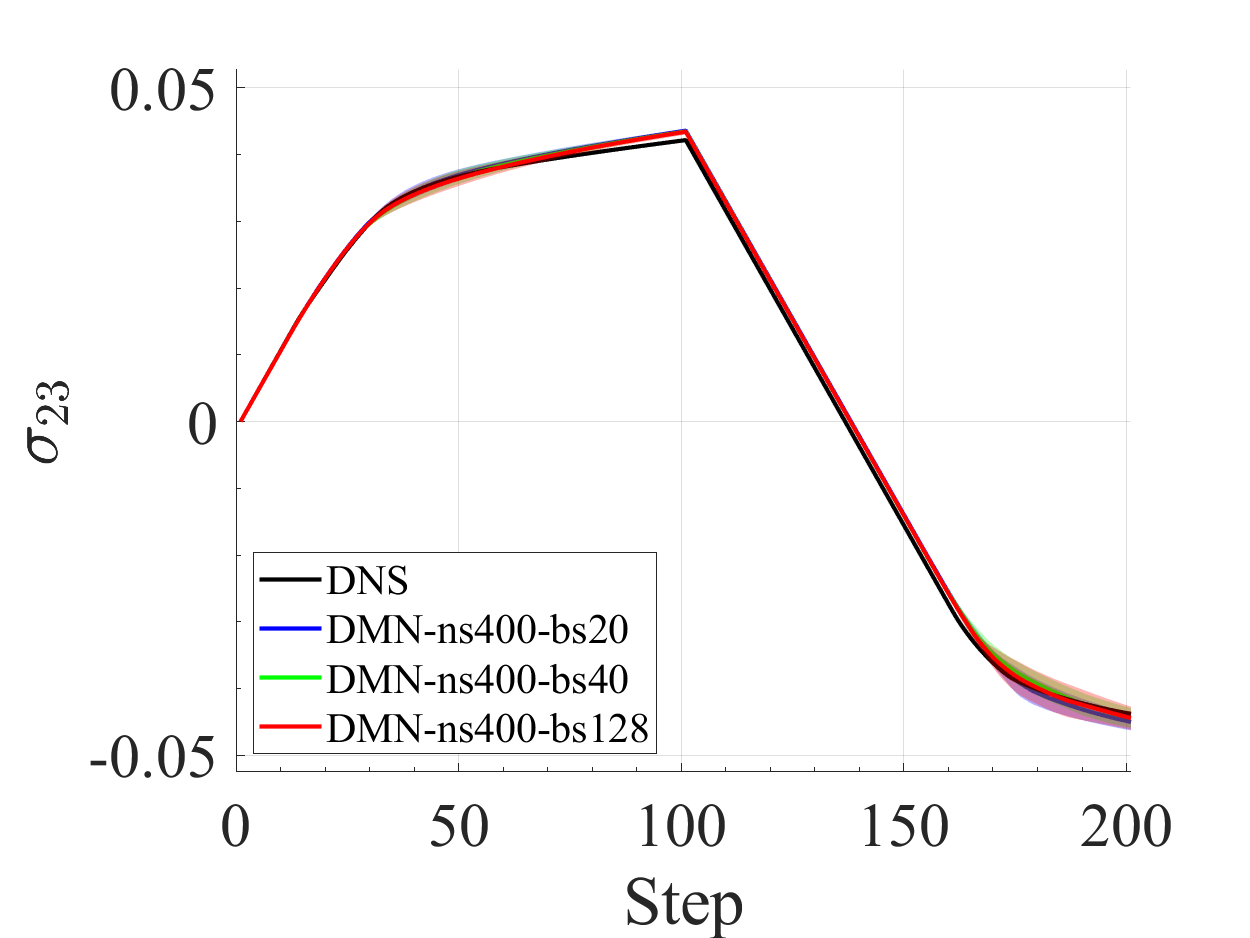}
        \caption{$\sigma_{23}$}
    \end{subfigure}
    \begin{subfigure}{0.325\textwidth}
        \centering
        \includegraphics[width=1\linewidth]{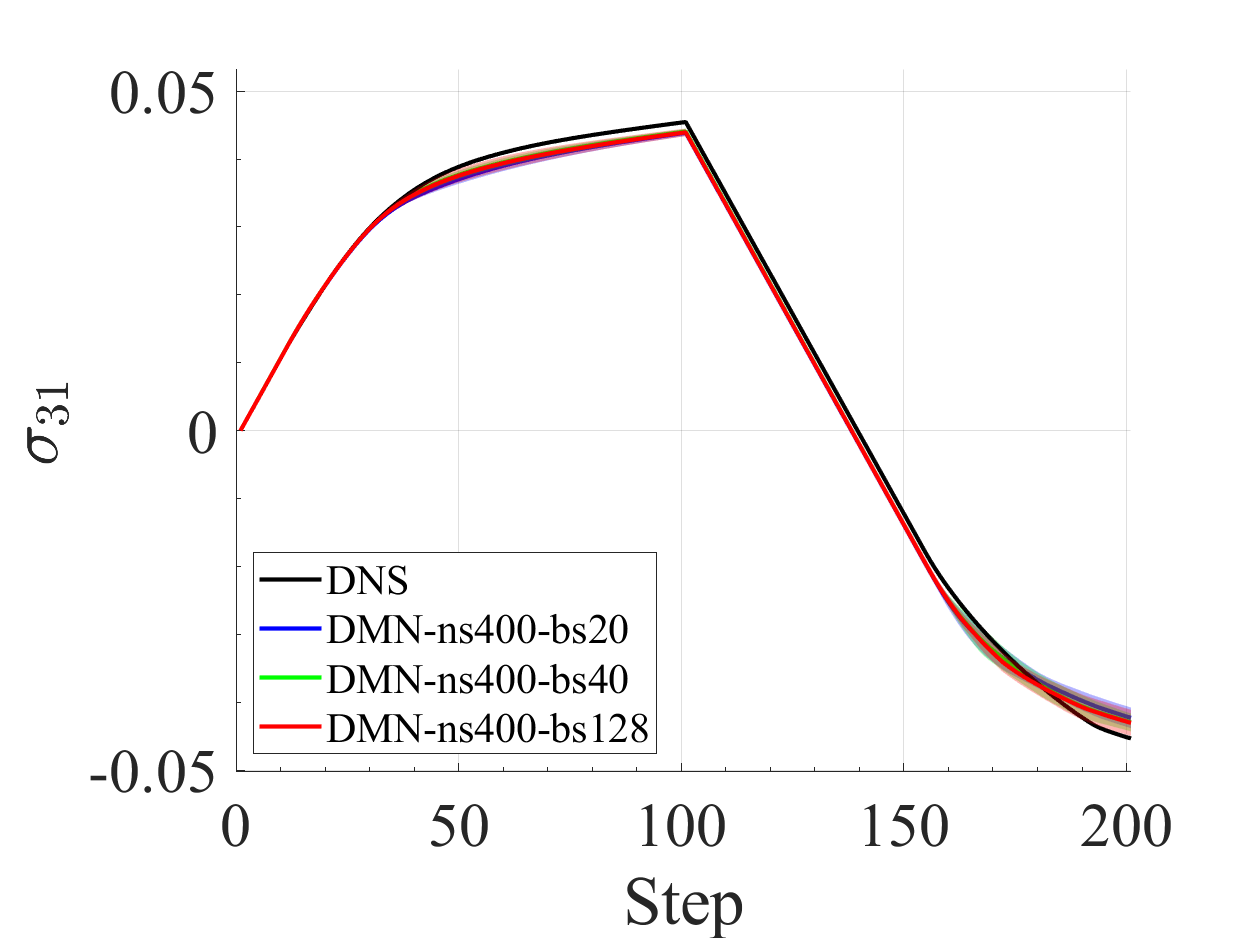}
        \caption{$\sigma_{31}$}
    \end{subfigure}
\caption{Composite 2 - Stress predictions from DMNs trained with 400 samples using batch sizes of 20, 40, and 128: (a) $\sigma_{11}$; (b) $\sigma_{22}$; (c) $\sigma_{33}$; (d) $\sigma_{12}$; (e) $\sigma_{23}$; (f) $\sigma_{31}$; Solid lines show the mean over 10 runs with random initializations, and the shaded regions indicate the corresponding standard deviations.}\label{fig.mat2_ns400}
\end{figure}

\begin{figure}[htp]
\centering
    \begin{subfigure}{0.325\textwidth}
        \centering
        \includegraphics[width=1\linewidth]{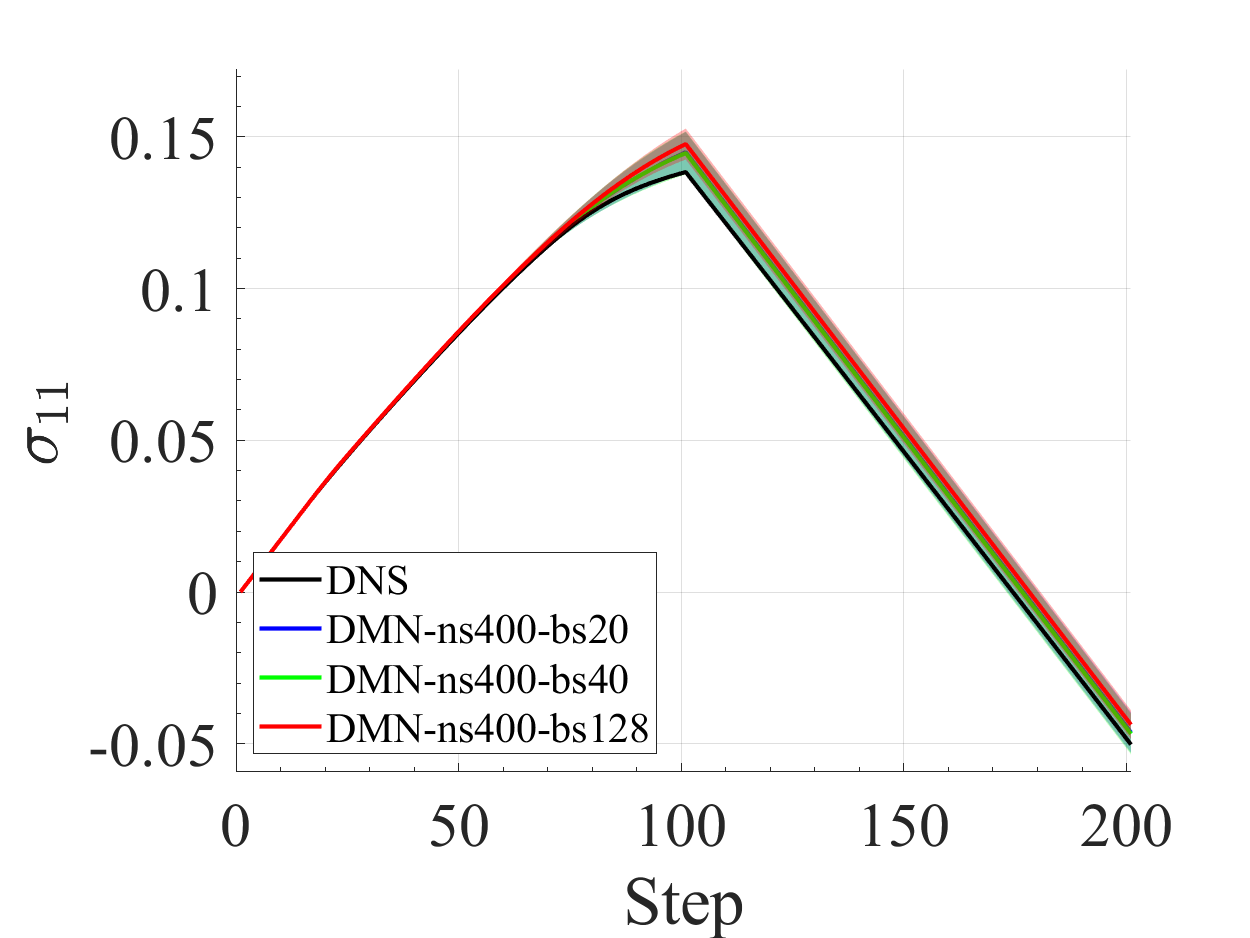}
        \caption{$\sigma_{11}$}
    \end{subfigure}
    \begin{subfigure}{0.325\textwidth}
        \centering
        \includegraphics[width=1\linewidth]{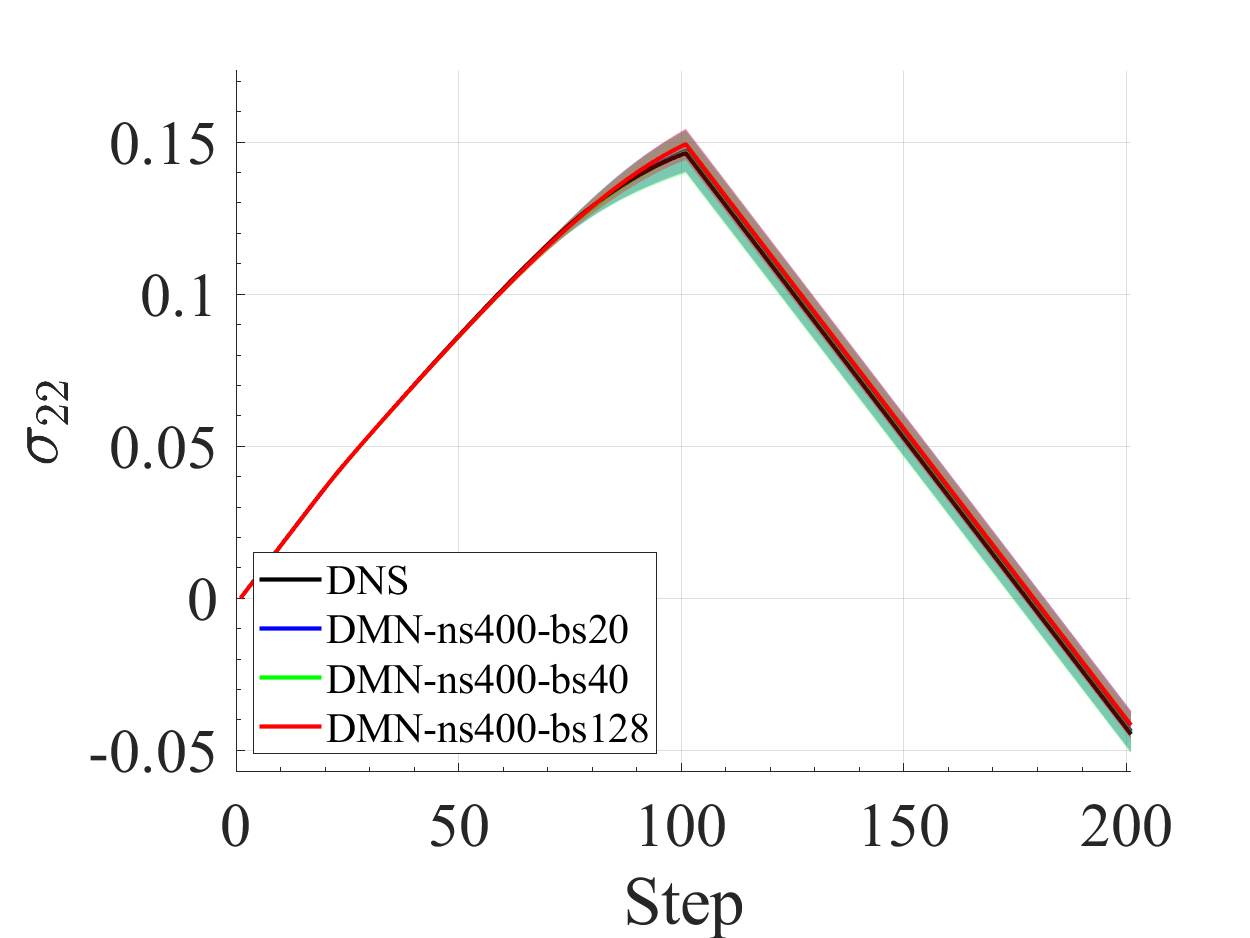}
        \caption{$\sigma_{22}$}
    \end{subfigure}
    \begin{subfigure}{0.325\textwidth}
        \centering
        \includegraphics[width=1\linewidth]{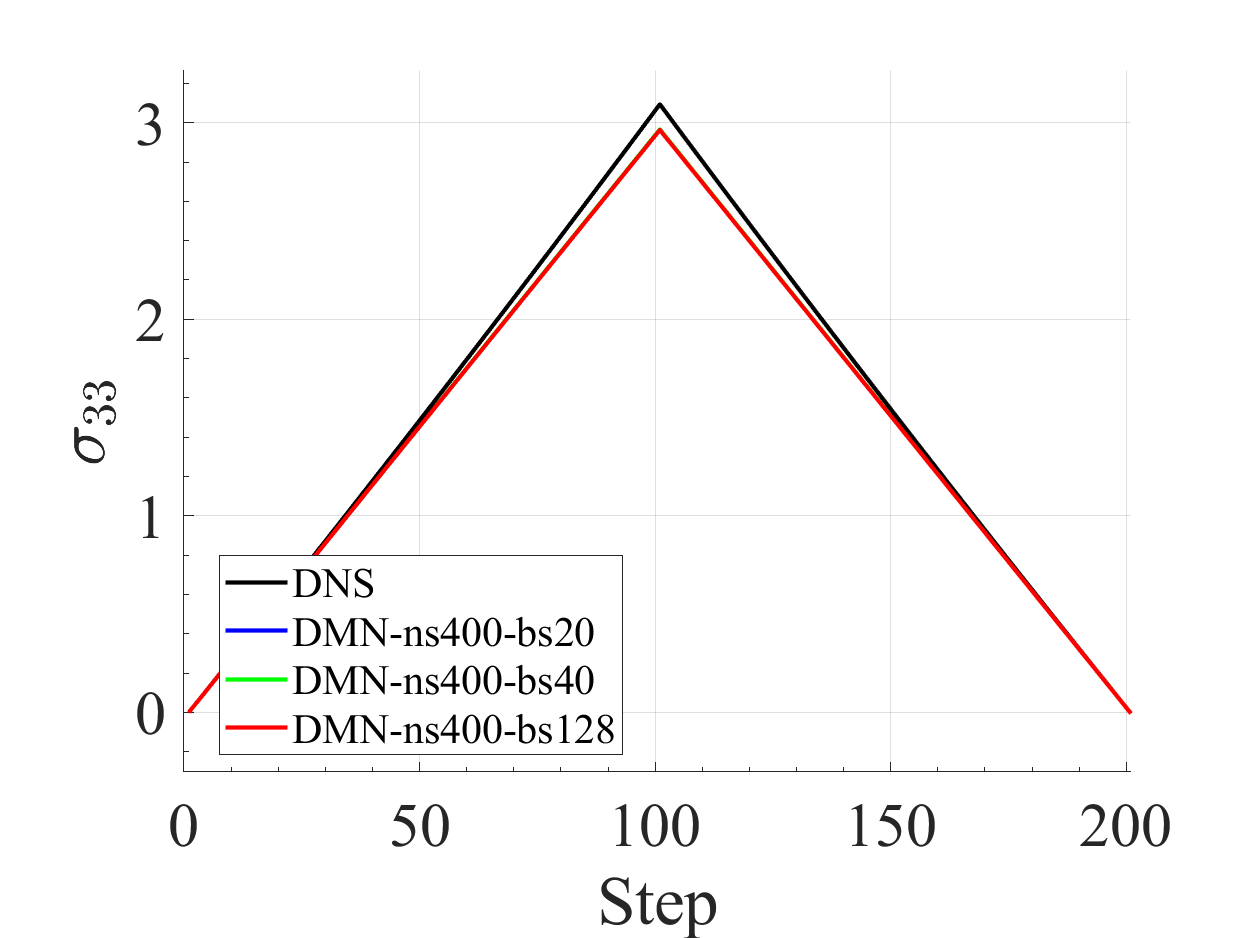}
        \caption{$\sigma_{33}$}
    \end{subfigure}
    \begin{subfigure}{0.325\textwidth}
        \centering
        \includegraphics[width=1\linewidth]{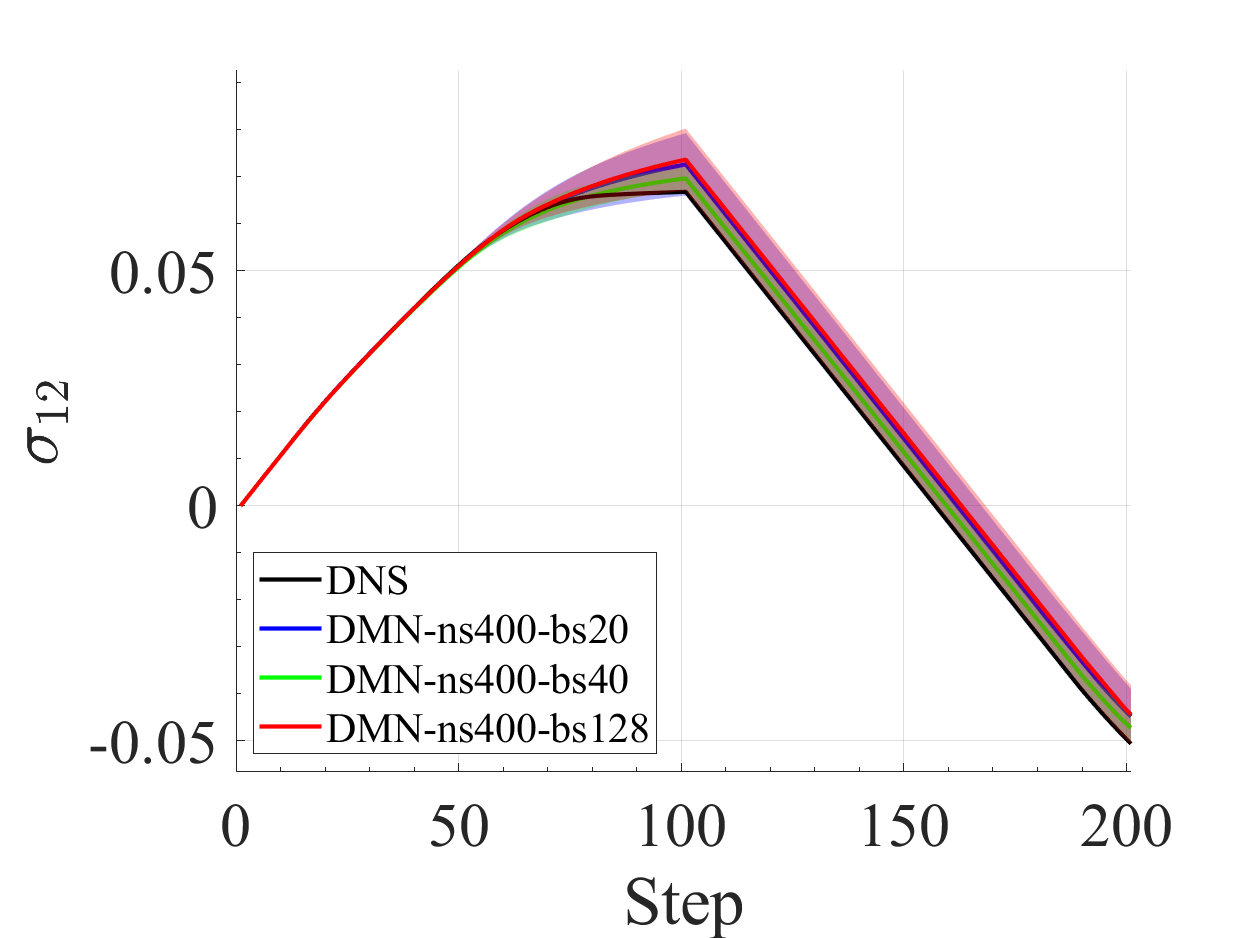}
        \caption{$\sigma_{12}$}
    \end{subfigure}
    \begin{subfigure}{0.325\textwidth}
        \centering
        \includegraphics[width=1\linewidth]{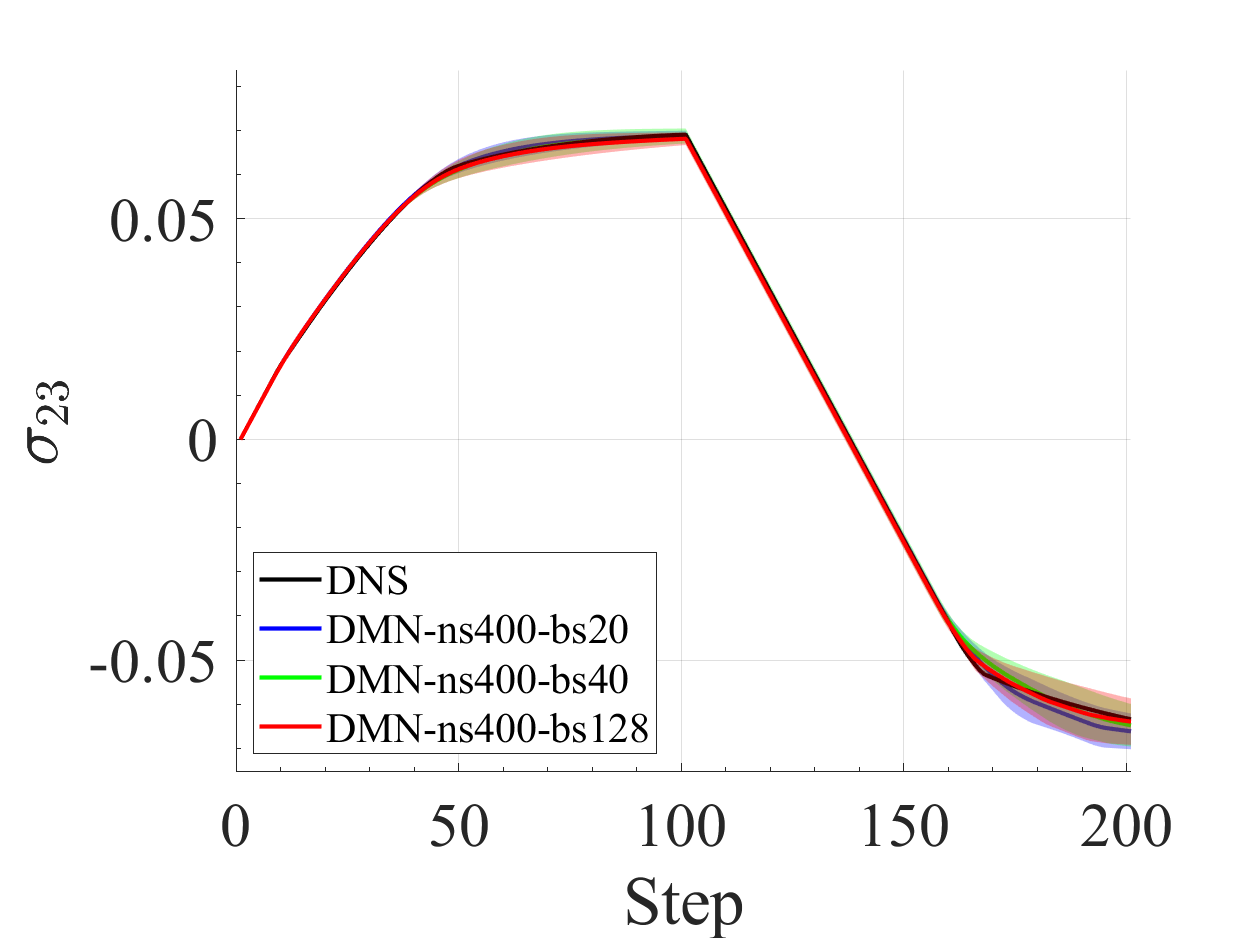}
        \caption{$\sigma_{23}$}
    \end{subfigure}
    \begin{subfigure}{0.325\textwidth}
        \centering
        \includegraphics[width=1\linewidth]{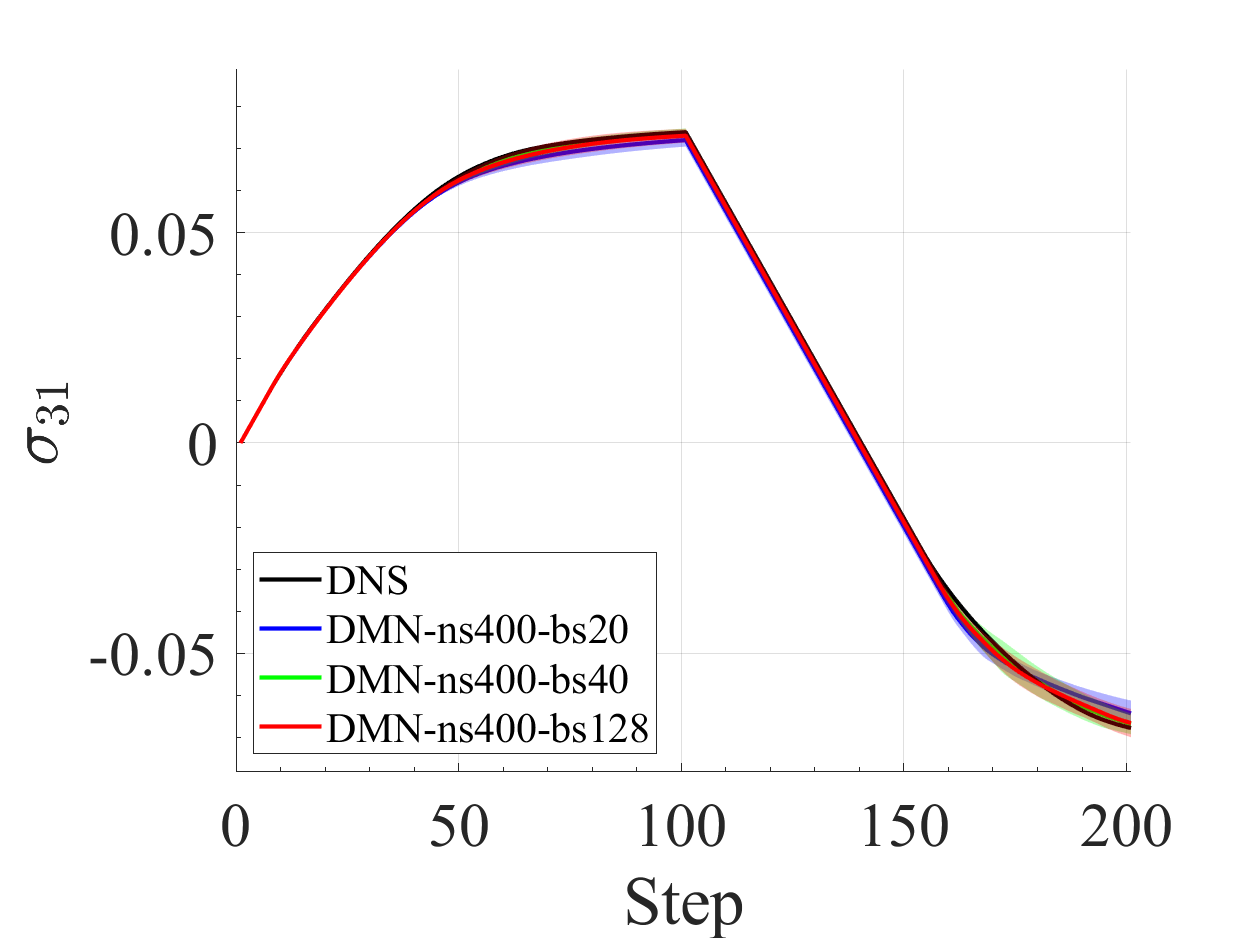}
        \caption{$\sigma_{31}$}
    \end{subfigure}
\caption{Composite 3 - Stress predictions from DMNs trained with 400 samples using batch sizes of 20, 40, and 128: (a) $\sigma_{11}$; (b) $\sigma_{22}$; (c) $\sigma_{33}$; (d) $\sigma_{12}$; (e) $\sigma_{23}$; (f) $\sigma_{31}$; Solid lines show the mean over 10 runs with random initializations, and the shaded regions indicate the corresponding standard deviations.}\label{fig.mat3_ns400}
\end{figure}

Figs. \ref{fig.mat1_ns400} - \ref{fig.mat3_ns400} compare the stress predictions of DMNs trained on 400 samples with DNS results for the three testing composites.
The results demonstrate that both initialization and the batch size have a significant impact on prediction accuracy.
The shaded regions quantify the uncertainty in the predictions arising from these effects.
Among the tested batch sizes (20, 40, and 128), a batch size of 40 yields the most accurate predictions with the least uncertainty across all tested composites, particularly for the 11, 22, and 12 directions.
This observation is aligned with the mean relative errors shown in Fig. \ref{fig.dmn_loss_data_size}(a).
The testing results for DMNs trained with 1024 and 2048 samples are provided in Appendix \ref{sec:dmn_ns1024} and \ref{sec:dmn_ns2048}, and their prediction accuracy and uncertainty trends are also consistent with the mean relative errors shown in Fig. \ref{fig.dmn_loss_data_size}.

\begin{figure}[htp]
    \centering
    \includegraphics[width=0.5\textwidth]{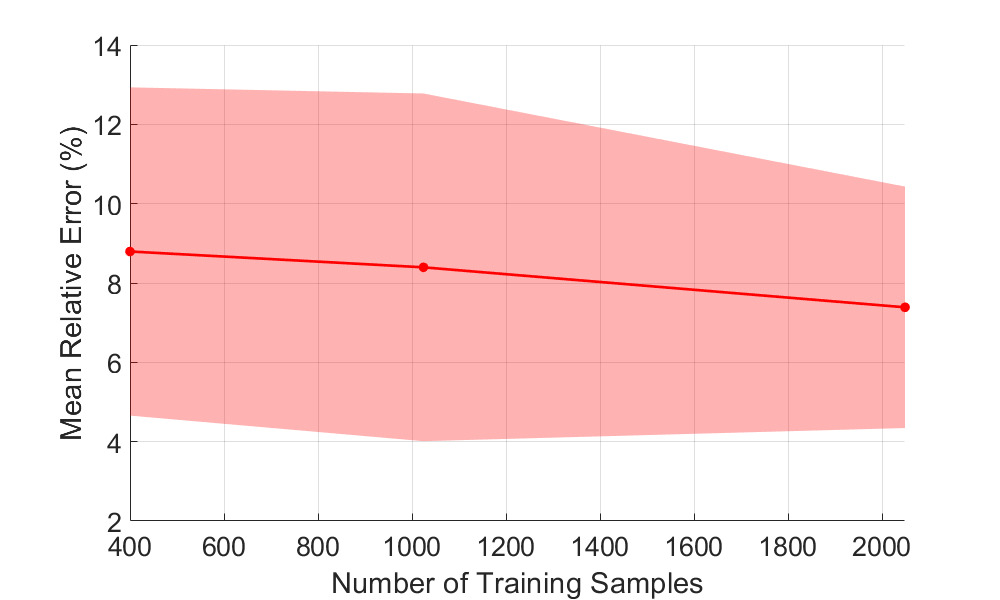}
    \caption{Mean relative errors across all 80 DMN models, covering all training data sizes, batch sizes, initialization runs, and testing composites. Solid line shows the mean and the shaded regions indicate the corresponding standard deviation.}
    \label{fig.error_vs_data}
\end{figure}

Finally, Fig. \ref{fig.error_vs_data} summarizes the mean relative errors across all 80 DMN models, covering all training data sizes, batch sizes, initialization runs, and testing composites.
Overall, the results indicate that increasing the training data size leads to improved prediction accuracy and reduced uncertainty.

\subsection{Effects of Regularization}\label{sec:regularization_effect}
In this section, we examine how the regularization term in the loss function defined in Eq. \eqref{eq.dmn_loss} affects online prediction performance.
IMN models are first trained with $\xi=1$ while varying $\eta$ from $10^{-2}$ to $10^{2}$.
To account for variability due to initialization, each $(\xi,\eta)$ configuration is trained with 5 random initializations. 

\begin{figure}[htp]
	\centering
	\begin{subfigure}{0.495\textwidth}
		\centering
		\includegraphics[width=1\linewidth]{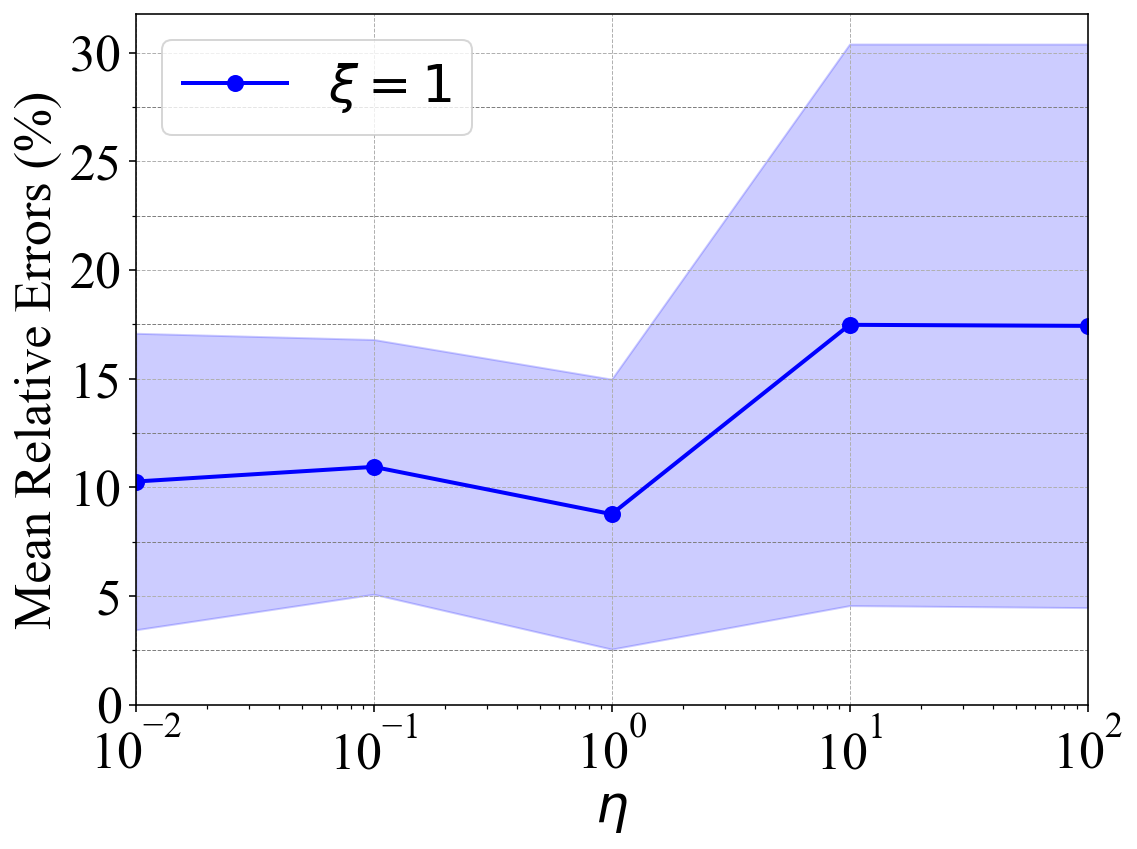}
		\caption{}
	\end{subfigure}
	\begin{subfigure}{0.495\textwidth}
		\centering
		\includegraphics[width=1\linewidth]{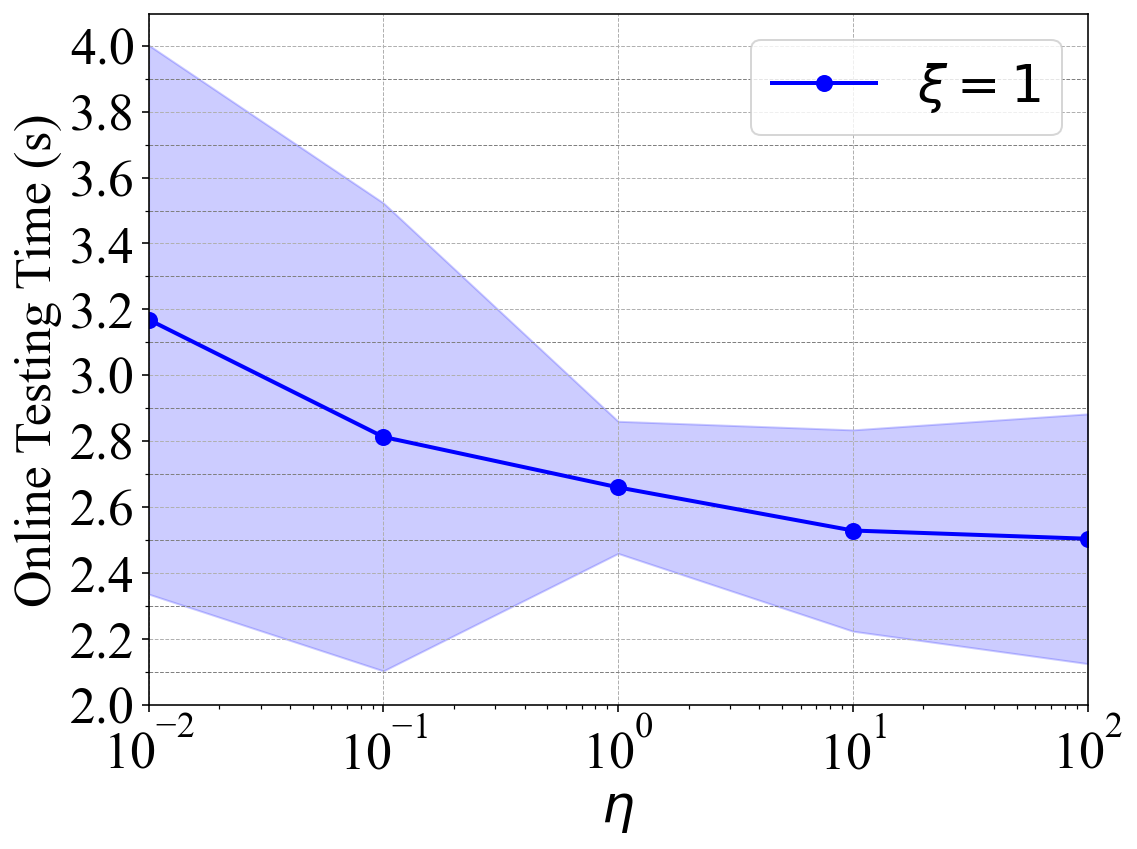}
		\caption{}
	\end{subfigure}
	\begin{subfigure}{0.495\textwidth}
		\centering
		\includegraphics[width=1\linewidth]{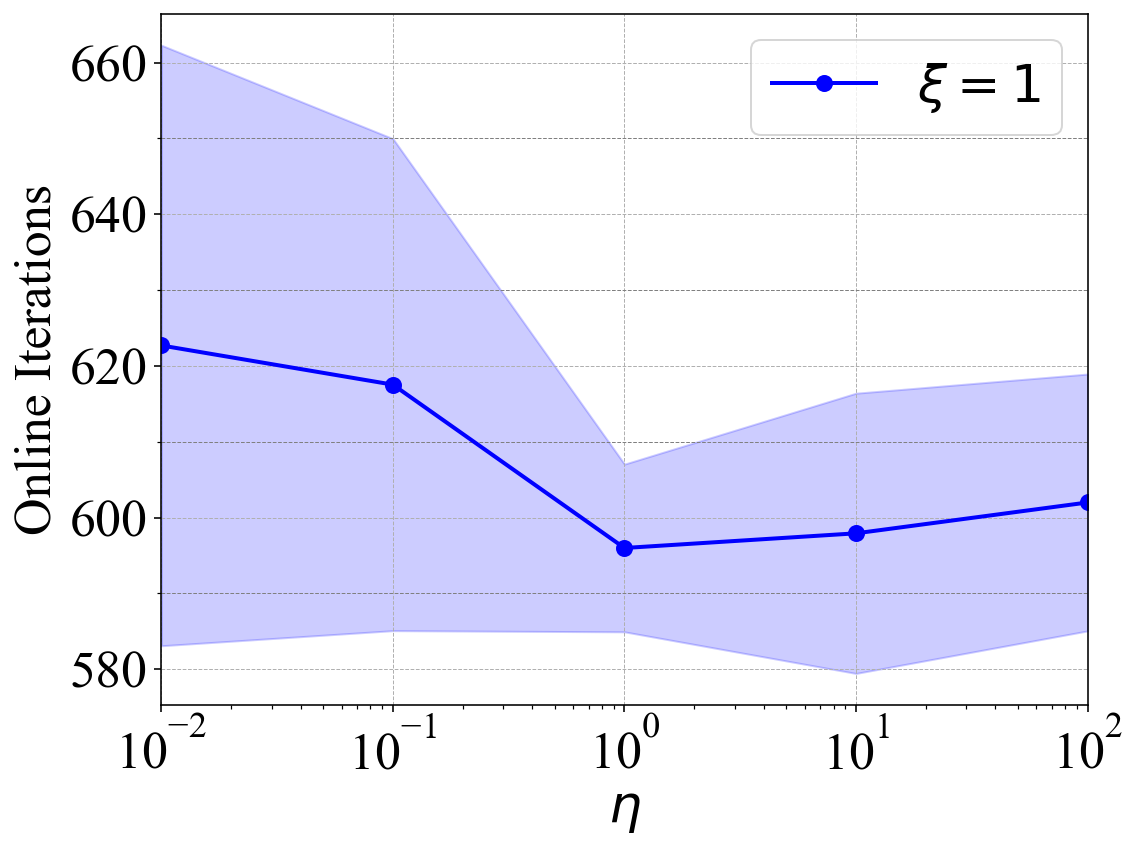}
		\caption{}
	\end{subfigure}
	\begin{subfigure}{0.495\textwidth}
		\centering
		\includegraphics[width=1\linewidth]{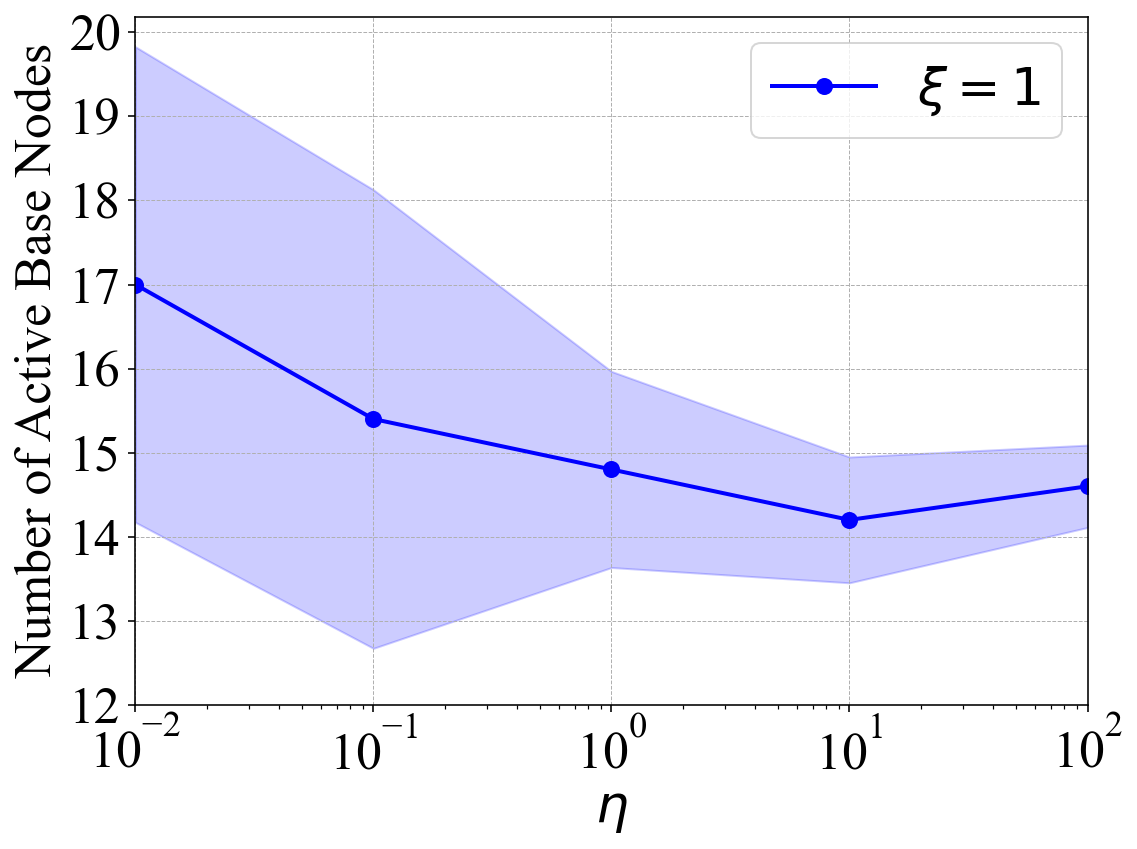}
		\caption{}
	\end{subfigure}
	\caption{Effects of $\eta$ in the loss function (Eq. \eqref{eq.dmn_loss}) on IMN online prediction performance: (a) mean relative errors; (b) online testing time; (c) number of iterations required for convergence; (d) number of active base nodes; Solid line shows the mean over 5 runs with random initializations, and the shaded regions indicate the corresponding standard deviation.}\label{fig.imn_effect_eta}
\end{figure}

Fig. \ref{fig.imn_effect_eta}(a) shows that smaller values of $\eta$ generally yield lower prediction errors, as measured by the metric defined in Eq. \eqref{eq.metric_stress}, with the minimum prediction error occurring at $\eta=1$.
As shown in Fig. \ref{fig.imn_effect_eta}(c)-(d), both the number of iterations required for convergence and the number of active base nodes decrease slightly as $\eta$ increases.
As a result, the online testing time, shown in Fig. \ref{fig.imn_effect_eta}(b), exhibits a modest reduction with increasing $\eta$.
These results suggest that $\eta$ has a relatively limited effect on the number of active base nodes, which reflects the network's complexity and, consequently, its online computational efficiency. 
However, excessively large values of $\eta$ lead to an imbalanced contribution of the regularization term to the total loss, resulting in degraded model performance (Fig. \ref{fig.imn_effect_eta}(a)).
Overall, $\eta=1$ provides the best trade-off between online prediction accuracy and efficiency.

\begin{figure}[htp]
	\centering
	\begin{subfigure}{0.495\textwidth}
		\centering
		\includegraphics[width=1\linewidth]{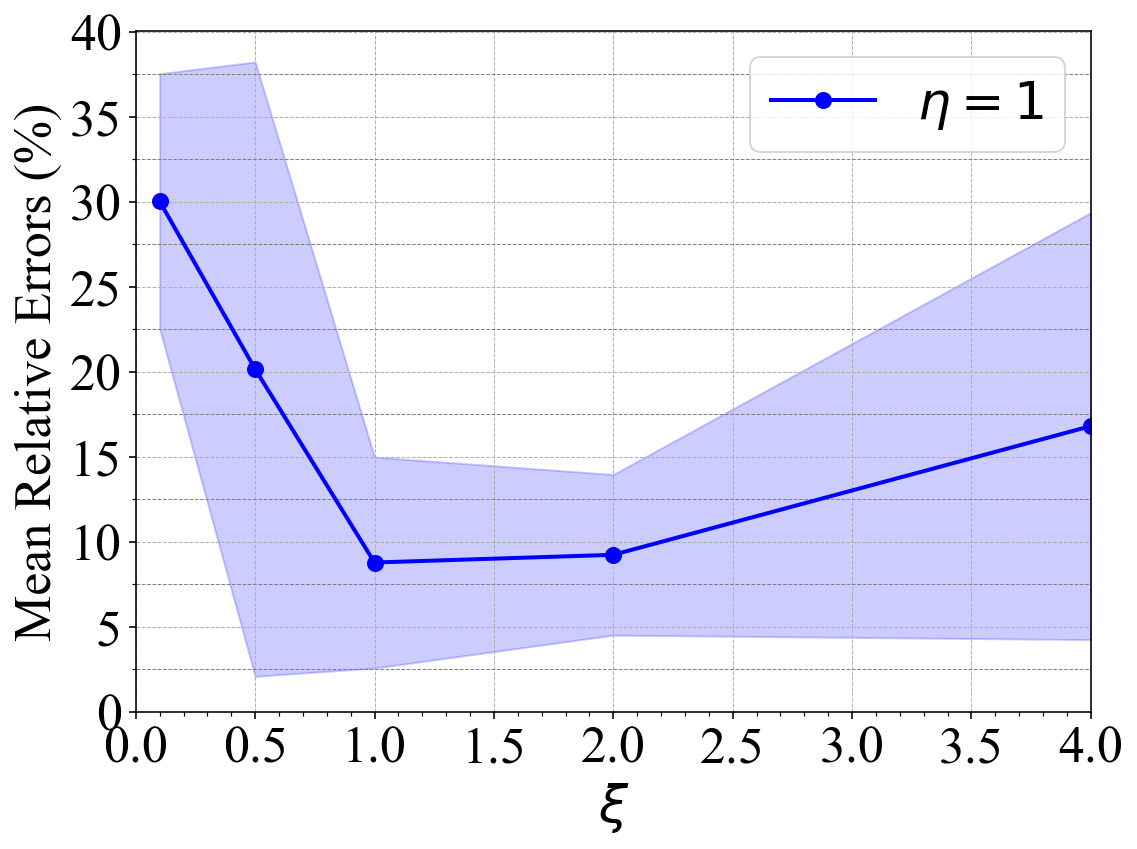}
		\caption{}
	\end{subfigure}
	\begin{subfigure}{0.495\textwidth}
		\centering
		\includegraphics[width=1\linewidth]{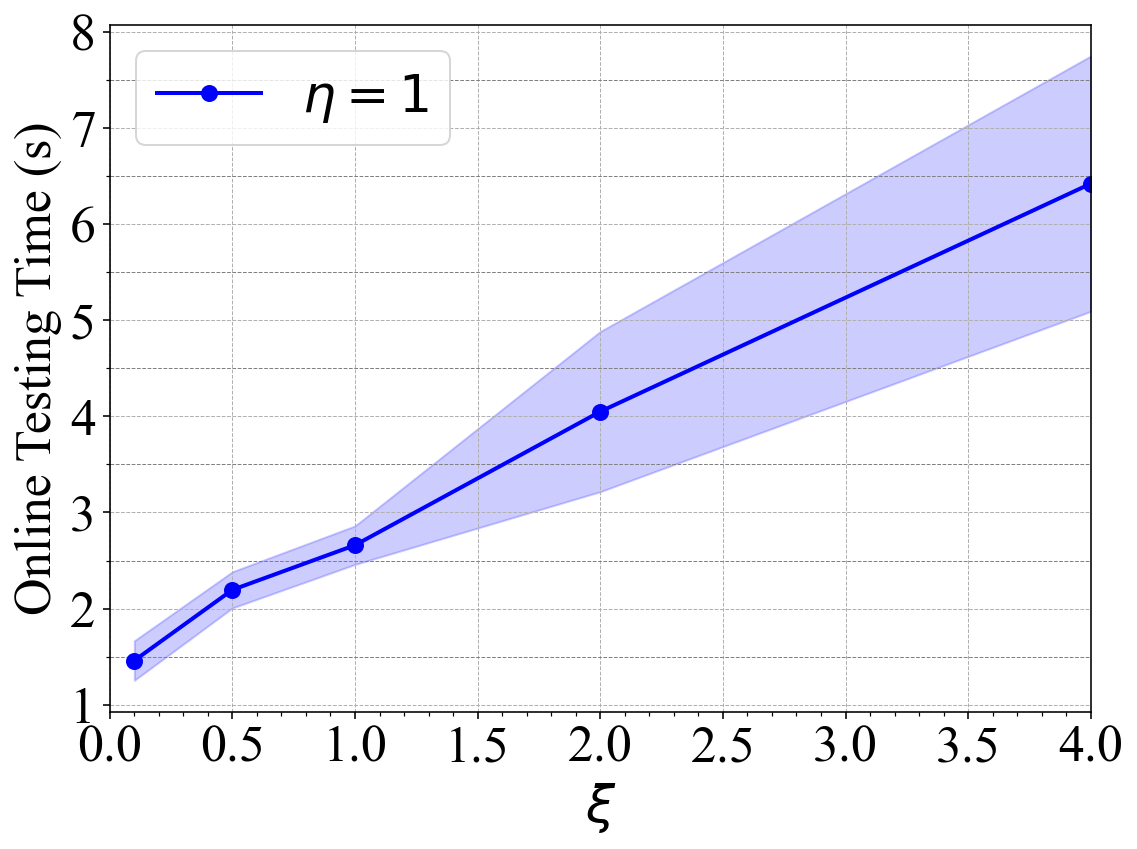}
		\caption{}
	\end{subfigure}
	\begin{subfigure}{0.495\textwidth}
		\centering
		\includegraphics[width=1\linewidth]{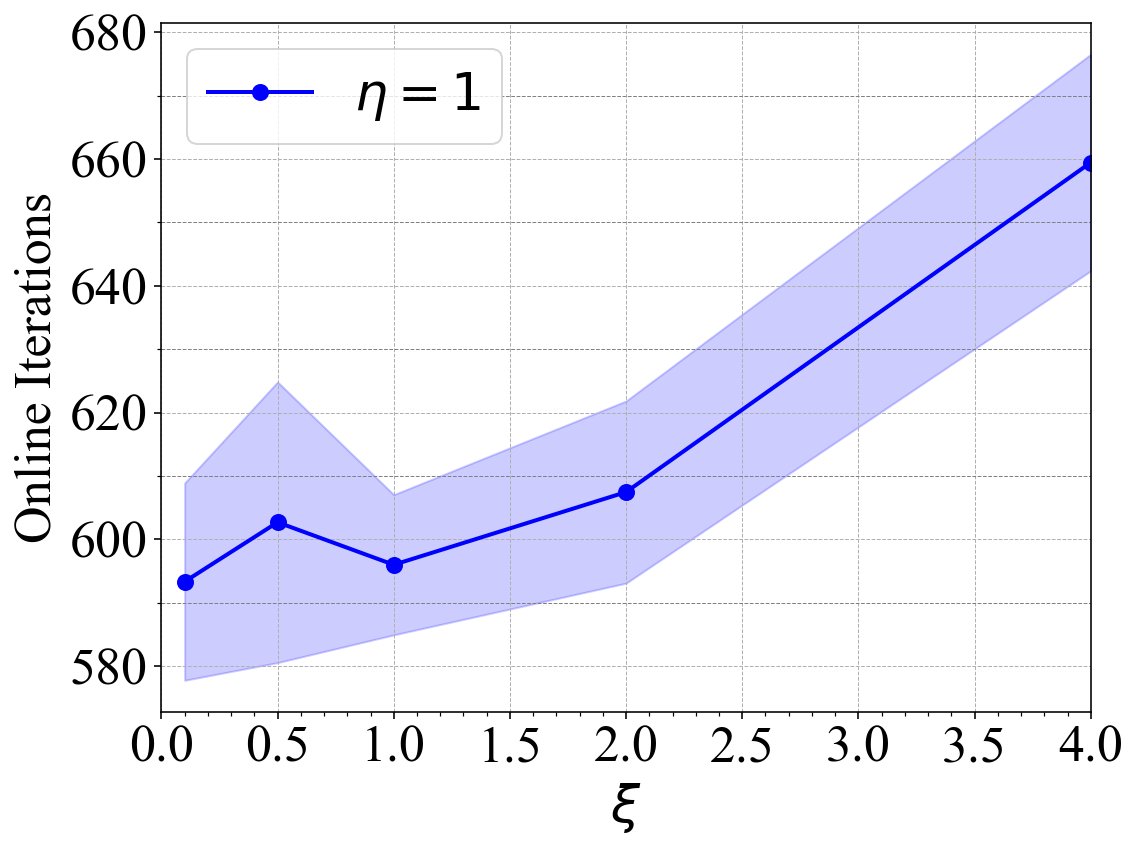}
		\caption{}
	\end{subfigure}
	\begin{subfigure}{0.495\textwidth}
		\centering
		\includegraphics[width=1\linewidth]{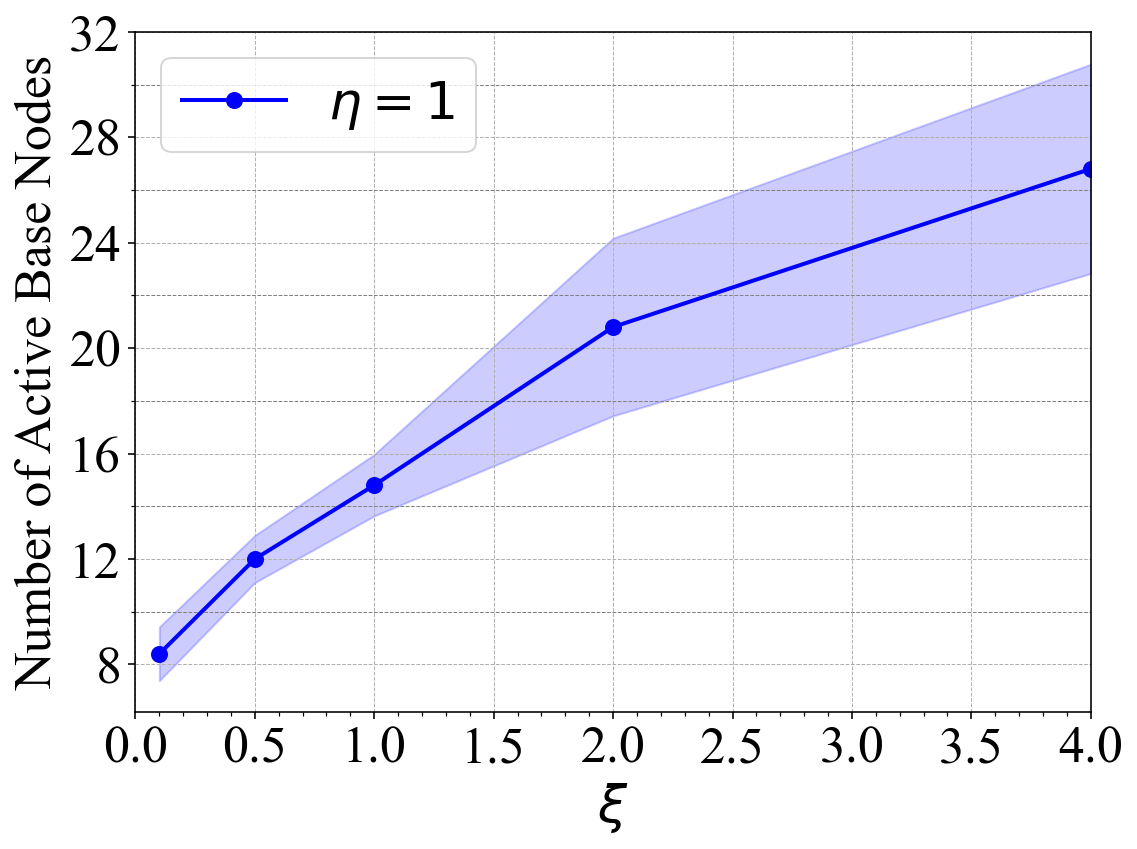}
		\caption{}
	\end{subfigure}
	\caption{Effects of $\xi$ in the loss function (Eq. \eqref{eq.dmn_loss}) on IMN online prediction performance: (a) mean relative errors; (b) online testing time; (c) number of iterations required for convergence; (d) number of active base nodes; Solid line shows the mean over 5 runs with random initializations, and the shaded regions indicate the corresponding standard deviation.}\label{fig.imn_effect_xi}
\end{figure}

Next, we investigate the effect of $\xi$ by training IMN models with $\eta=1$ while varying $\xi$ from 0.1 to 4.
Fig. \ref{fig.imn_effect_xi}(a) shows that the prediction error initially decreases and then increases as $\xi$ grows, reaching a minimum at $\xi=1$.
As shown in Fig. \ref{fig.imn_effect_xi}(b), the online testing time increases with $\xi$, which can be attributed to both the increasing number of active base nodes (Fig. \ref{fig.imn_effect_xi}(d)) and more online iterations required for convergence (Fig. \ref{fig.imn_effect_xi}(c)).

These observations indicate that larger values of $\xi$ promote more active base nodes, resulting in increased network complexity and higher online computational costs. 
Conversely, when $\xi$ is too small, the network contains too few active base nodes to adequately capture essential microstructural interactions and homogenization behavior, leading to degraded prediction accuracy.
Increasing $\xi$ enhances network expressiveness and improves accuracy up to an optimal point.
Beyond this point, however, excessive network complexity may lead to deteriorated performance. 
Based on these parametric studies, $\eta=1$ and $\xi=1$ are selected for the loss function in all subsequent DMN and IMN experiments.

\subsection{DMN vs IMN - Offline Training}\label{sec:dmn_vs_imn_offline}
\begin{figure}[htp]
\centering
    \begin{subfigure}{0.325\textwidth}
        \centering
        \includegraphics[width=1\linewidth]{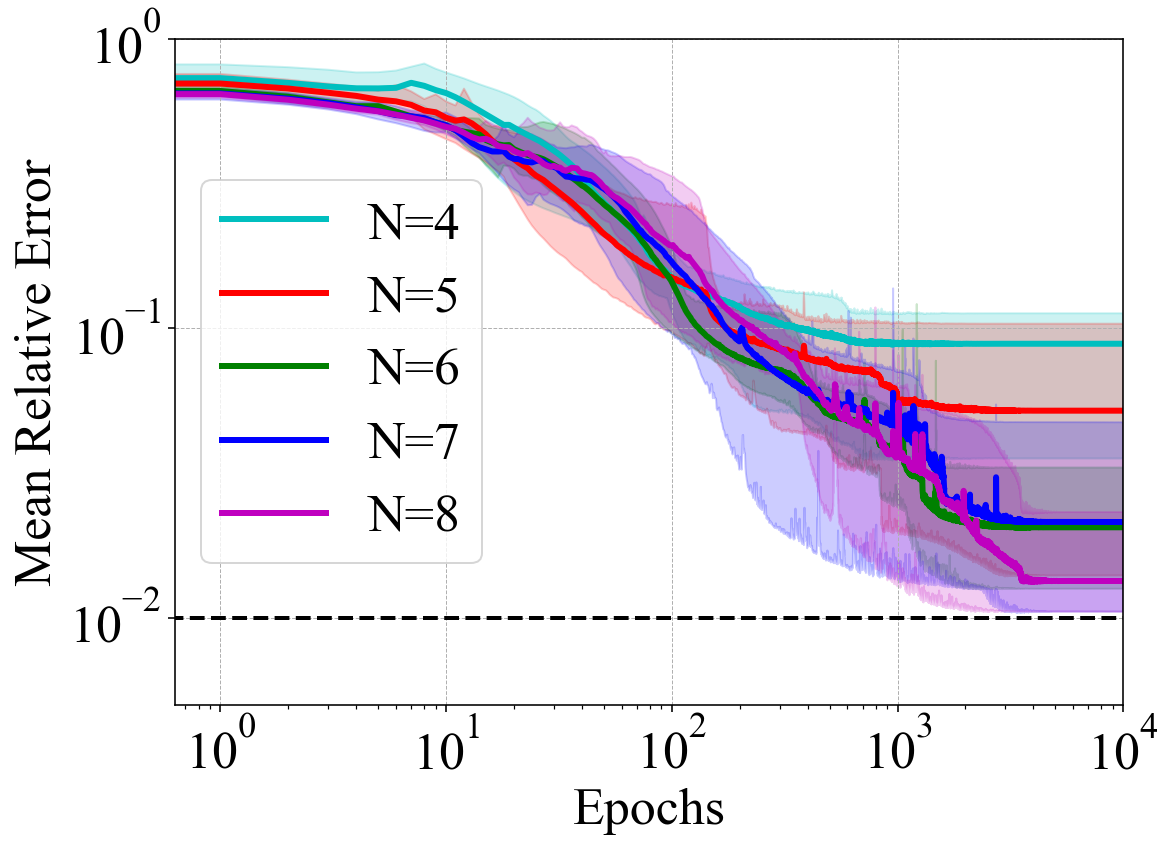}
        \caption{}
    \end{subfigure}
    \begin{subfigure}{0.325\textwidth}
        \centering
        \includegraphics[width=1\linewidth]{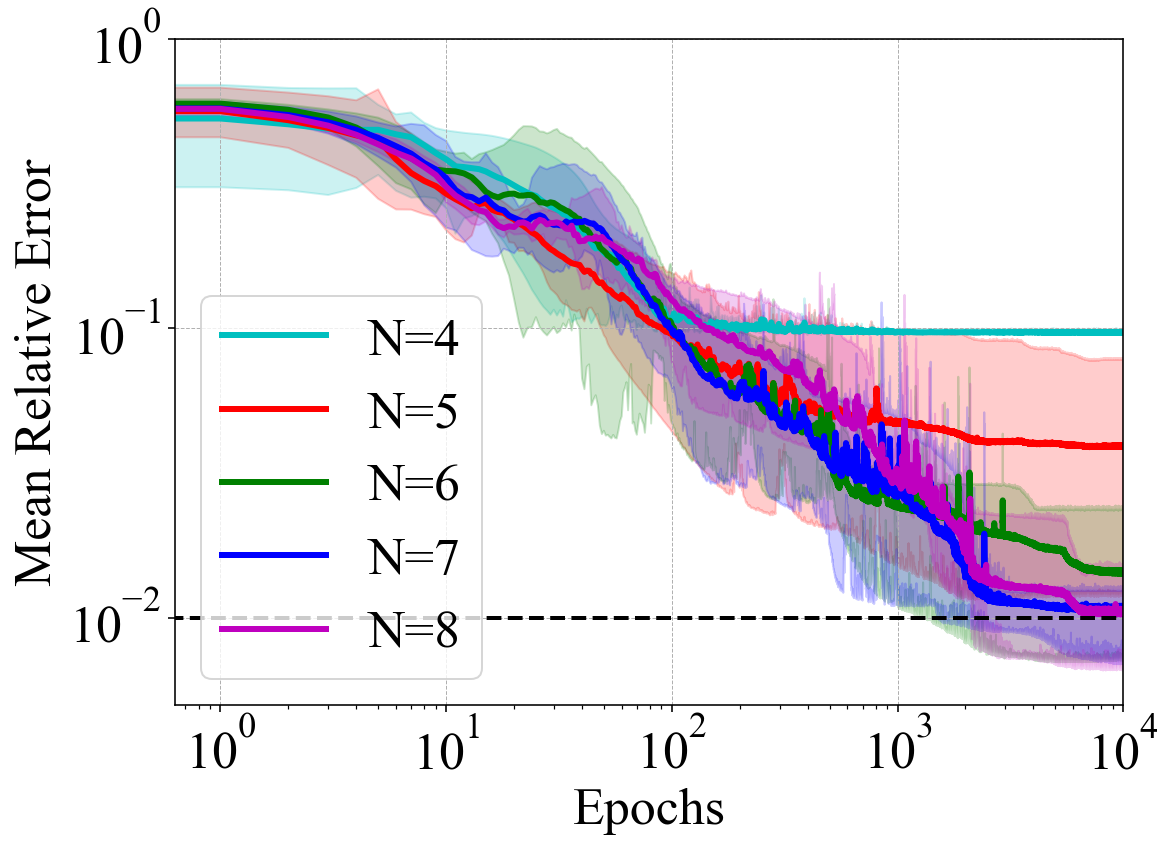}
        \caption{}
    \end{subfigure}
        \begin{subfigure}{0.325\textwidth}
        \centering
        \includegraphics[width=1\linewidth]{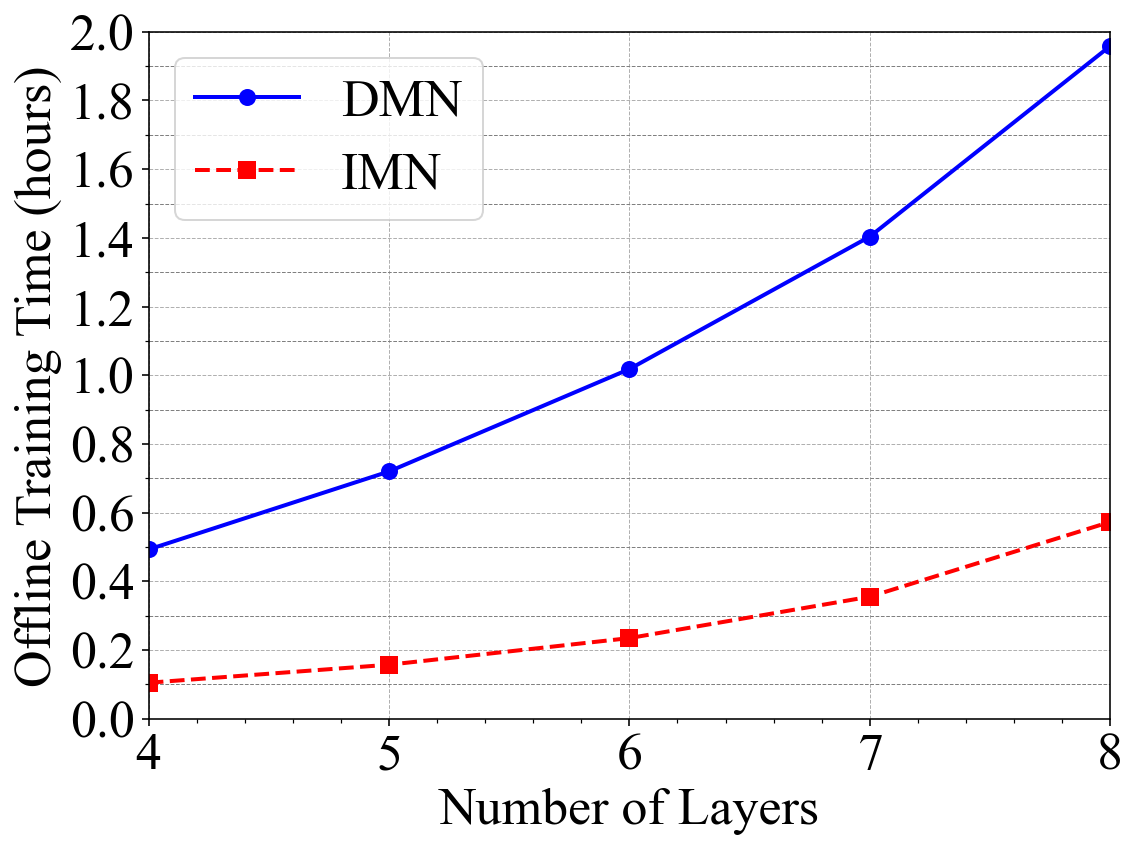}
        \caption{}
    \end{subfigure}
\caption{(a) Training histories of mean relative errors of DMNs trained using network depths ranging from 4 to 8; (b) histories of training mean relative errors of IMNs trained with various numbers of layers; (c) Offline training time of DMN and IMN; Solid line shows the mean over 5 runs with random initializations, and the shaded regions indicate the corresponding standard deviation.}
\label{fig.dmn_vs_imn_offline}
\end{figure}

In this section, we compare the offline training costs of DMN and IMN. 
Both models are trained on Dataset 1 (400 samples) using network depths ranging from 4 to 8 layers for 10,000 epochs.
To account for variability due to initialization, each configuration is trained with 5 random initializations.
Fig. \ref{fig.dmn_vs_imn_offline}(a) and (b) show training histories of the mean relative errors for DMN and IMN, respectively.
For both models, deeper networks generally converge to lower training errors.
It is observed that IMN achieves lower errors than DMN for network depths $N \ge 6$.

Fig. \ref{fig.dmn_vs_imn_offline}(c) compares the offline training times of DMN and IMN.
Compared with DMN, IMN achieves a $3.4 \times$ - $4.7 \times$ speed-up in offline training.
A primary contributor to this acceleration is the reduced number of trainable parameters in IMN for a given network depth, as discussed in Section \ref{sec:imn}.
Furthermore, IMN exhibits lower computational complexity in its forward pass, as will be demonstrated in Section \ref{sec:dmn_vs_imn_online}.
The improved offline training efficiency not only facilitates faster fine-tuning from pre-trained models in practical applications, but also enables the use of larger and more diverse training datasets, leading to enhanced robustness and accuracy, as demonstrated in Section \ref{sec:dmn_ns400}.

\subsection{DMN Online Prediction - Effects of Residual Stress}\label{sec:dmn_residual}
In this section, we investigate the effects of the residual stress term in the DMN online iteration scheme (Eqs. \eqref{eq.dmn_online_stress_strain_relation}-\eqref{eq.dmn_rotated_residual_stress}) on prediction performance, as described in Section \ref{sec:dmn_online}.
Each trained DMN is evaluated using online algorithms with and without the residual stress.
The trained DMN models are applied to predict the material responses of Composite 1 subjected to six loading conditions.

\begin{figure}[htp]
	\centering
	\begin{subfigure}{0.495\textwidth}
		\centering
		\includegraphics[width=1\linewidth]{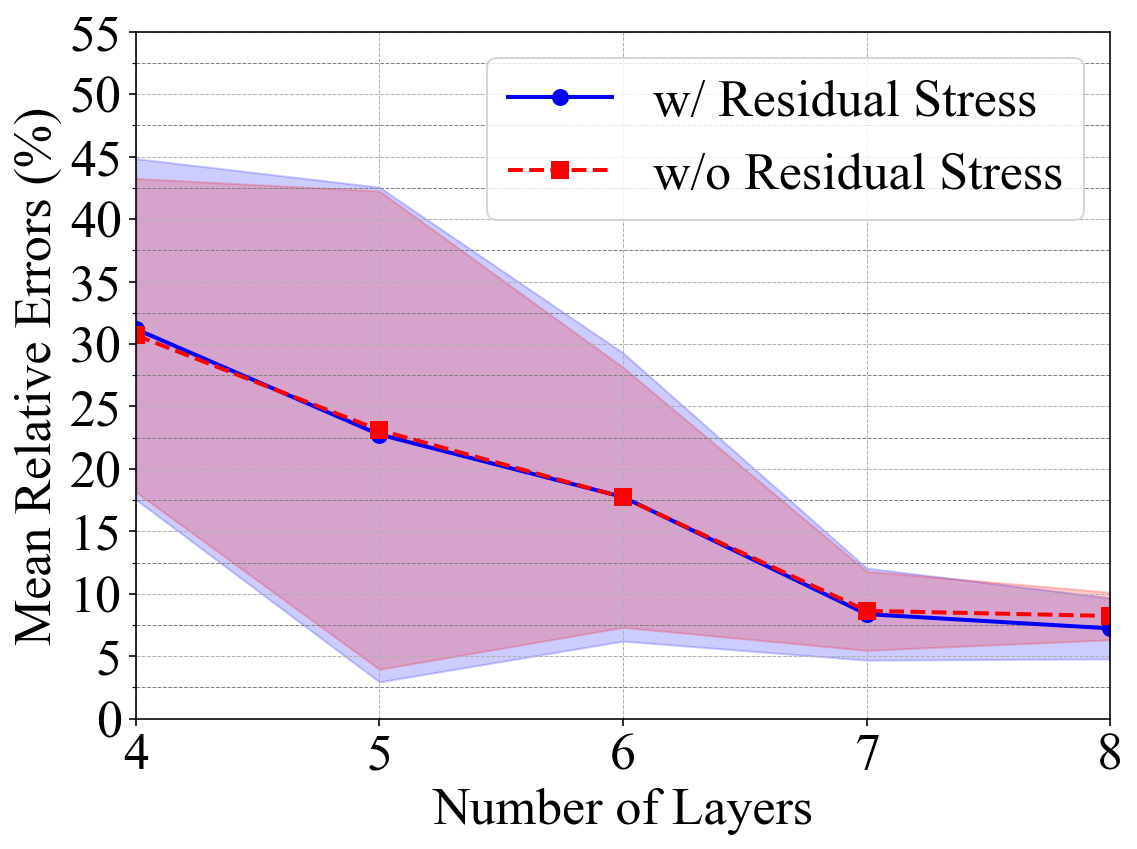}
		\caption{}
	\end{subfigure}
	\begin{subfigure}{0.495\textwidth}
		\centering
		\includegraphics[width=1\linewidth]{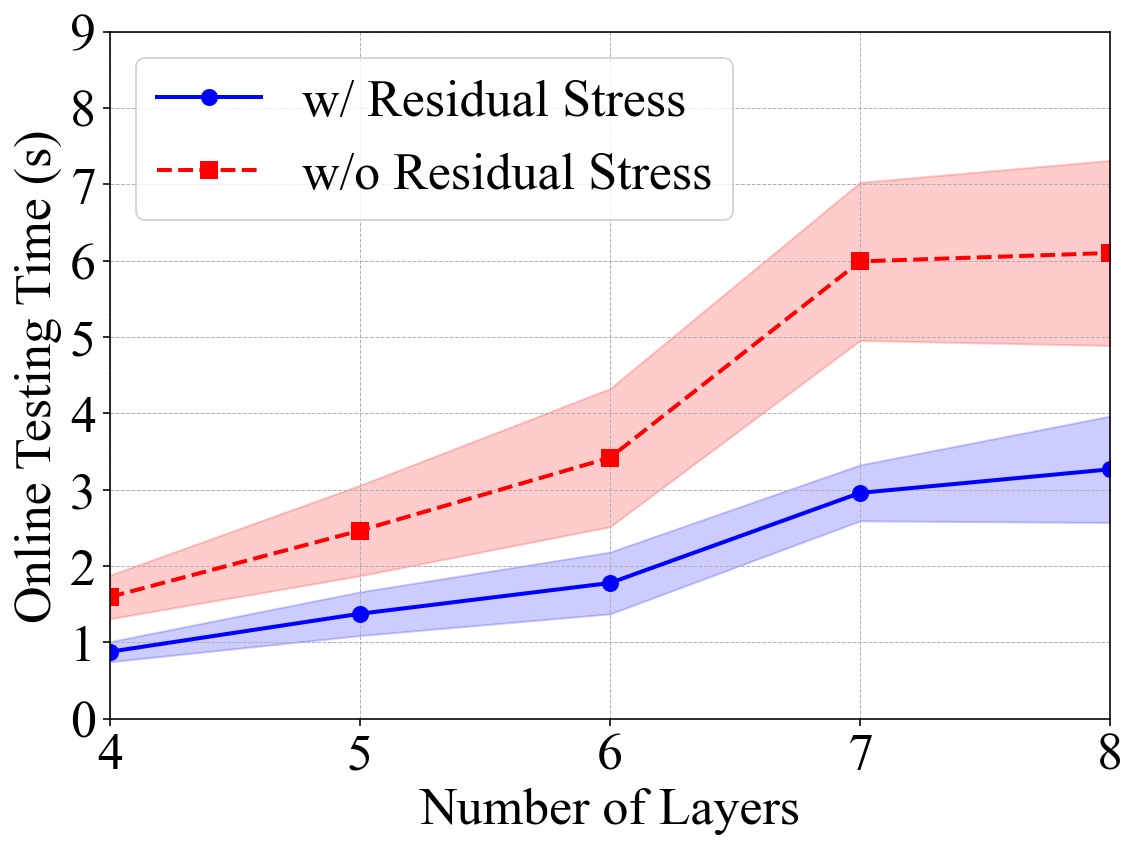}
		\caption{}
	\end{subfigure}
	\begin{subfigure}{0.495\textwidth}
		\centering
		\includegraphics[width=1\linewidth]{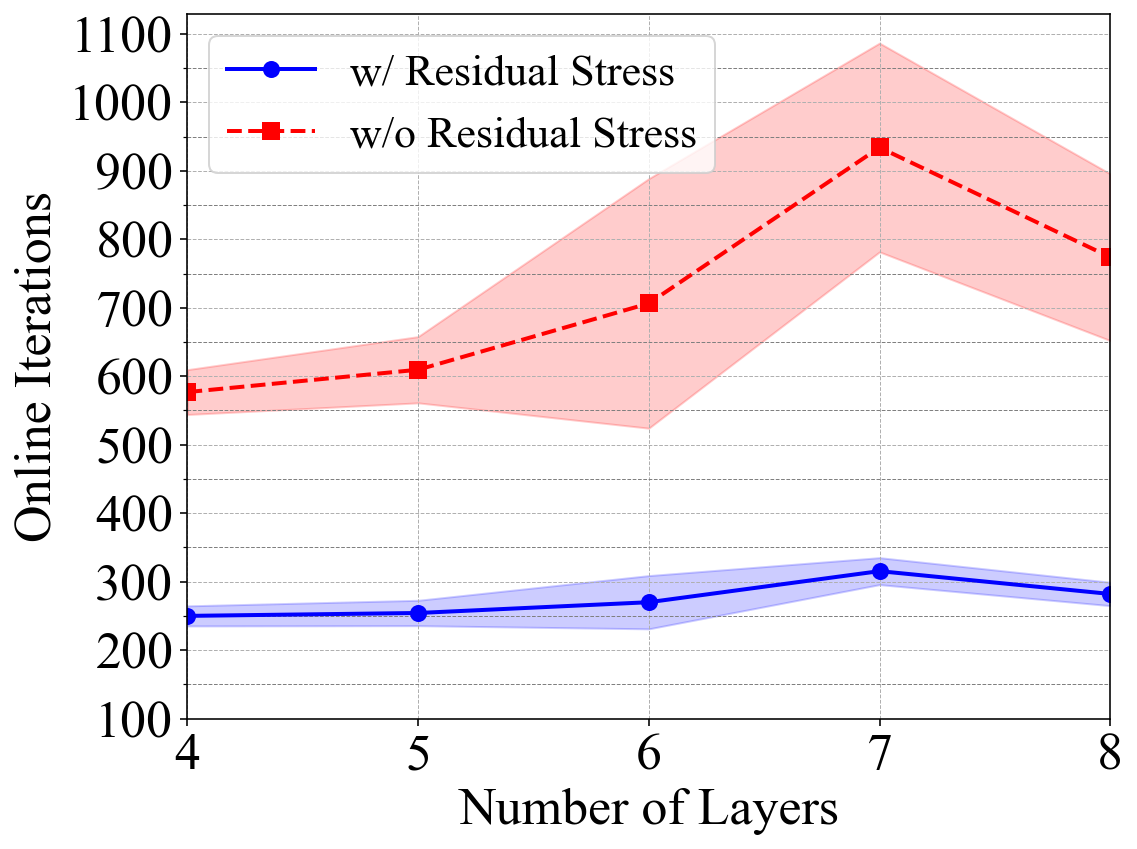}
		\caption{}
	\end{subfigure}
	\begin{subfigure}{0.495\textwidth}
		\centering
		\includegraphics[width=1\linewidth]{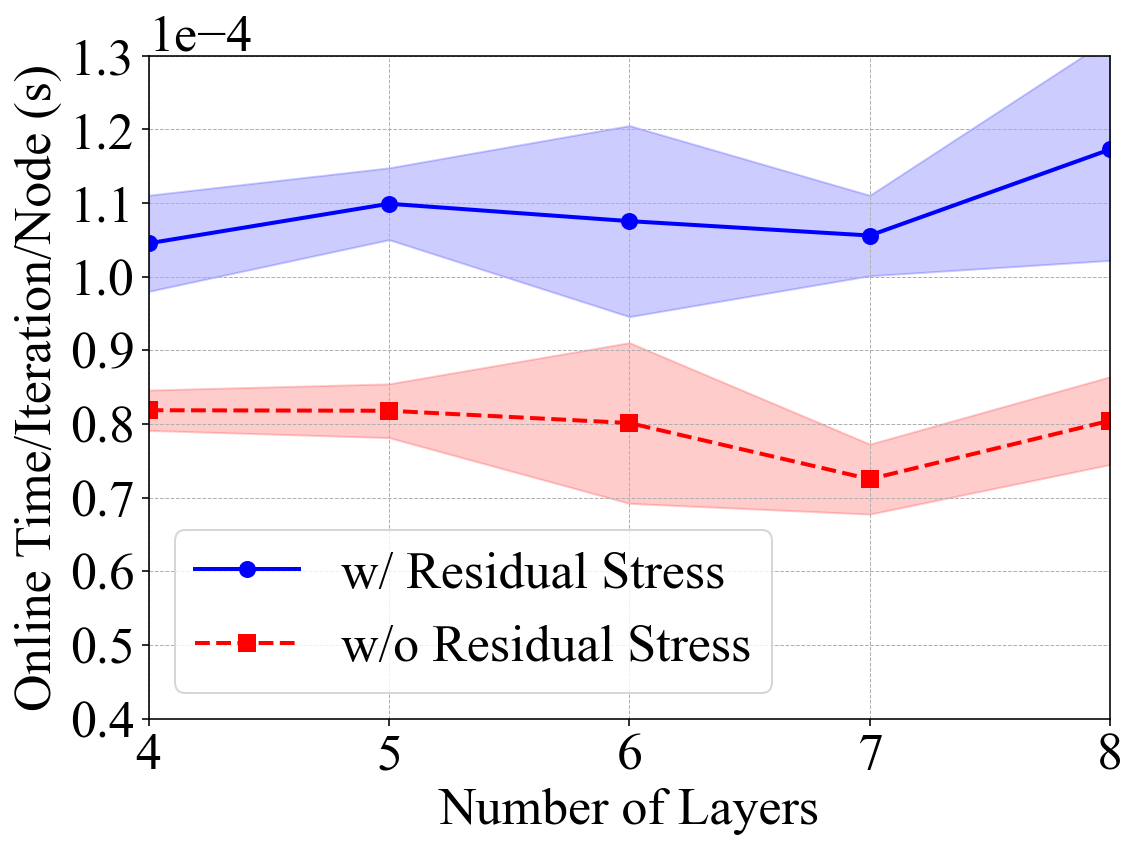}
		\caption{}
	\end{subfigure}
	\caption{Effects of the residual stress on IMN online prediction performance: (a) mean relative errors; (b) online testing time; (c) number of iterations required for convergence; (d) online testing time per iteration per node; Solid line shows the mean over 5 runs with random initializations, and the shaded regions indicate the corresponding standard deviation.}\label{fig.dmn_residual}
\end{figure}

Fig. \ref{fig.dmn_residual}(a) shows that including the residual stress term has no significant effect on prediction accuracy, as measured by the metric defined in Eq. \eqref{eq.metric_stress}.
As shown in Fig. \ref{fig.dmn_residual}(b), the online scheme that incorporates the residual stress requires less computational time during online testing, achieving an average speed-up of 1.8$\times$ - 2.0$\times$.
Moreover, Fig. \ref{fig.dmn_residual}(c) shows that this scheme converges in significantly fewer iterations, owing to the correction effects of the residual stress term.

In general, the online computational cost depends on the number of active base nodes, the number of iterations required for convergence, and the computational complexity of the forward and backward passes.
To isolate the intrinsic algorithmic cost, the online testing time is normalized by both the number of active base nodes and the number of iterations, yielding the per-iteration-per-node cost, as shown in Fig. \ref{fig.dmn_residual}(d).
This metric reflects the intrinsic computational complexity of forward and backward passes and shows that the online scheme with residual stress incurs a higher per-iteration per-node cost, around 1.3$\times$ - 1.5$\times$ that of the scheme without residual stress, due to the additional computations associated with the residual stress term.

Overall, these results demonstrate that despite a higher per-iteration per-node computational cost, the DMN online scheme that incorporates residual stress achieves superior overall efficiency compared with the scheme without residual stress. 

\subsection{IMN Online Prediction - Fixed-Point vs Newton}\label{sec:imn_FPvsNT}
In this section, we investigate the online performance of IMN using both the fixed-point iteration scheme and the Newton iteration scheme, as described in Section \ref{sec:imn_online}.
Each trained IMN is evaluated with both online algorithms.
The trained IMN models are used to predict the material responses of Composite 1 subjected to six loading conditions.

\begin{figure}[htp]
\centering
    \begin{subfigure}{0.495\textwidth}
        \centering
        \includegraphics[width=1\linewidth]{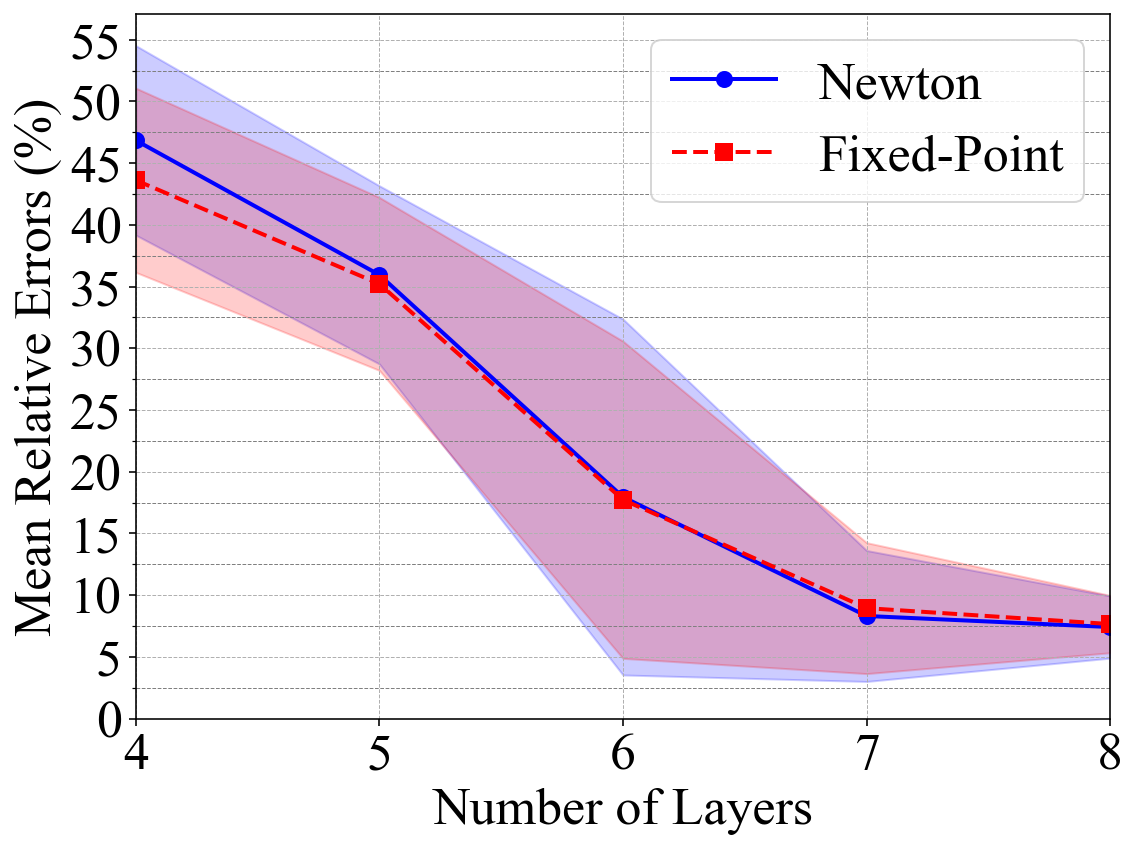}
        \caption{}
    \end{subfigure}
    \begin{subfigure}{0.495\textwidth}
        \centering
        \includegraphics[width=1\linewidth]{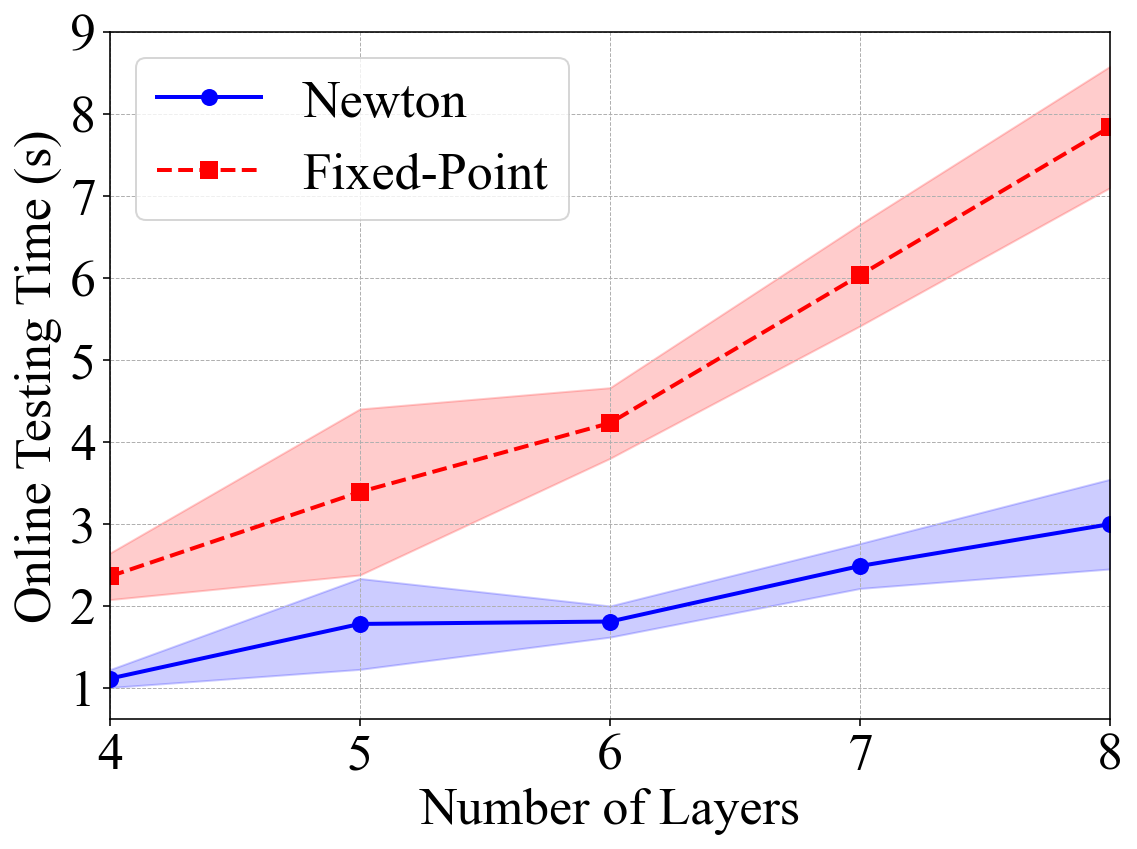}
        \caption{}
    \end{subfigure}
    \begin{subfigure}{0.495\textwidth}
        \centering
        \includegraphics[width=1\linewidth]{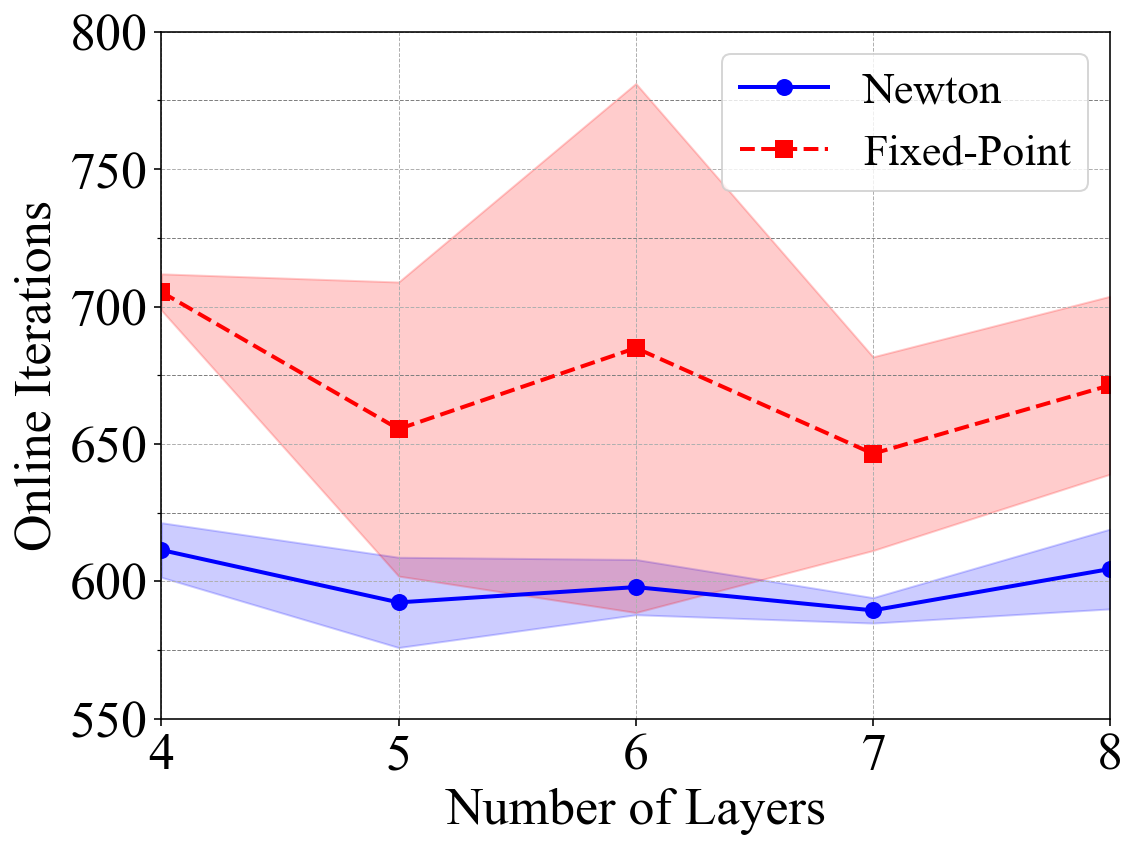}
        \caption{}
    \end{subfigure}
    \begin{subfigure}{0.495\textwidth}
        \centering
        \includegraphics[width=1\linewidth]{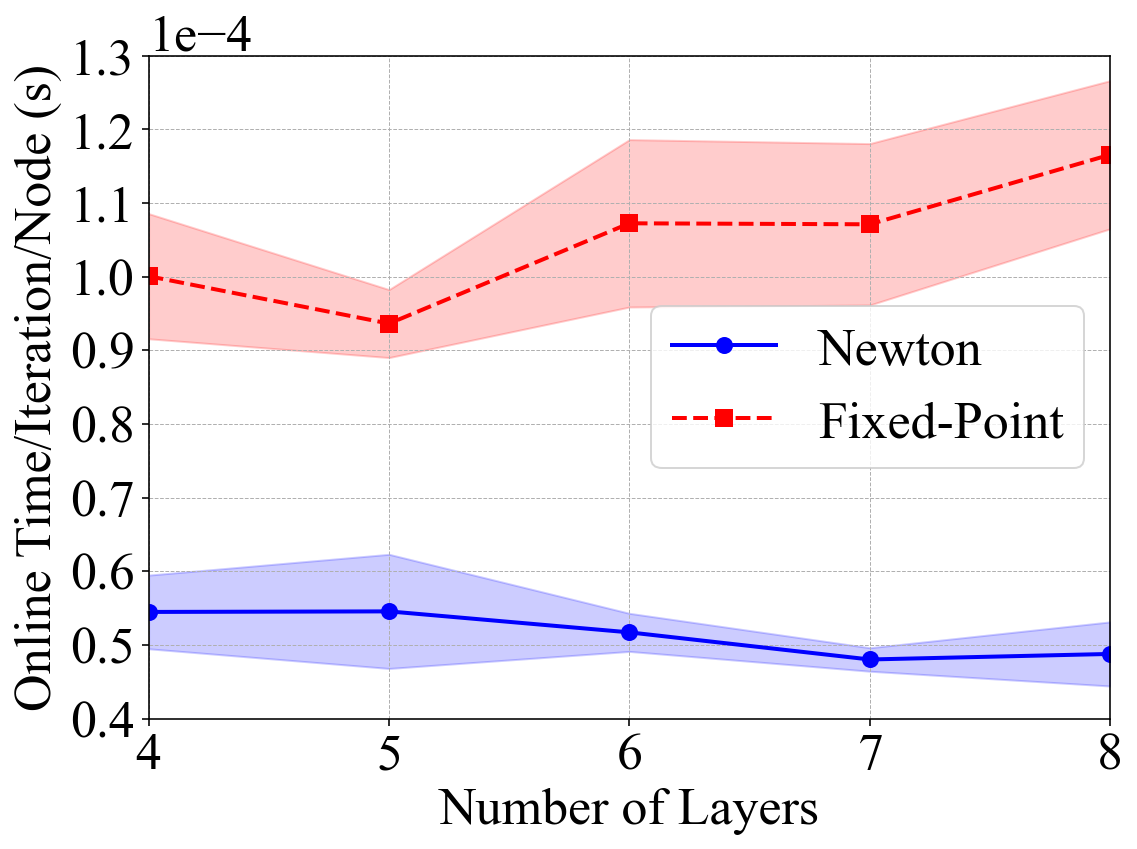}
        \caption{}
    \end{subfigure}
\caption{Fixed-point iterations vs Newton iterations for IMN online prediction: (a) mean relative errors; (b) online testing time; (c) number of iterations required for convergence; (d) online testing time per iteration per node; Solid line shows the mean over 5 runs with random initializations, and the shaded regions indicate the corresponding standard deviation.}\label{fig.imn_FPvsNT}
\end{figure}

Fig. \ref{fig.imn_FPvsNT}(a) shows that the two online schemes achieve comparable prediction accuracy, as measured by the mean relative error defined in Eq. \eqref{eq.metric_stress}.
As shown in Fig. \ref{fig.imn_FPvsNT}(b), the Newton iteration scheme requires less computational time during online testing, achieving an average speed-up of 1.9$\times$ - 2.6$\times$.
In addition, Fig. \ref{fig.imn_FPvsNT}(c) shows that the Newton iteration scheme requires fewer iterations to converge, owing to its more accurate search directions obtained from the Jacobian.

To isolate the intrinsic algorithmic cost, the online testing time is normalized by both the number of active base nodes and the number of iterations, yielding the per-iteration-per-node cost, as shown in Fig. \ref{fig.imn_FPvsNT}(d).
This metric reflects the intrinsic computational complexity of the forward and backward passes for the two online algorithms and shows that the Newton iteration scheme is more efficient, achieving an average speed-up of 1.7$\times$ - 2.4$\times$ per iteration per node.

Overall, these results demonstrate that the Newton iteration scheme provides superior efficiency for IMN. 
In the following subsection, IMN with Newton iterations is compared against DMN in terms of online performance.

\subsection{DMN vs IMN - Online Prediction}\label{sec:dmn_vs_imn_online}
In this section, we compare the online prediction performance of DMN and IMN. 
The trained models are used to predict the material responses of Composite 1 subjected to six loading conditions.

\begin{figure}[htp]
\centering
    \begin{subfigure}{0.325\textwidth}
        \centering
        \includegraphics[width=1\linewidth]{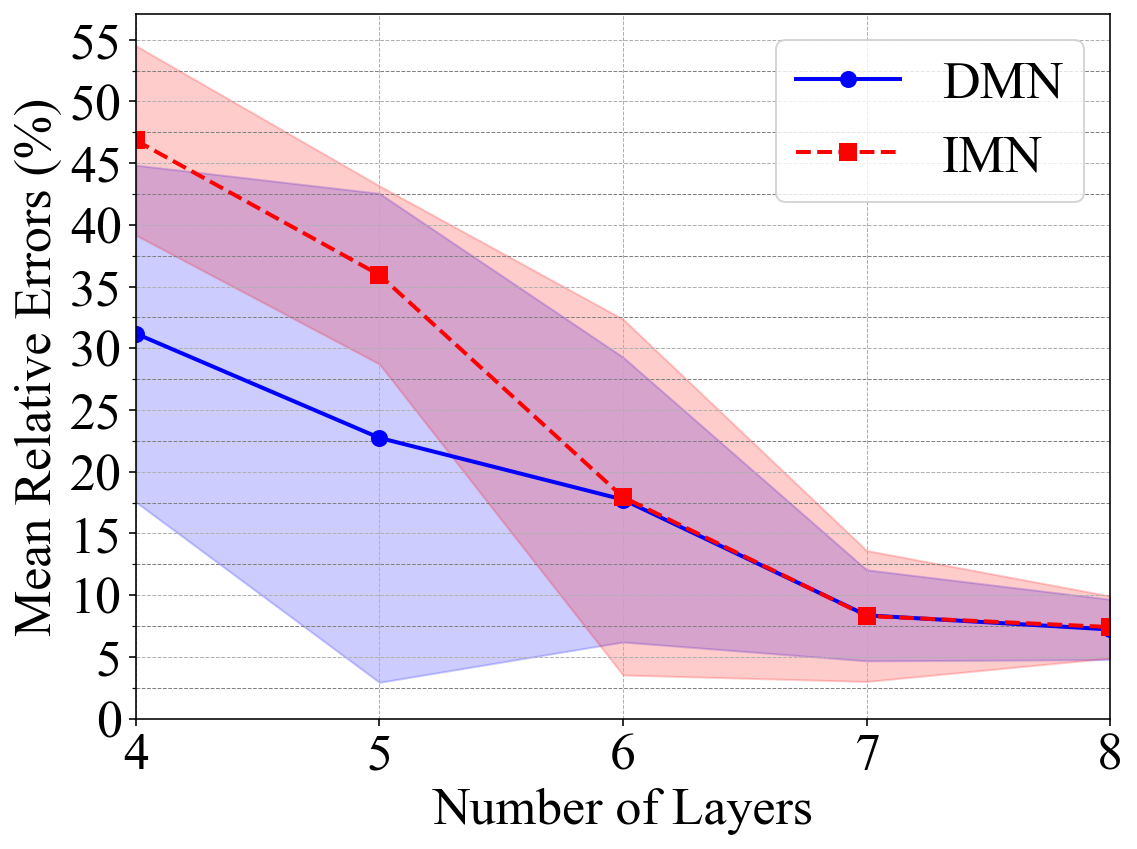}
        \caption{}
    \end{subfigure}
    \begin{subfigure}{0.325\textwidth}
        \centering
        \includegraphics[width=1\linewidth]{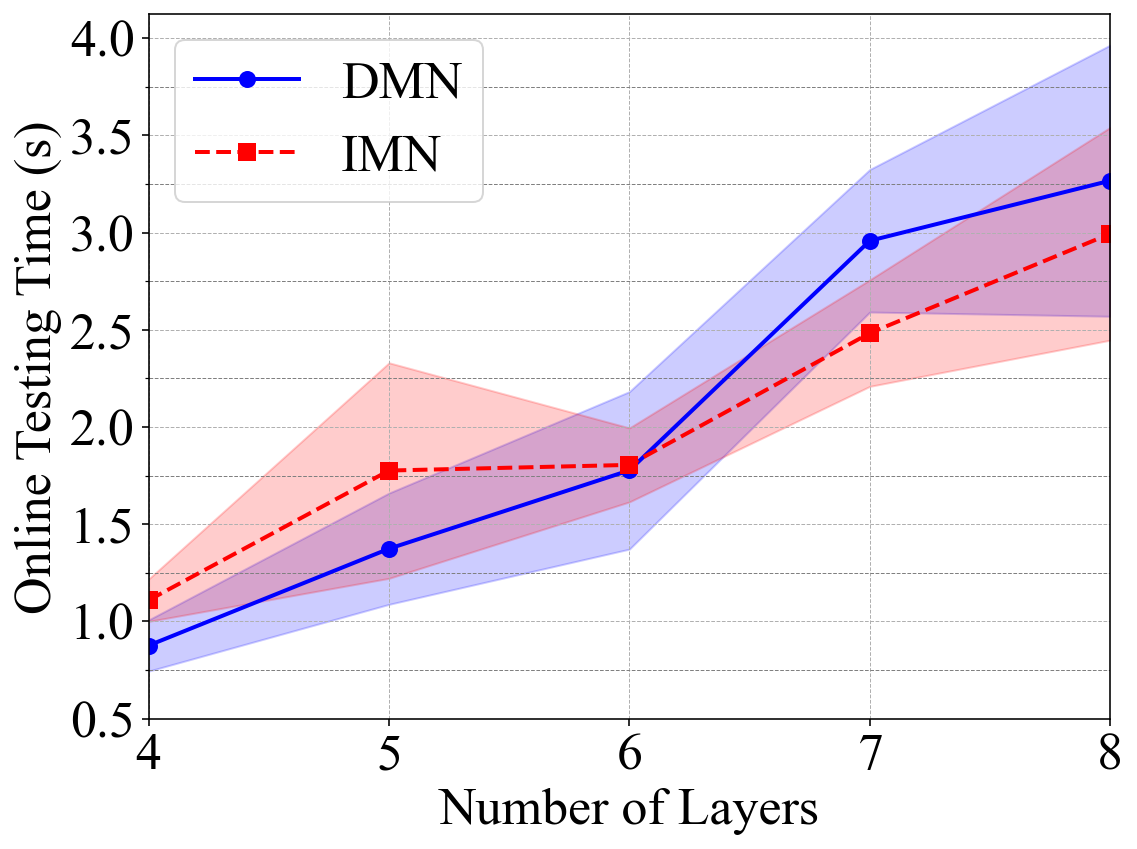}
        \caption{}
    \end{subfigure}
    \begin{subfigure}{0.325\textwidth}
        \centering
        \includegraphics[width=1\linewidth]{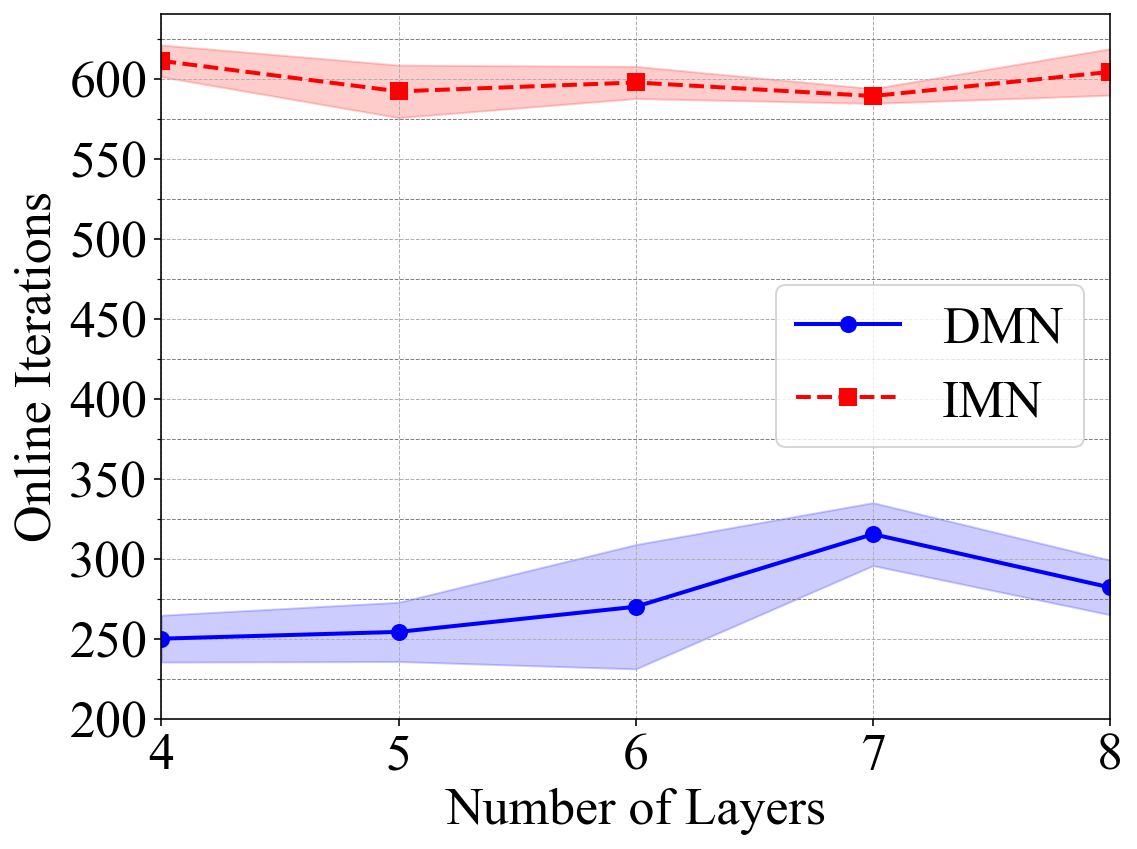}
        \caption{}
    \end{subfigure}
        \begin{subfigure}{0.325\textwidth}
        \centering
        \includegraphics[width=1\linewidth]{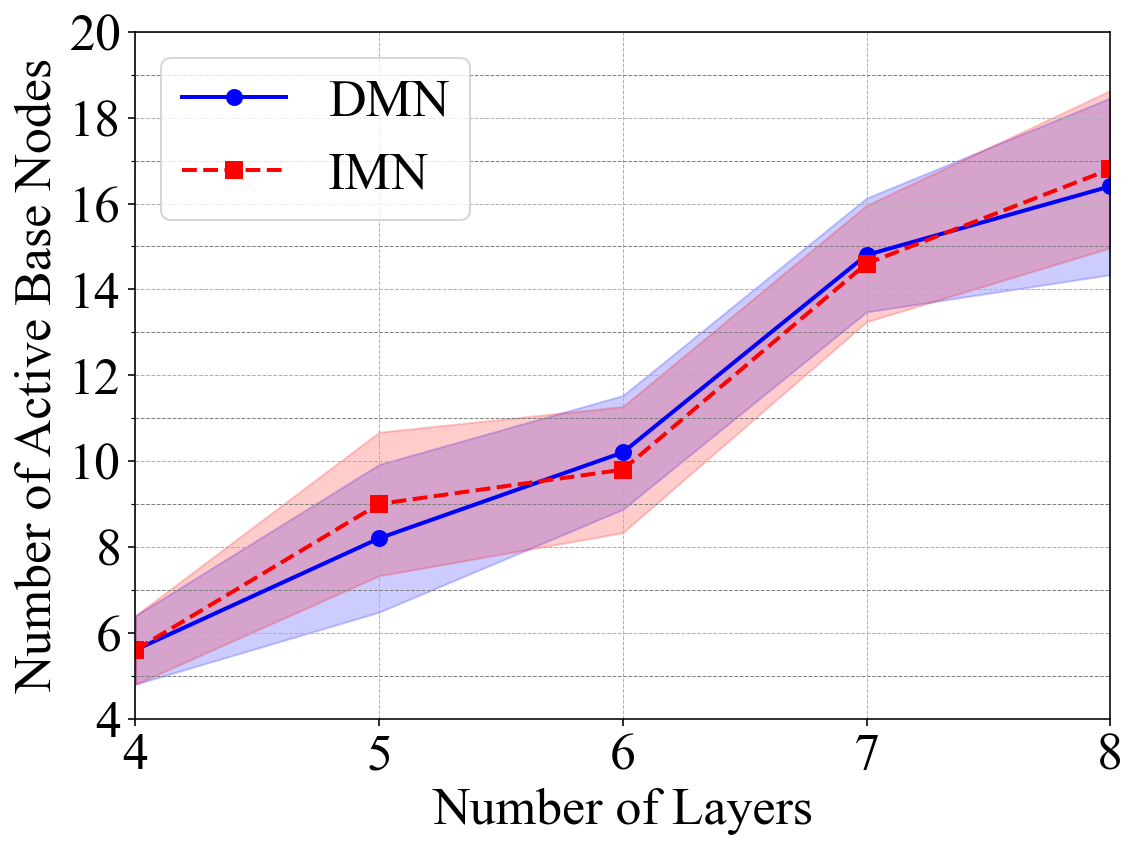}
        \caption{}
    \end{subfigure}
    \begin{subfigure}{0.325\textwidth}
	\centering
	\includegraphics[width=1\linewidth]{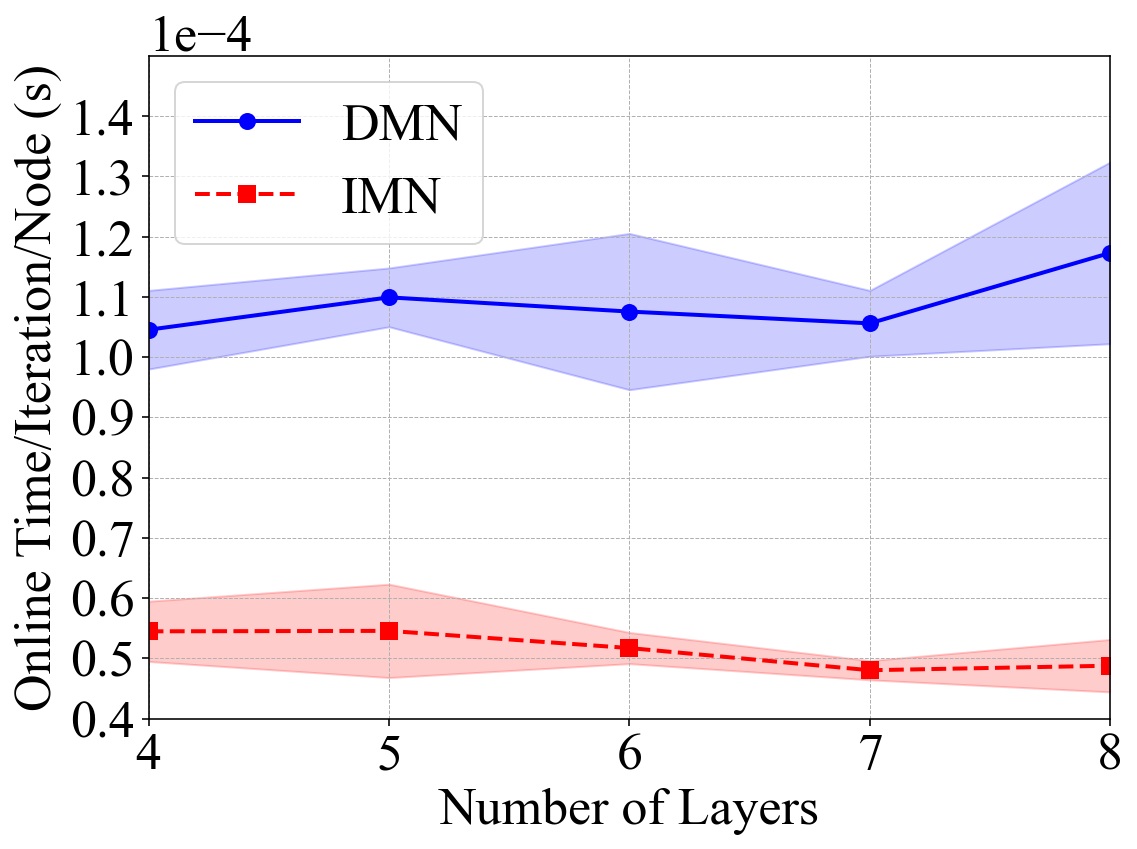}
	\caption{}
\end{subfigure}
\caption{DMN vs IMN for online prediction: (a) mean relative errors; (b) online testing time; (c) number of iterations required for convergence; (d) number of active base nodes; (e) online testing time per iteration per node; Solid line shows the mean over 5 runs with random initializations, and the shaded regions indicate the corresponding standard deviation.}\label{fig.dmn_imn_online}
\end{figure}

Fig. \ref{fig.dmn_imn_online}(a) shows that both models achieve comparable prediction accuracy (Eq. \eqref{eq.metric_stress}) when the network depth exceeds six layers, whereas DMN exhibits higher accuracy in shallower networks.
In general, prediction uncertainty decreases with increasing network depth.
Fig. \ref{fig.dmn_imn_online}(b) shows that the online testing times of both models increase with network depth and remain comparable across all configurations.

As shown in Fig. \ref{fig.dmn_imn_online}(c), DMN requires significantly fewer online iterations to converge (0.4$\times$ - 0.5$\times$ those of IMN), owing to the residual stress in its online algorithm (see Sections \ref{sec:dmn_online} and \ref{sec:dmn_residual}).
The iteration counts for both DMN and IMN are relatively insensitive to network depth, though DMN exhibits slightly greater variability.
Fig. \ref{fig.dmn_imn_online}(d) shows that the number of active base nodes increases with network depth for both models, with comparable values for each depth.
This indicates that, under the same regularization settings in the loss function, DMN and IMN yield similar network complexity, enabling a fair performance comparison.

To isolate the intrinsic computational complexity of the two models, the online testing time is normalized by the number of active base nodes and the number of iterations, yielding the per-iteration-per-node cost shown in Fig. \ref{fig.dmn_imn_online}(e).
This metric captures the cost of forward and backward passes for online prediction, independent of iteration and node counts.
The results show that IMN is more efficient, achieving a speed-up of 1.9$\times$ - 2.4$\times$ per iteration per node.
However, this advantage is offset by the larger number of iterations required for IMN to converge (Fig. \ref{fig.dmn_imn_online}(c)), resulting in an overall online computational cost comparable to that of DMN (Fig. \ref{fig.dmn_imn_online}(b)). 
Overall, these results demonstrate that DMN and IMN achieve comparable online prediction accuracy and efficiency.

\section{Conclusion}\label{sec:conclusion}
This work presents a systematic performance assessment of structure-preserving material networks, namely the Deep Material Network (DMN) originally proposed in \cite{liu2019deep,liu2019exploring} and its variant, the rotation-free interaction-based material network (IMN) \cite{gajek2020micromechanics,noels2022interaction}.
Unlike purely data-driven models, DMN embeds micromechanical building blocks within a hierarchical binary-tree architecture with physically interpretable parameters.
This structure-preserving design enables strong generalization capability: although trained exclusively on linear elastic data, the networks accurately predict the nonlinear inelastic responses of composites composed of previously unseen base materials. This property highlights their significant potential for practical multiscale material modeling.

Through extensive parametric studies, we investigated the effects of training data size, batch size, initialization, and loss-function regularization on model accuracy, robustness, and computational efficiency. 
The results demonstrate that while initialization and batch size influence model training and performance, increasing the training data size consistently improves prediction accuracy and reduces uncertainty. 
In addition, the regularization parameters ($\xi, \eta$) for base node activation directly influence network complexity and consequently, online prediction accuracy and computational cost. 
Specifically, $\eta$ controls the strength of regularization, whereas $\xi$ strongly affects network complexity. 
Based on the parametric analysis, a balanced trade-off between accuracy and efficiency is achieved with $\eta=\xi=1$ for the examples considered in this study.

A comprehensive comparison between DMN and IMN reveals distinct advantages and trade-offs. Offline training results show that IMN achieves higher training efficiency, with a $3.4\times$–$4.7\times$ speed-up over DMN, primarily due to fewer trainable parameters (Section \ref{sec:imn}) and lower forward-pass computational complexity (Section \ref{sec:dmn_vs_imn_online}). 
Online performance evaluations demonstrate that DMN and IMN achieve comparable prediction accuracy and overall computational cost. 
While DMN benefits from fewer iterations to converge due to the use of residual stresses (Sections \ref{sec:dmn_residual} and \ref{sec:dmn_vs_imn_online}) in its online algorithm, IMN exhibits substantially lower per-iteration-per-node computational cost. 
These effects counterbalance each other, resulting in similar overall online efficiency. Furthermore, the Newton-based online iteration scheme for IMN is shown to significantly outperform fixed-point iterations, delivering up to a $2.6\times$ speed-up while maintaining comparable accuracy.
The results of this study provide practical guidance for their deployment in multiscale simulations. 

Overall, the reduced number of trainable parameters in IMN leads to improved offline training efficiency while preserving online prediction accuracy and efficiency comparable to those of DMN.
Future work will focus on developing more efficient and robust training strategies to mitigate the model's sensitivity to training settings, e.g., initialization \cite{shin2023deep} and activation regularization \ref{sec:regularization_effect}, as well as on designing improved sampling strategies \cite{dey2022training,srinivas2026rapid} to reduce training data requirements while maximizing learning performance.

\newpage

\bibliographystyle{elsarticle-num} 
\bibliography{reference}

\newpage
\appendix
\section{DMN Trained on Dataset 2}\label{sec:dmn_ns1024}
\begin{figure}[htp]
\centering
    \begin{subfigure}{0.325\textwidth}
        \centering
        \includegraphics[width=1\linewidth]{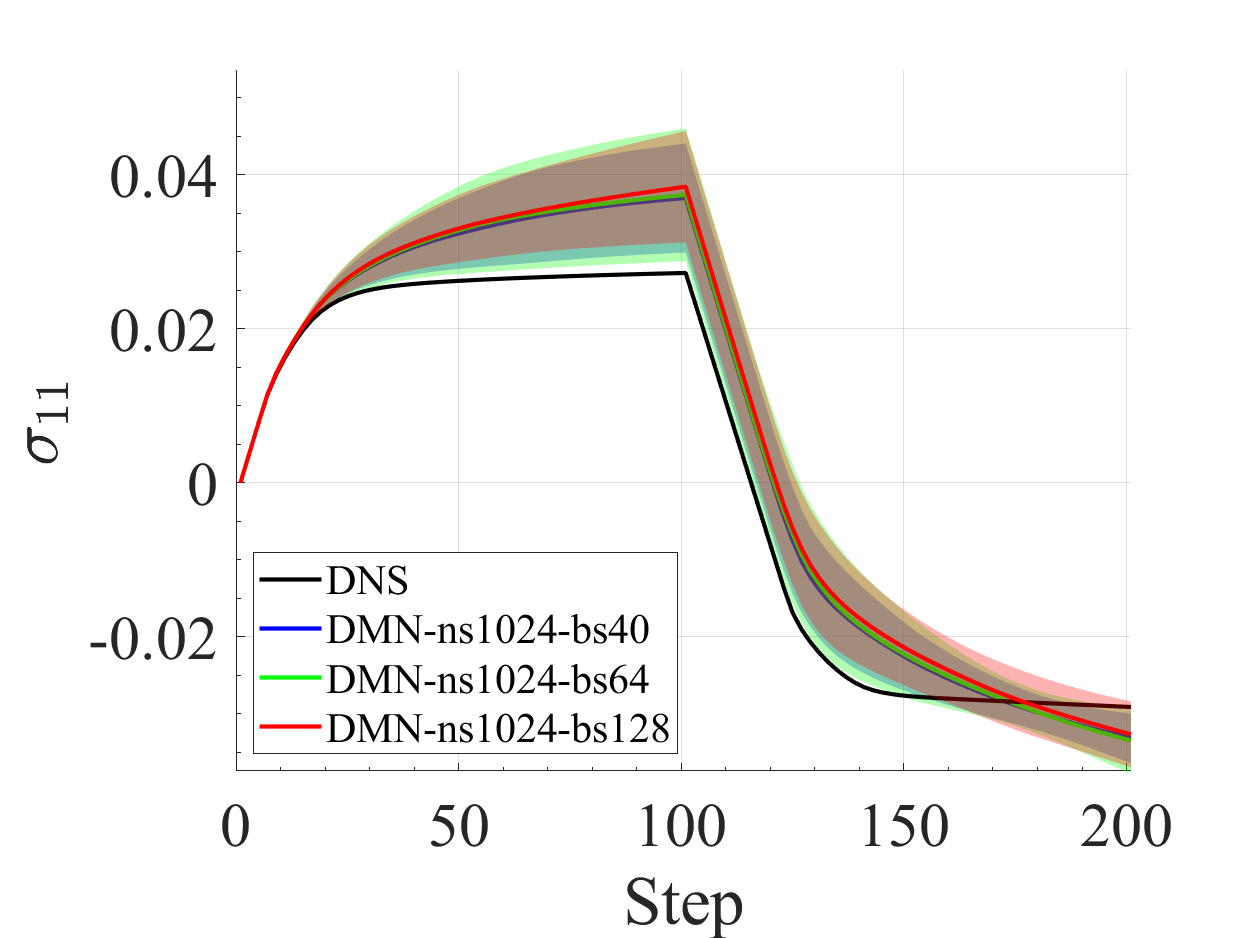}
        \caption{$\sigma_{11}$}
    \end{subfigure}
    \begin{subfigure}{0.325\textwidth}
        \centering
        \includegraphics[width=1\linewidth]{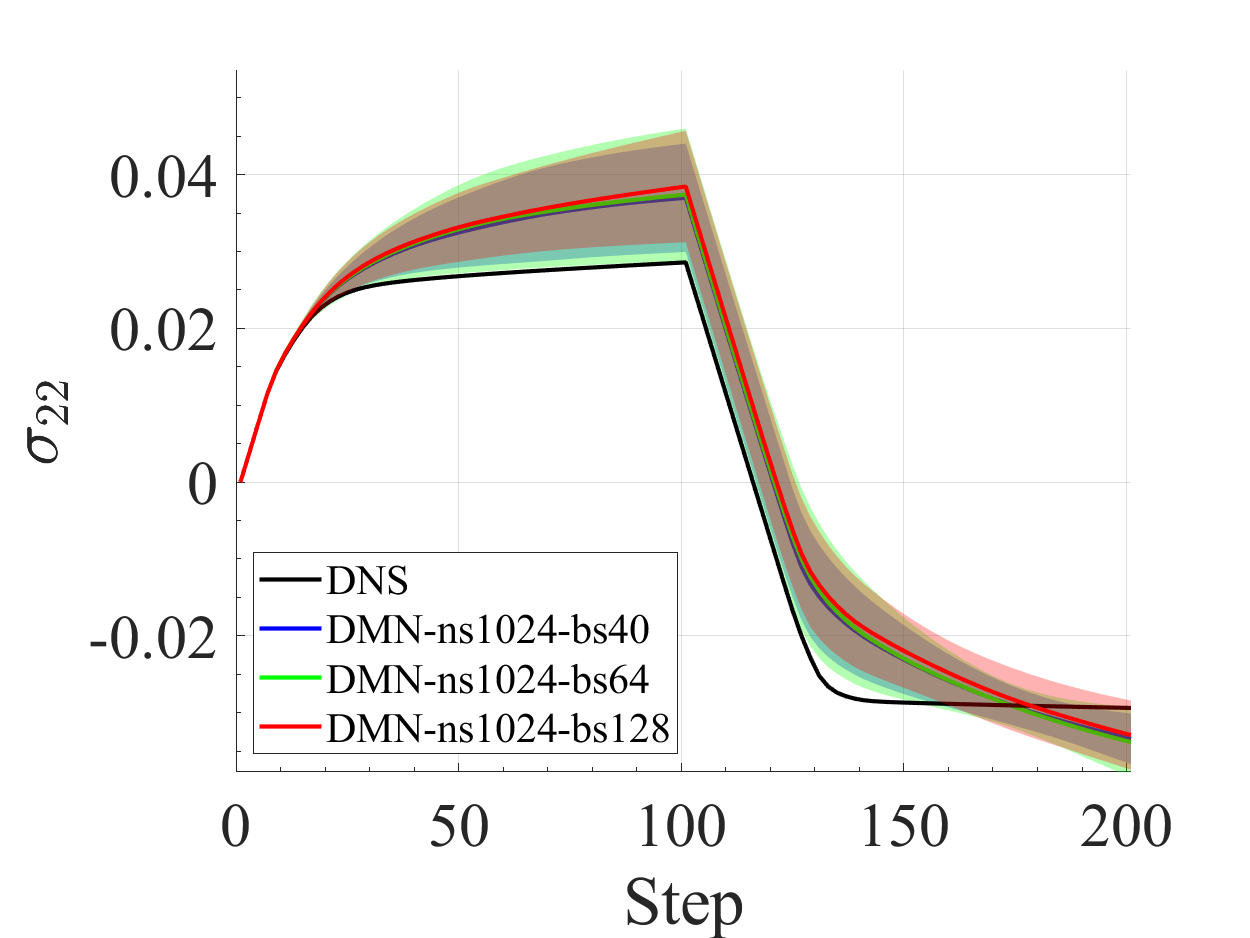}
        \caption{$\sigma_{22}$}
    \end{subfigure}
    \begin{subfigure}{0.325\textwidth}
        \centering
        \includegraphics[width=1\linewidth]{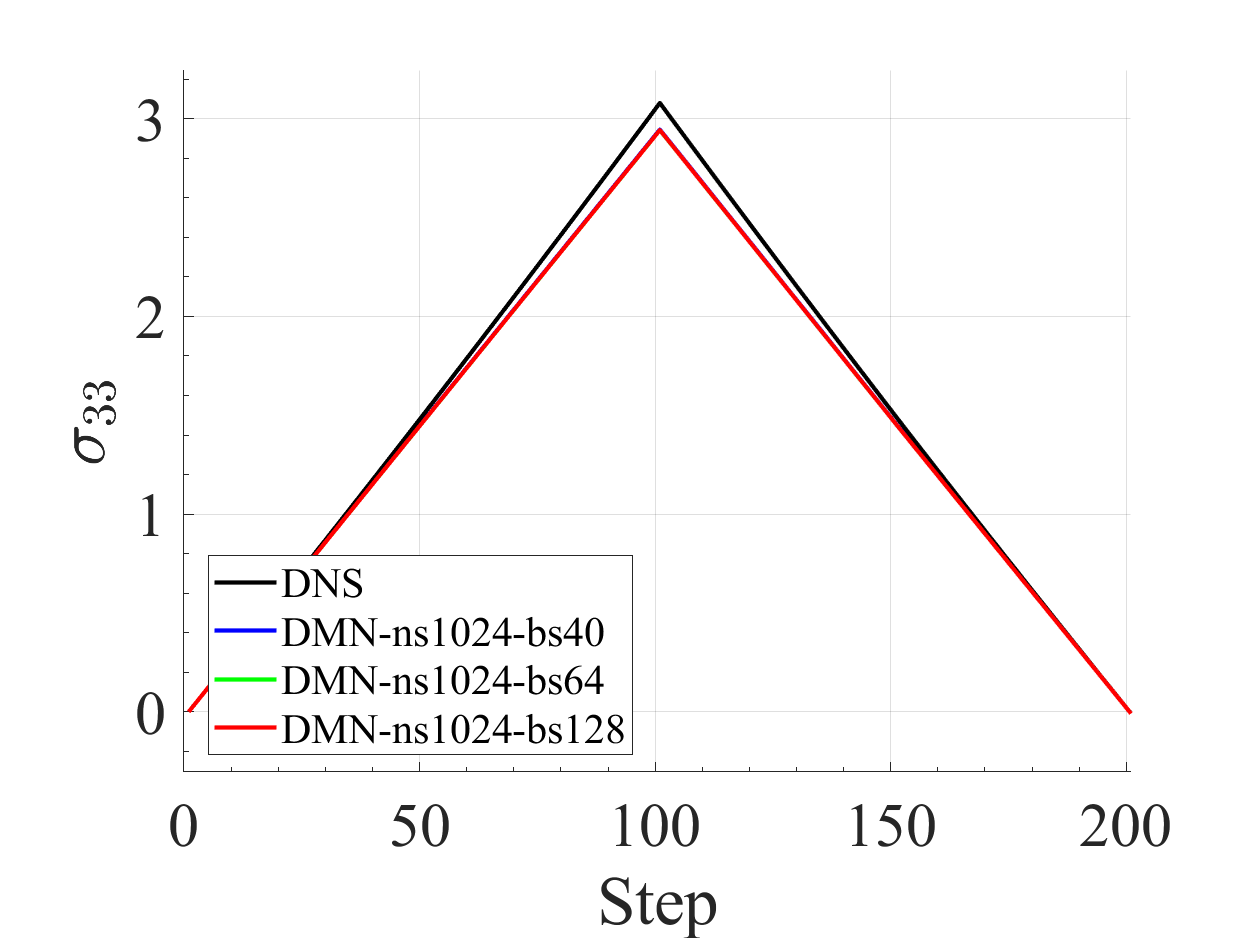}
        \caption{$\sigma_{33}$}
    \end{subfigure}
    \begin{subfigure}{0.325\textwidth}
        \centering
        \includegraphics[width=1\linewidth]{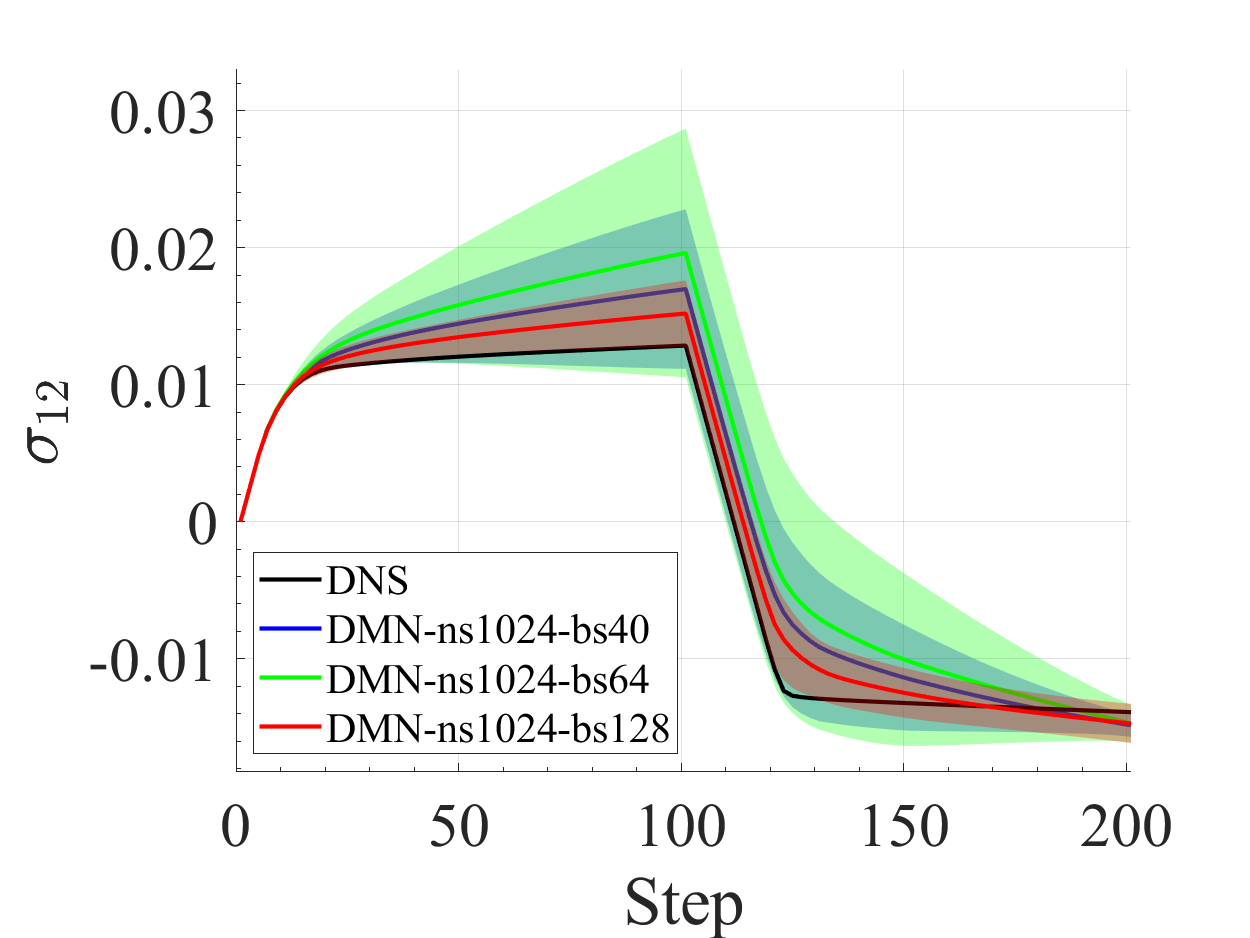}
        \caption{$\sigma_{12}$}
    \end{subfigure}
    \begin{subfigure}{0.325\textwidth}
        \centering
        \includegraphics[width=1\linewidth]{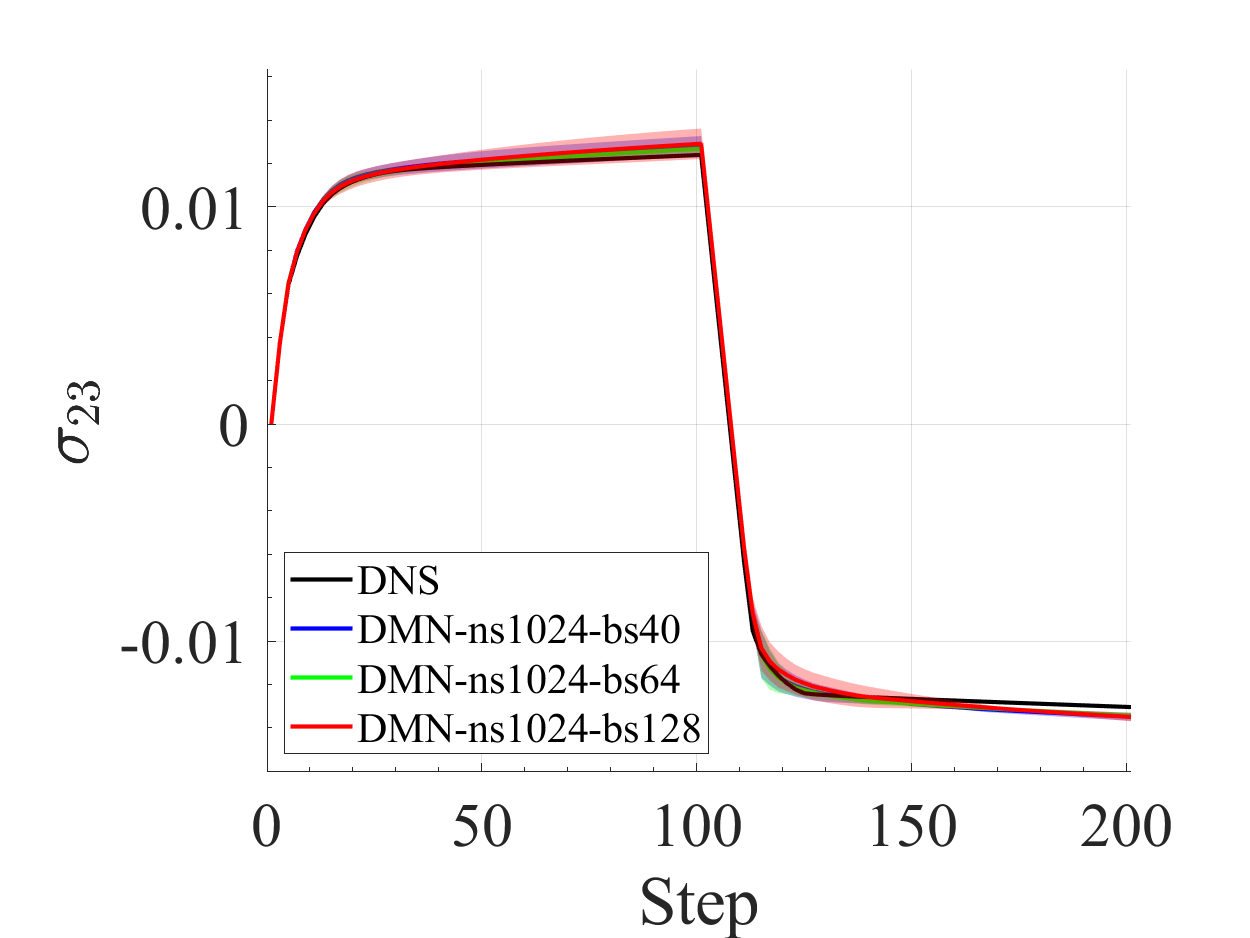}
        \caption{$\sigma_{23}$}
    \end{subfigure}
    \begin{subfigure}{0.325\textwidth}
        \centering
        \includegraphics[width=1\linewidth]{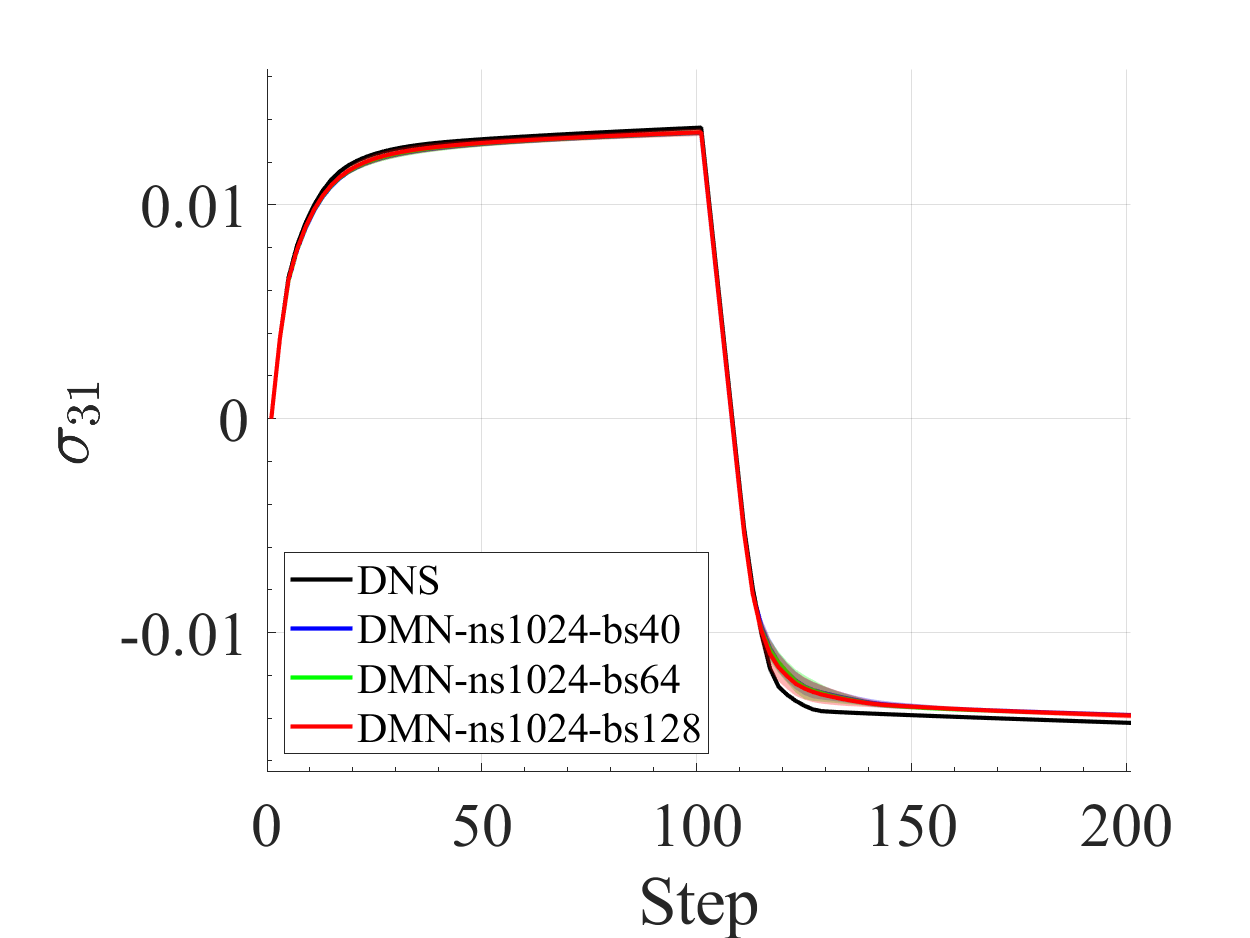}
        \caption{$\sigma_{31}$}
    \end{subfigure}
\caption{Composite 1 - Stress predictions from DMNs trained with 1024 samples using batch sizes of 40, 64, and 128: (a) $\sigma_{11}$; (b) $\sigma_{22}$; (c) $\sigma_{33}$; (d) $\sigma_{12}$; (e) $\sigma_{23}$; (f) $\sigma_{31}$; Solid lines show the mean over 10 runs with random initializations, and the shaded region indicate the corresponding standard deviations.}\label{fig.mat1_ns1024}
\end{figure}

\begin{figure}[htp]
\centering
    \begin{subfigure}{0.325\textwidth}
        \centering
        \includegraphics[width=1\linewidth]{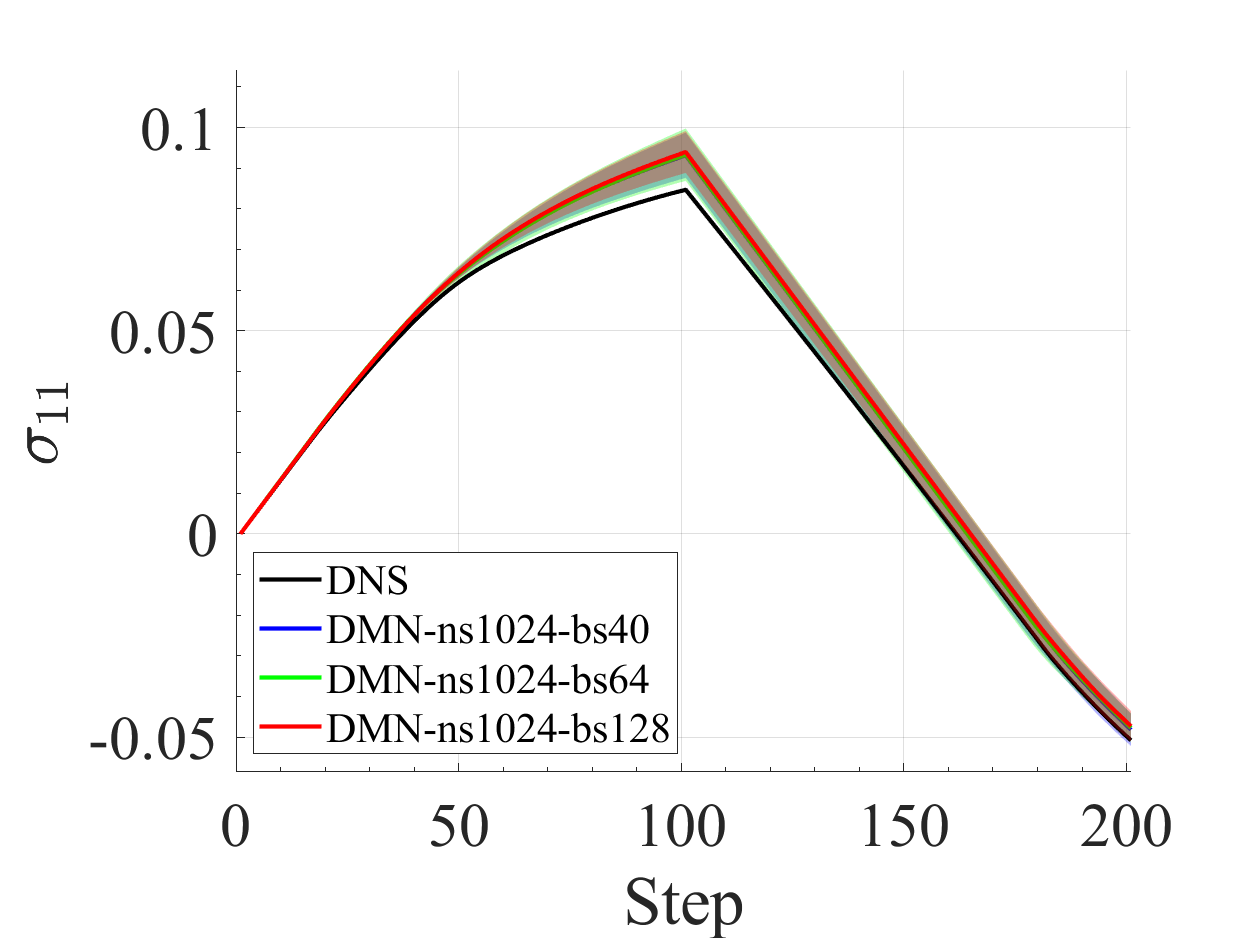}
        \caption{$\sigma_{11}$}
    \end{subfigure}
    \begin{subfigure}{0.325\textwidth}
        \centering
        \includegraphics[width=1\linewidth]{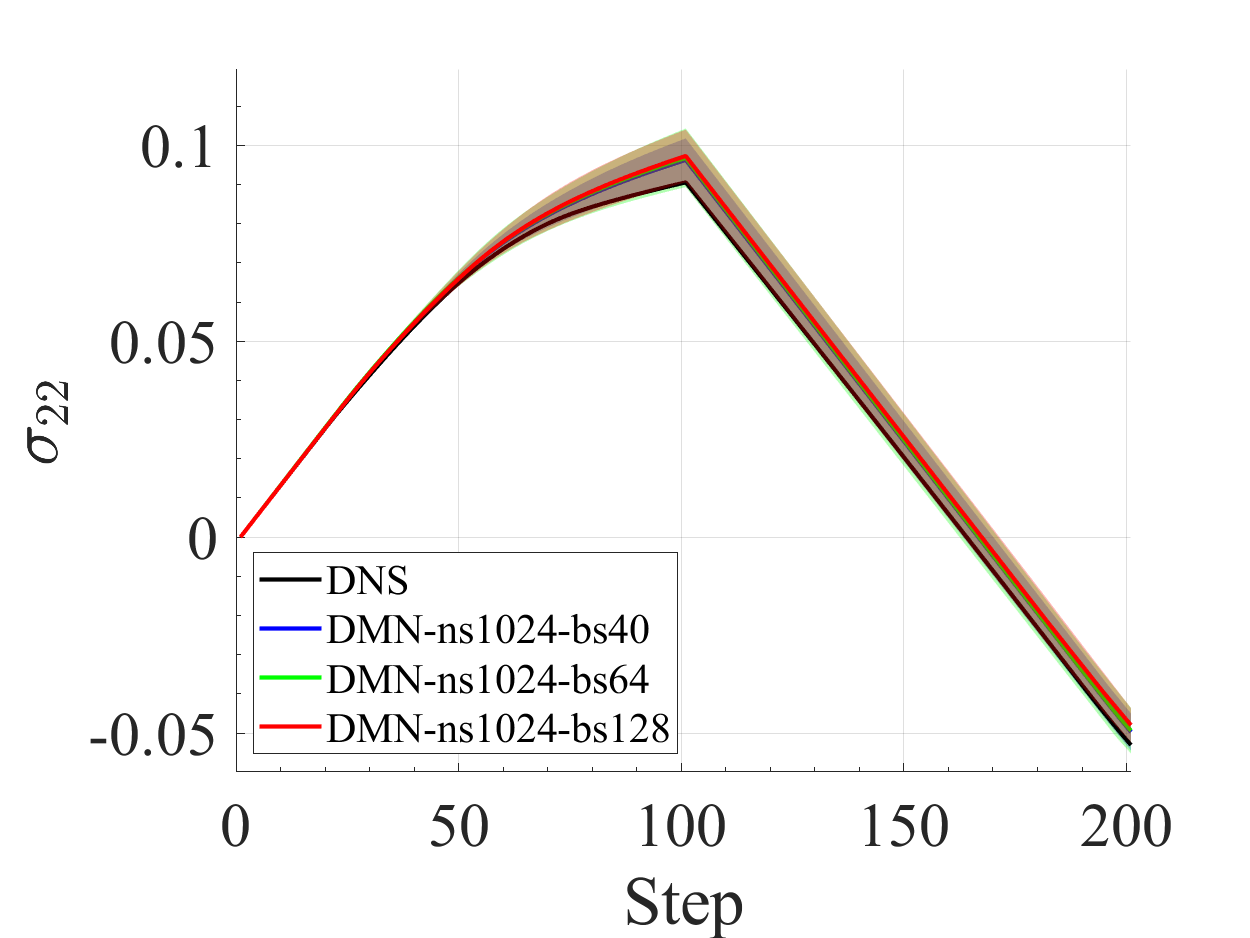}
        \caption{$\sigma_{22}$}
    \end{subfigure}
    \begin{subfigure}{0.325\textwidth}
        \centering
        \includegraphics[width=1\linewidth]{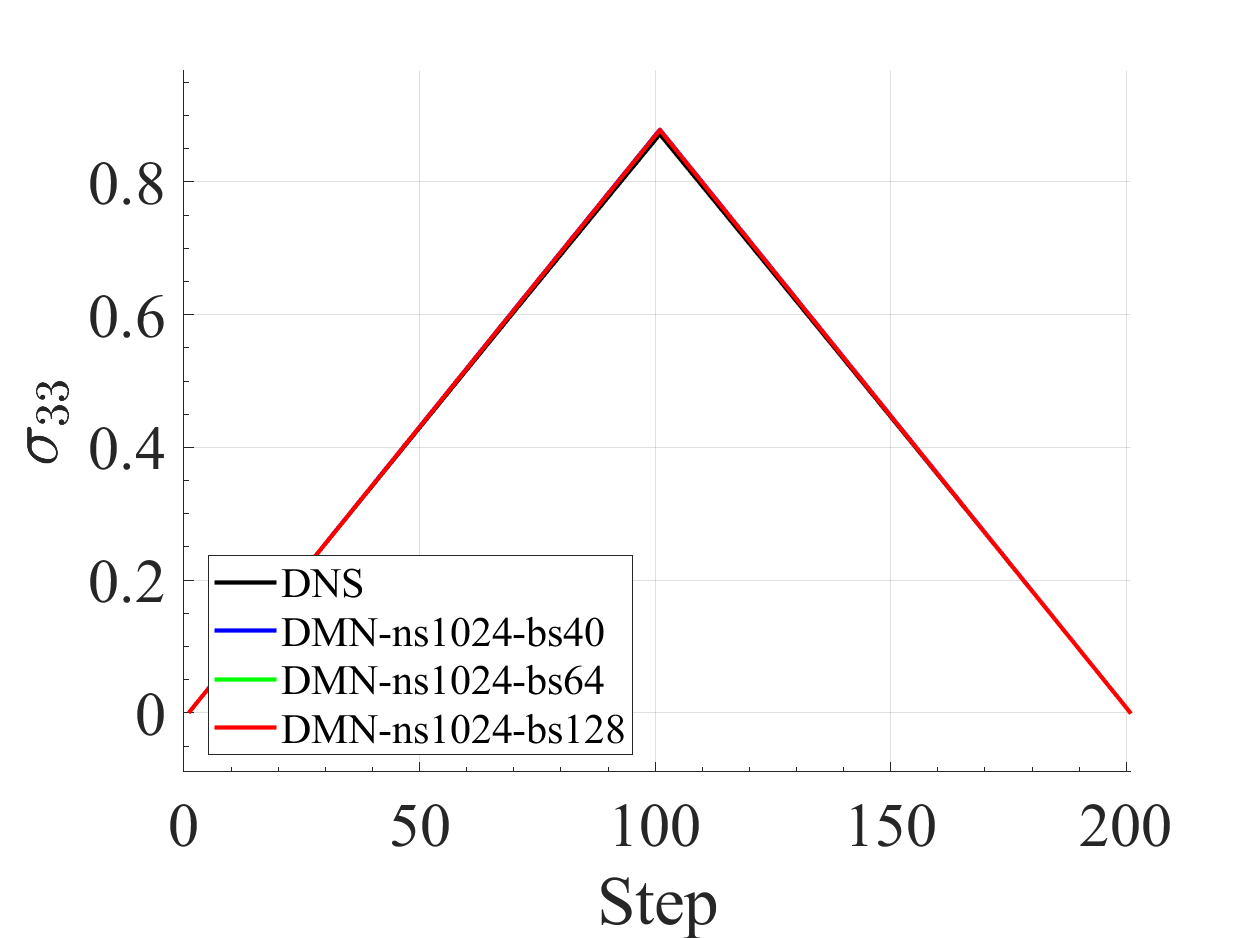}
        \caption{$\sigma_{33}$}
    \end{subfigure}
    \begin{subfigure}{0.325\textwidth}
        \centering
        \includegraphics[width=1\linewidth]{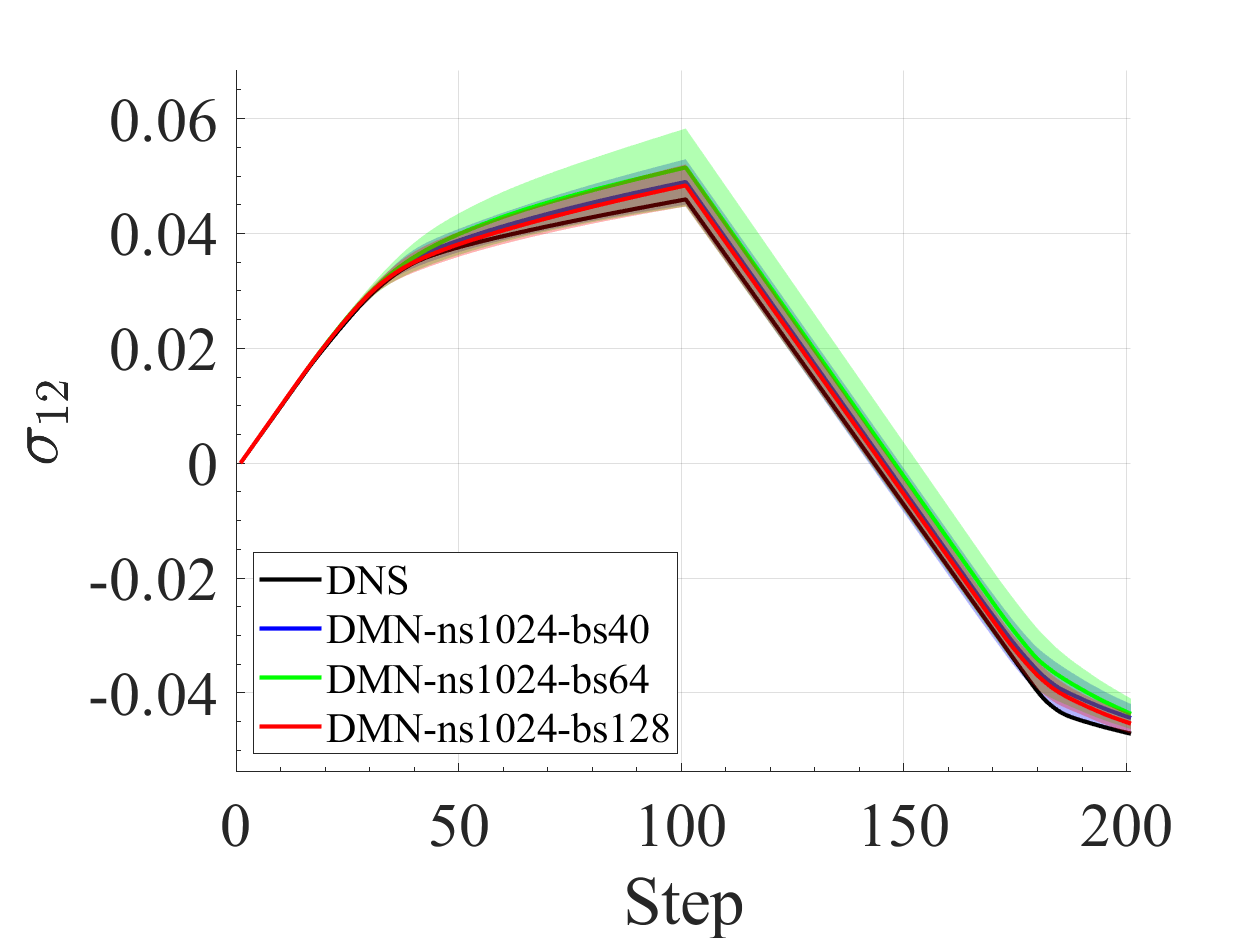}
        \caption{$\sigma_{12}$}
    \end{subfigure}
    \begin{subfigure}{0.325\textwidth}
        \centering
        \includegraphics[width=1\linewidth]{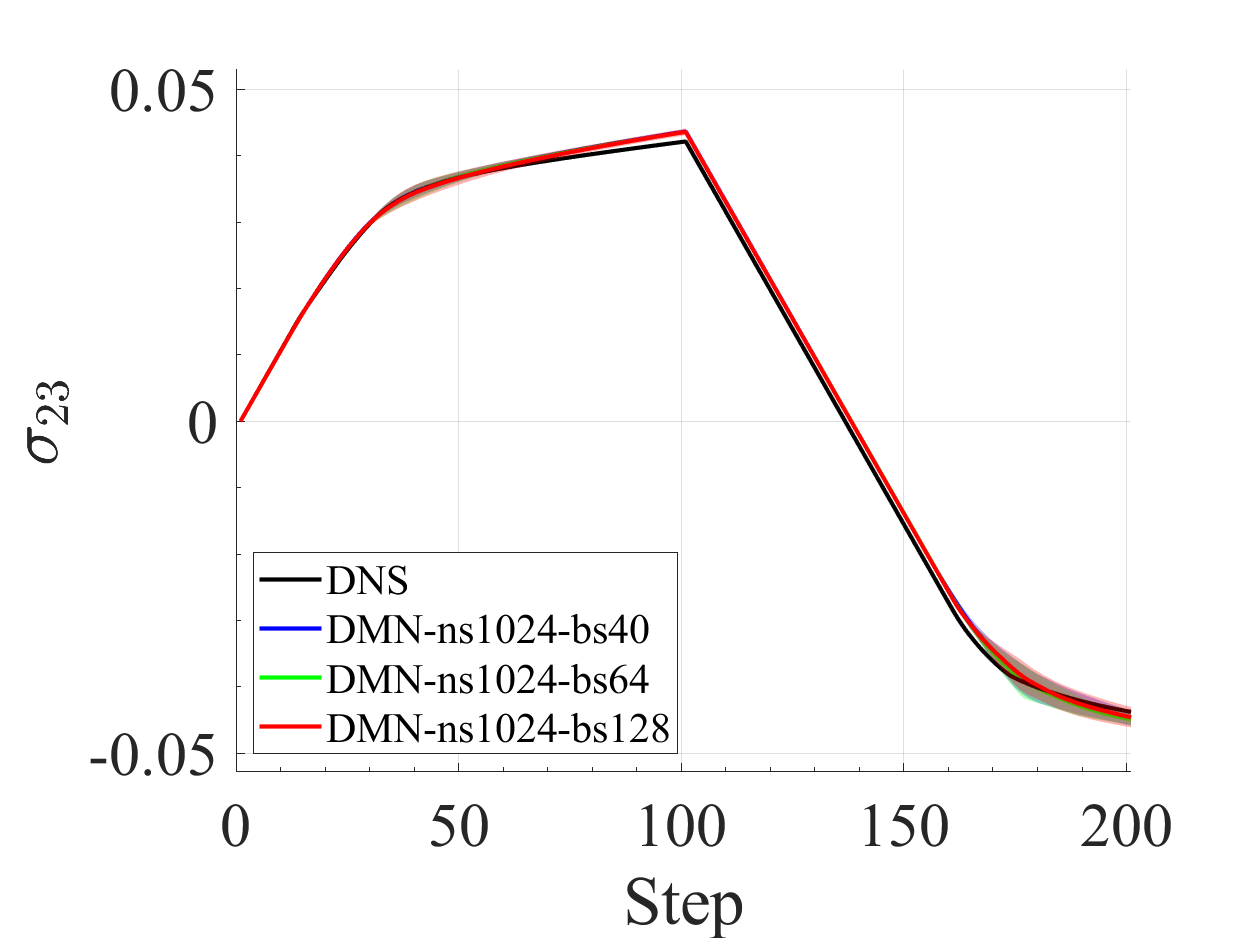}
        \caption{$\sigma_{23}$}
    \end{subfigure}
    \begin{subfigure}{0.325\textwidth}
        \centering
        \includegraphics[width=1\linewidth]{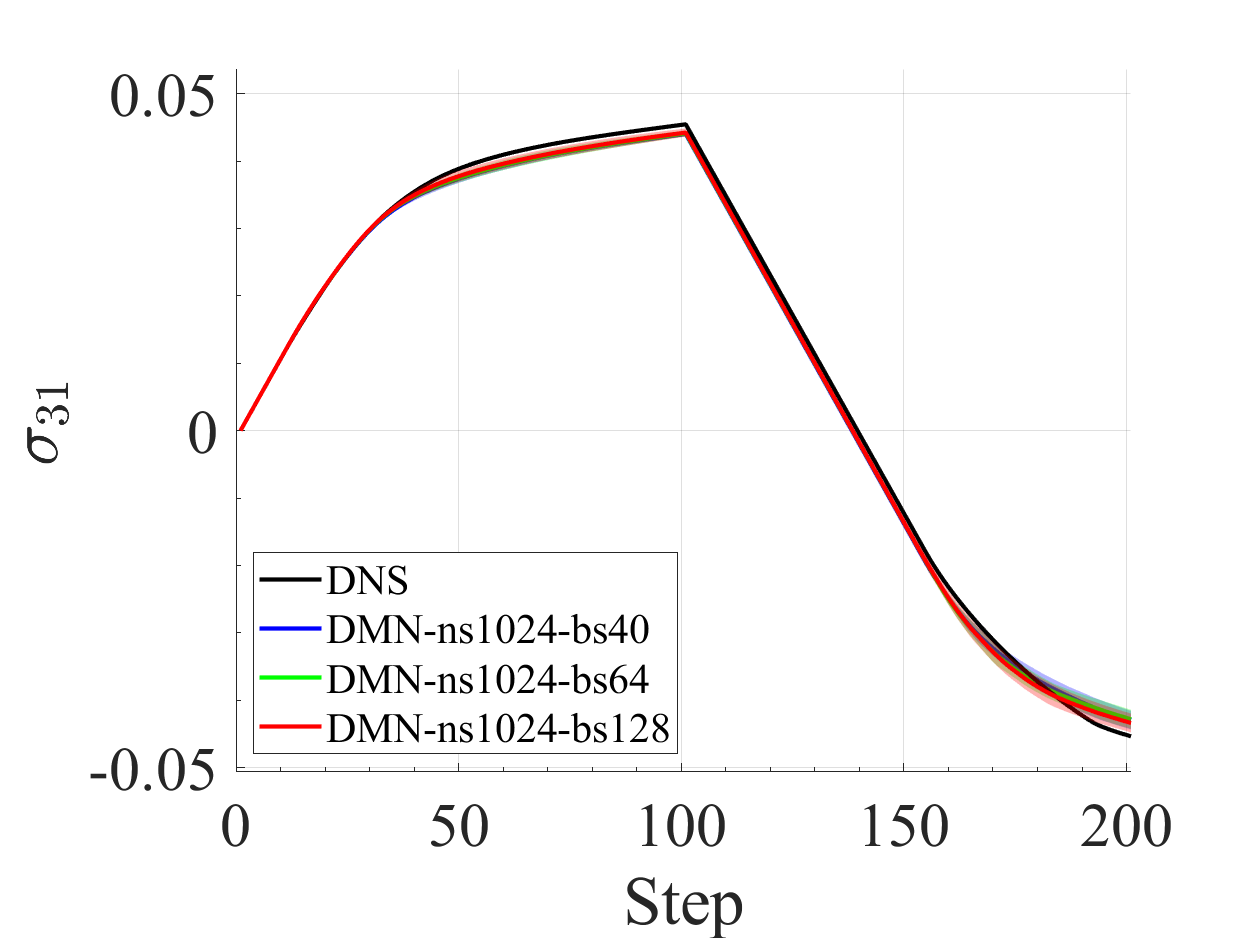}
        \caption{$\sigma_{31}$}
    \end{subfigure}
\caption{Composite 2 - Stress predictions from DMNs trained with 1024 samples using batch sizes of 40, 64, and 128: (a) $\sigma_{11}$; (b) $\sigma_{22}$; (c) $\sigma_{33}$; (d) $\sigma_{12}$; (e) $\sigma_{23}$; (f) $\sigma_{31}$; Solid lines show the mean over 10 runs with random initializations, and the shaded region indicate the corresponding standard deviations.}\label{fig.mat2_ns1024}
\end{figure}

\begin{figure}[htp]
\centering
    \begin{subfigure}{0.325\textwidth}
        \centering
        \includegraphics[width=1\linewidth]{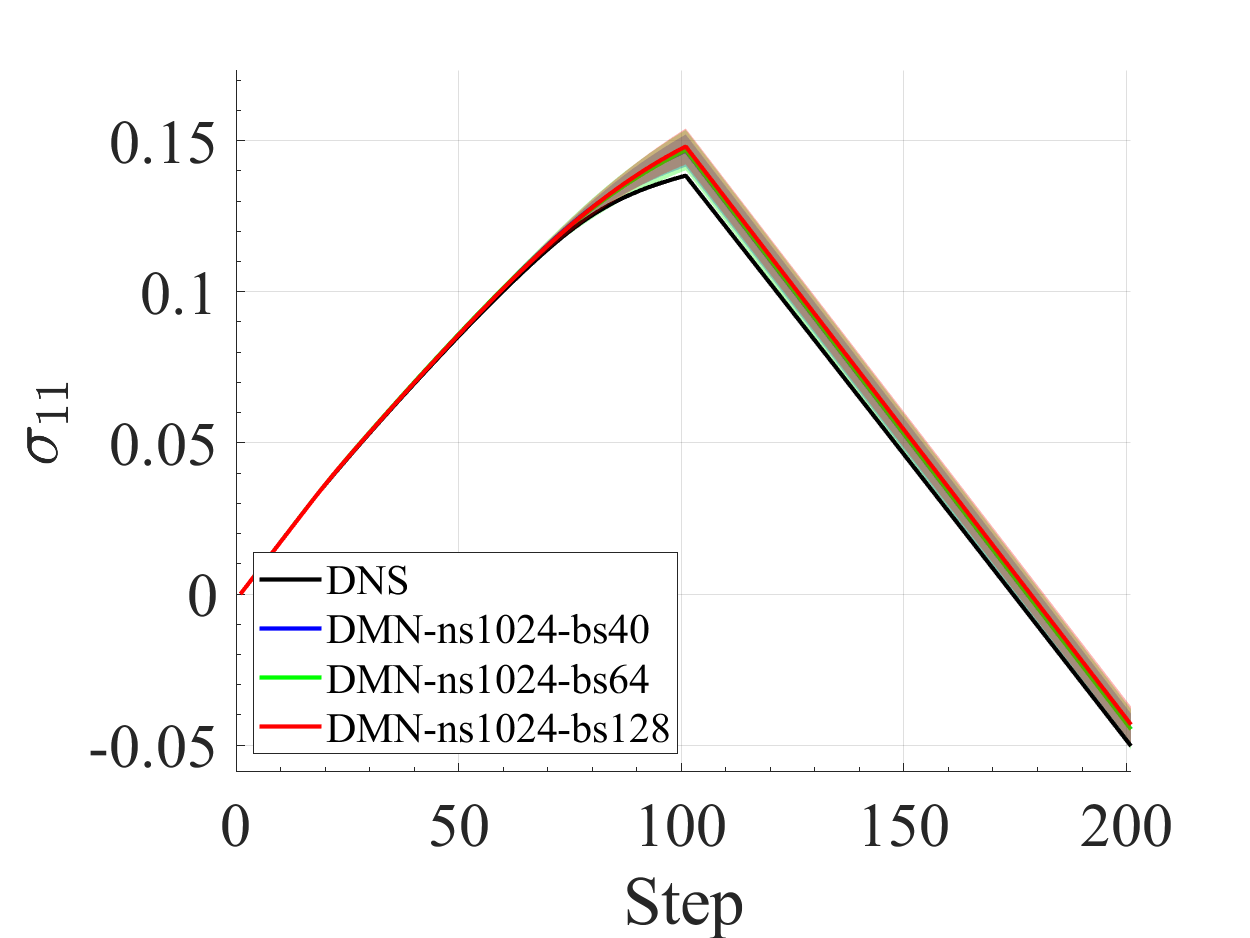}
        \caption{$\sigma_{11}$}
    \end{subfigure}
    \begin{subfigure}{0.325\textwidth}
        \centering
        \includegraphics[width=1\linewidth]{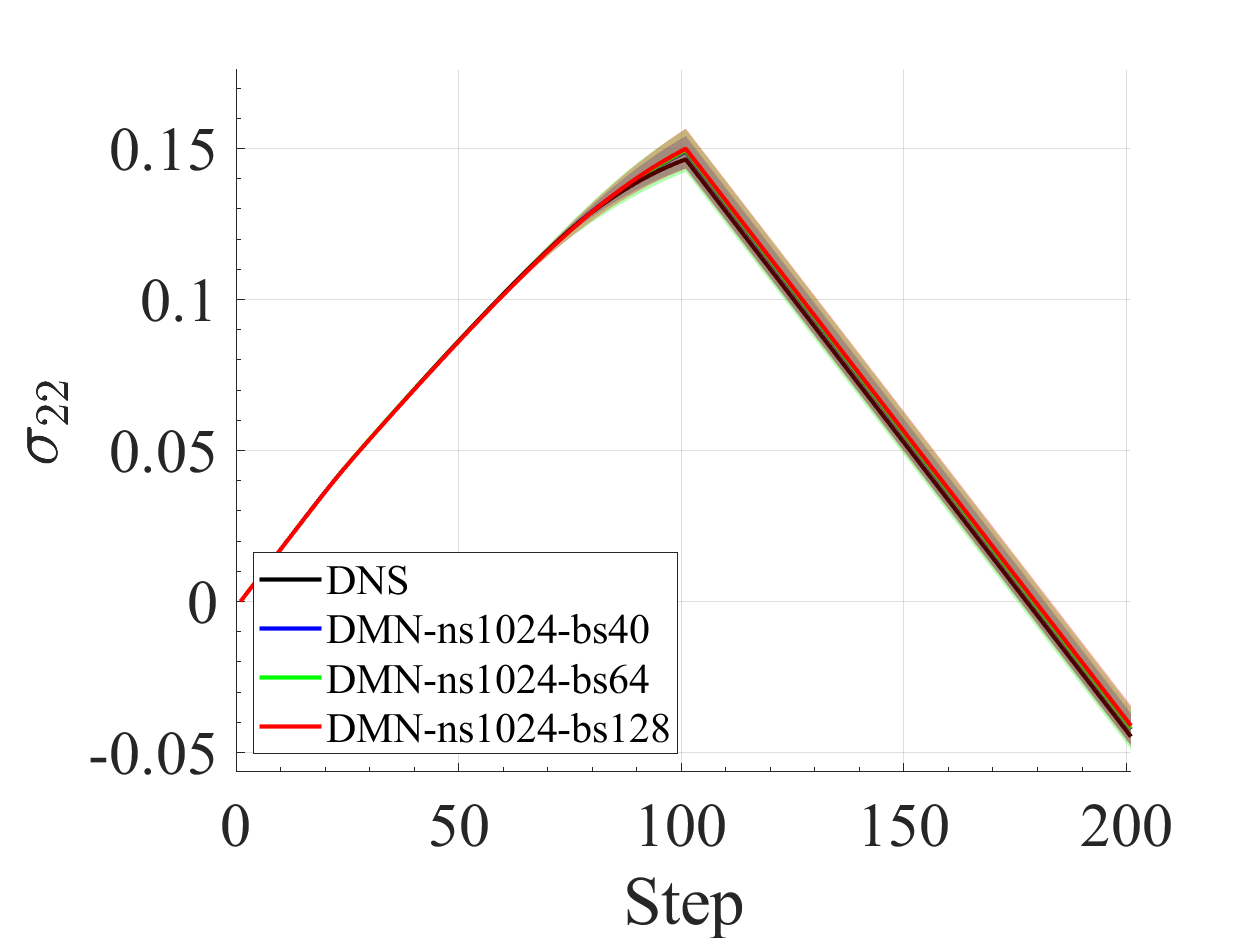}
        \caption{$\sigma_{22}$}
    \end{subfigure}
    \begin{subfigure}{0.325\textwidth}
        \centering
        \includegraphics[width=1\linewidth]{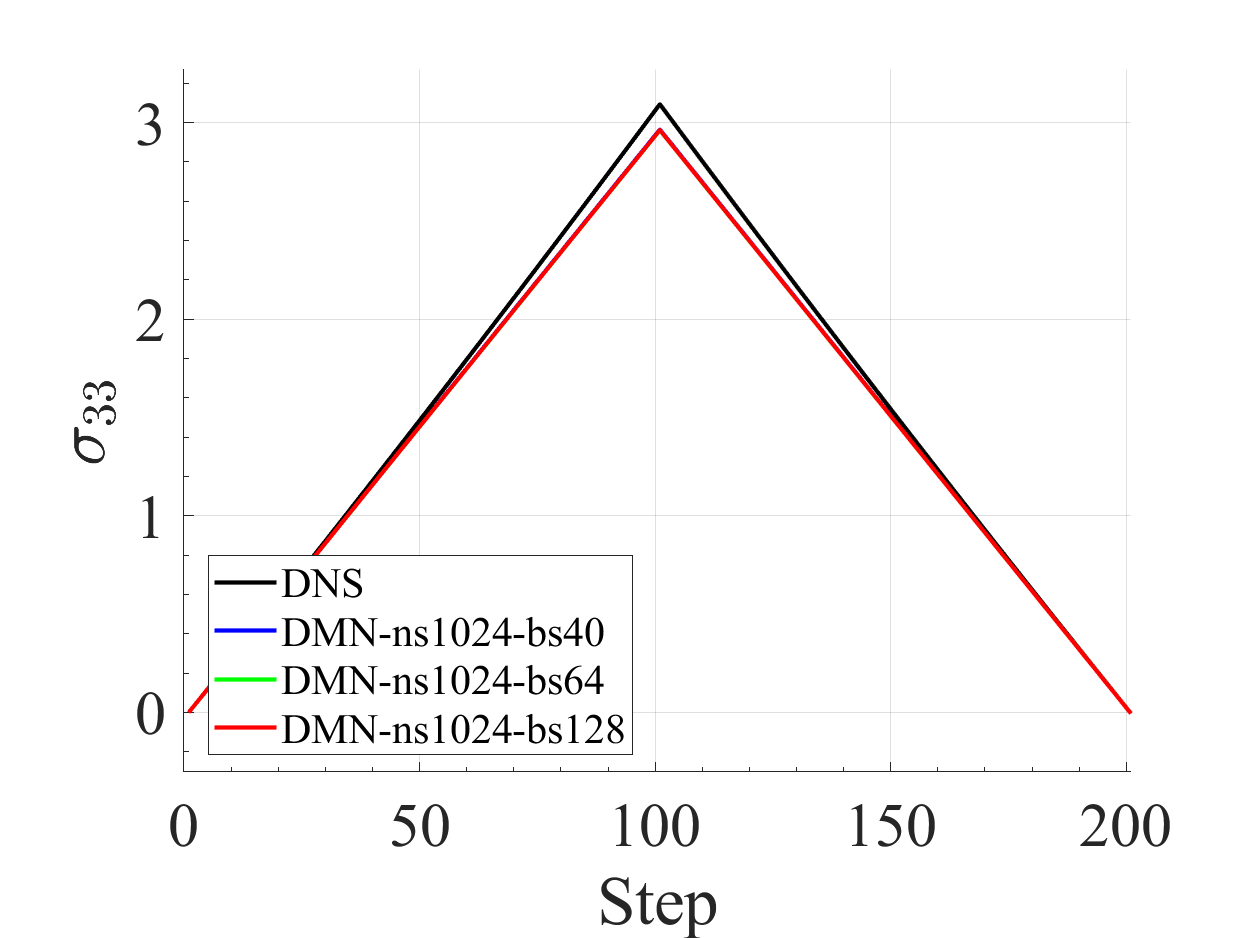}
        \caption{$\sigma_{33}$}
    \end{subfigure}
    \begin{subfigure}{0.325\textwidth}
        \centering
        \includegraphics[width=1\linewidth]{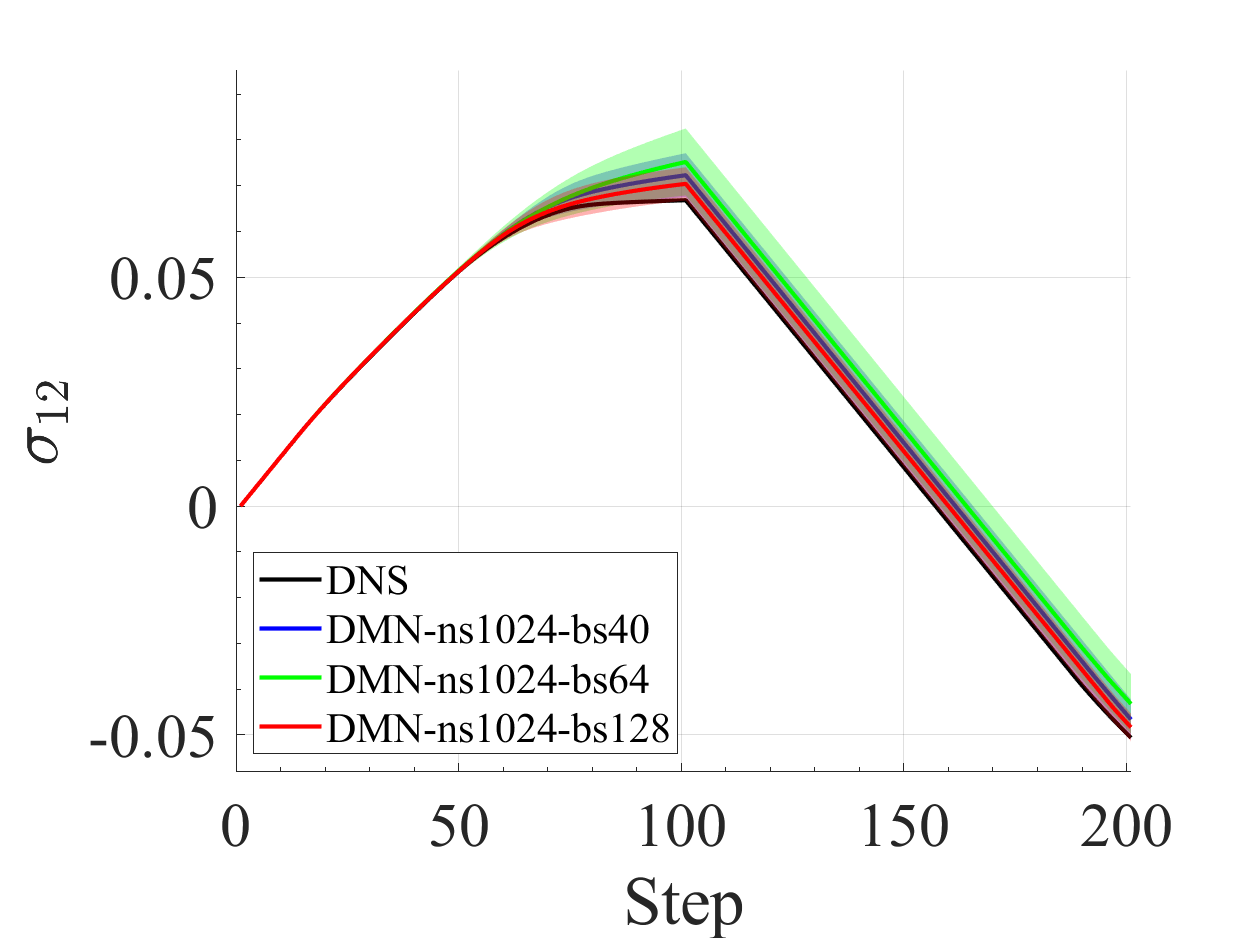}
        \caption{$\sigma_{12}$}
    \end{subfigure}
    \begin{subfigure}{0.325\textwidth}
        \centering
        \includegraphics[width=1\linewidth]{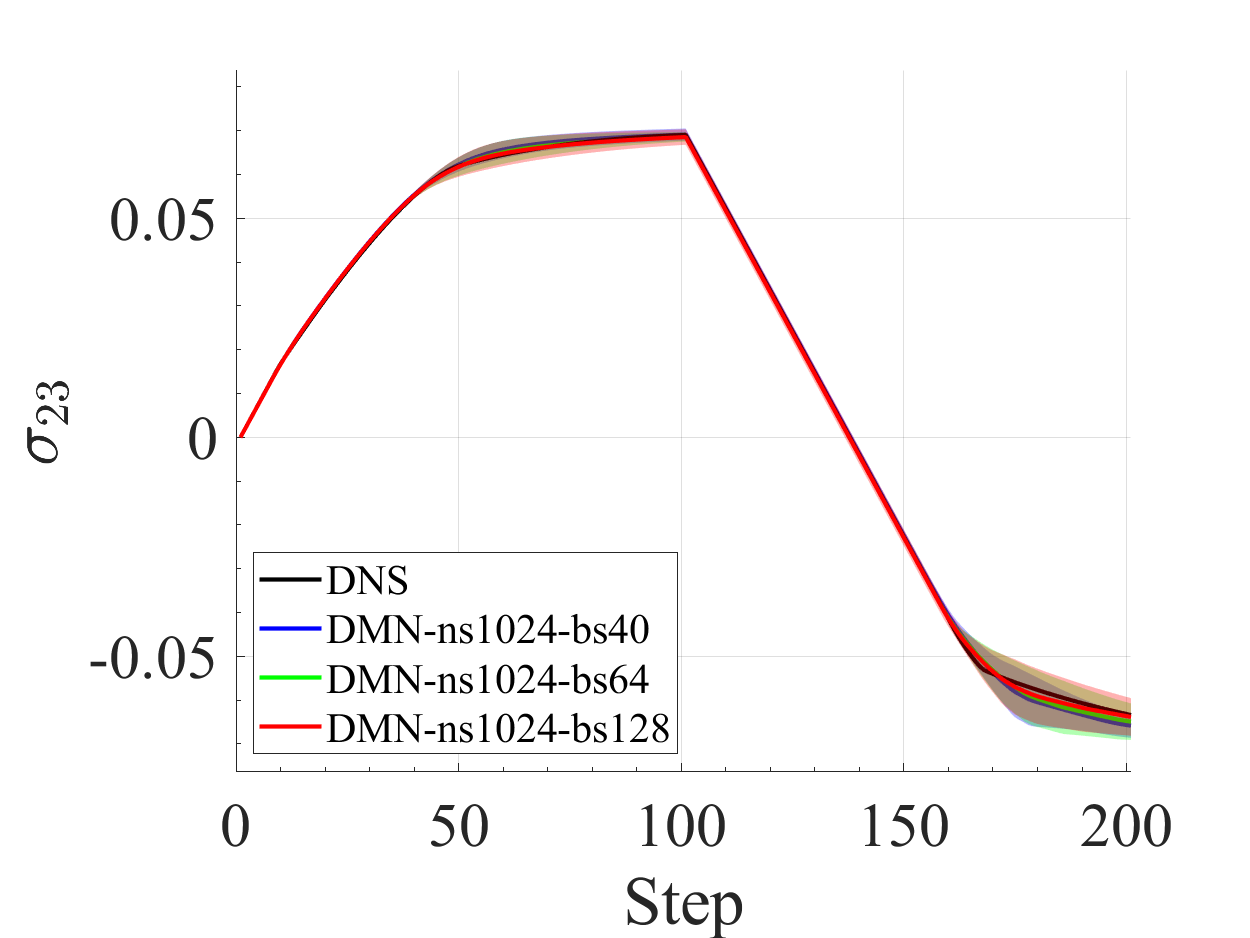}
        \caption{$\sigma_{23}$}
    \end{subfigure}
    \begin{subfigure}{0.325\textwidth}
        \centering
        \includegraphics[width=1\linewidth]{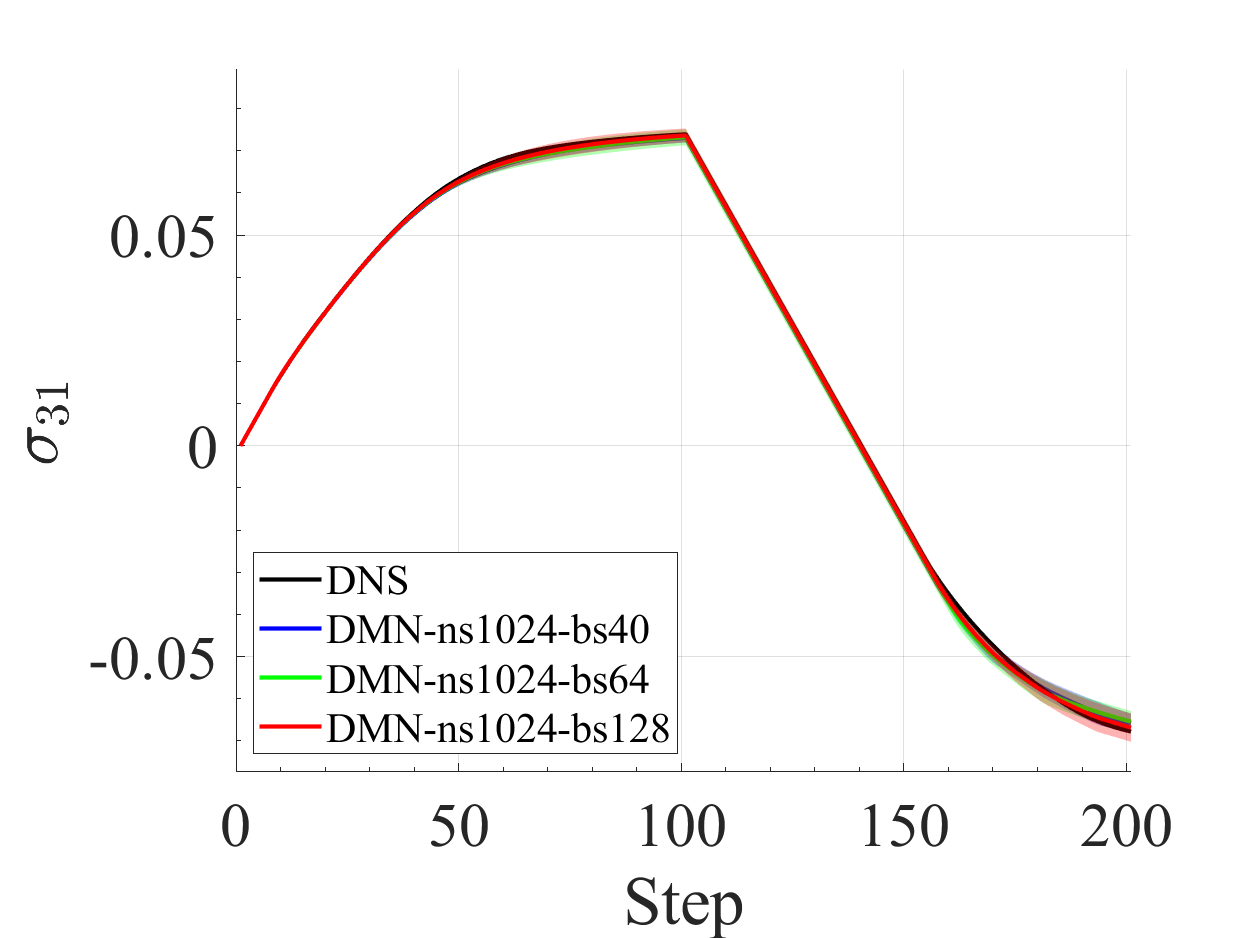}
        \caption{$\sigma_{31}$}
    \end{subfigure}
\caption{Composite 3 - Stress predictions from DMNs trained with 1024 samples using batch sizes of 40, 64, and 128: (a) $\sigma_{11}$; (b) $\sigma_{22}$; (c) $\sigma_{33}$; (d) $\sigma_{12}$; (e) $\sigma_{23}$; (f) $\sigma_{31}$; Solid lines show the mean over 10 runs with random initializations, and the shaded region indicate the corresponding standard deviations.}\label{fig.mat3_ns1024}
\end{figure}

This section presents predictions from DMNs trained on Dataset 2 (1024 samples) with batch sizes of 40, 64, and 128 for three composite materials, as shown in Figs. \ref{fig.mat1_ns1024} - \ref{fig.mat3_ns1024}.
Among the three configurations, a batch size of 128 yields more accurate predictions with lower uncertainty across all tested composite materials, especially in the 11, 22, and 12 directions, consistent with the corresponding training mean relative errors shown in Fig. \ref{fig.dmn_loss_data_size}(b).

\section{DMN Trained on Dataset 3}\label{sec:dmn_ns2048}
\begin{figure}[htp]
\centering
    \begin{subfigure}{0.325\textwidth}
        \centering
        \includegraphics[width=1\linewidth]{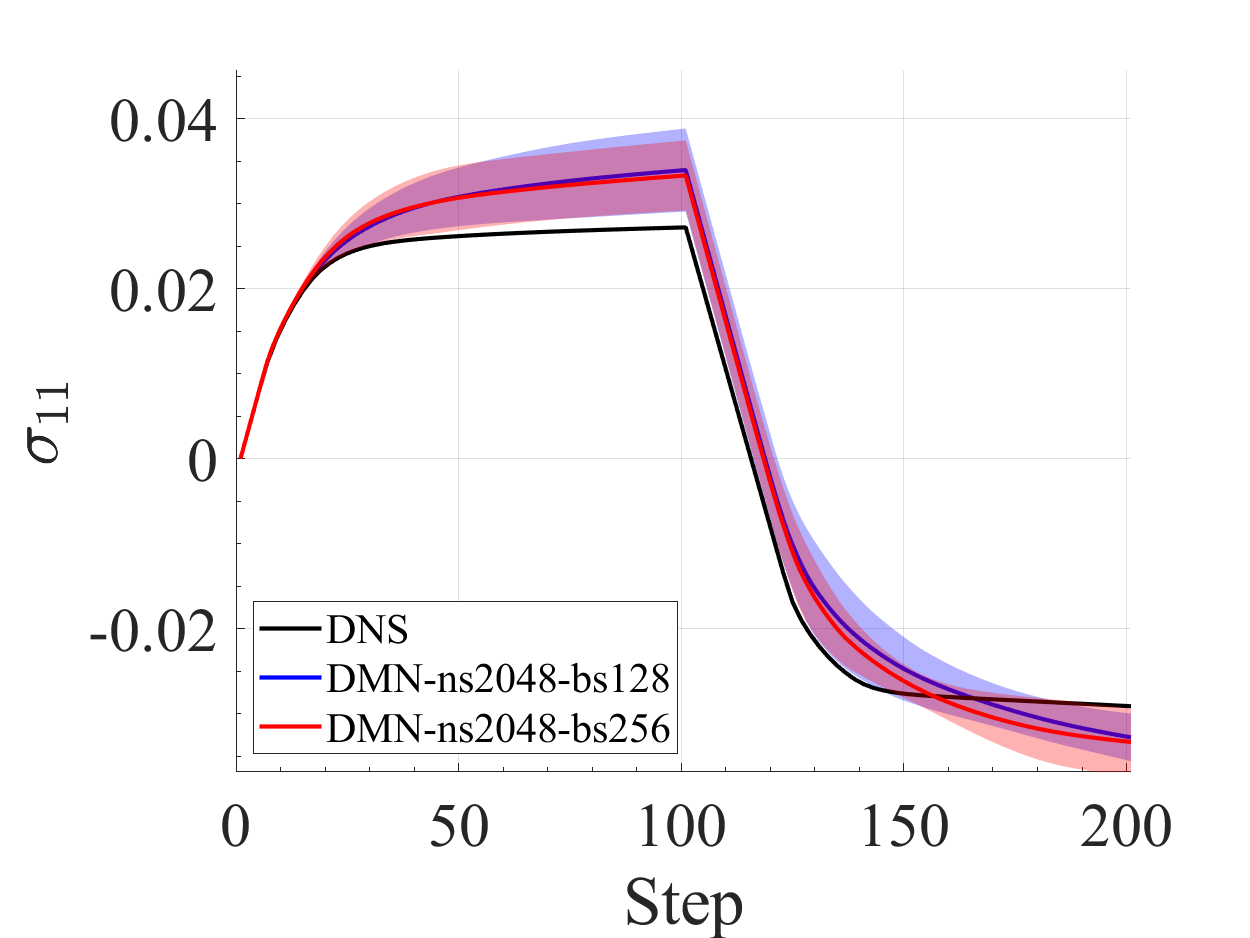}
        \caption{$\sigma_{11}$}
    \end{subfigure}
    \begin{subfigure}{0.325\textwidth}
        \centering
        \includegraphics[width=1\linewidth]{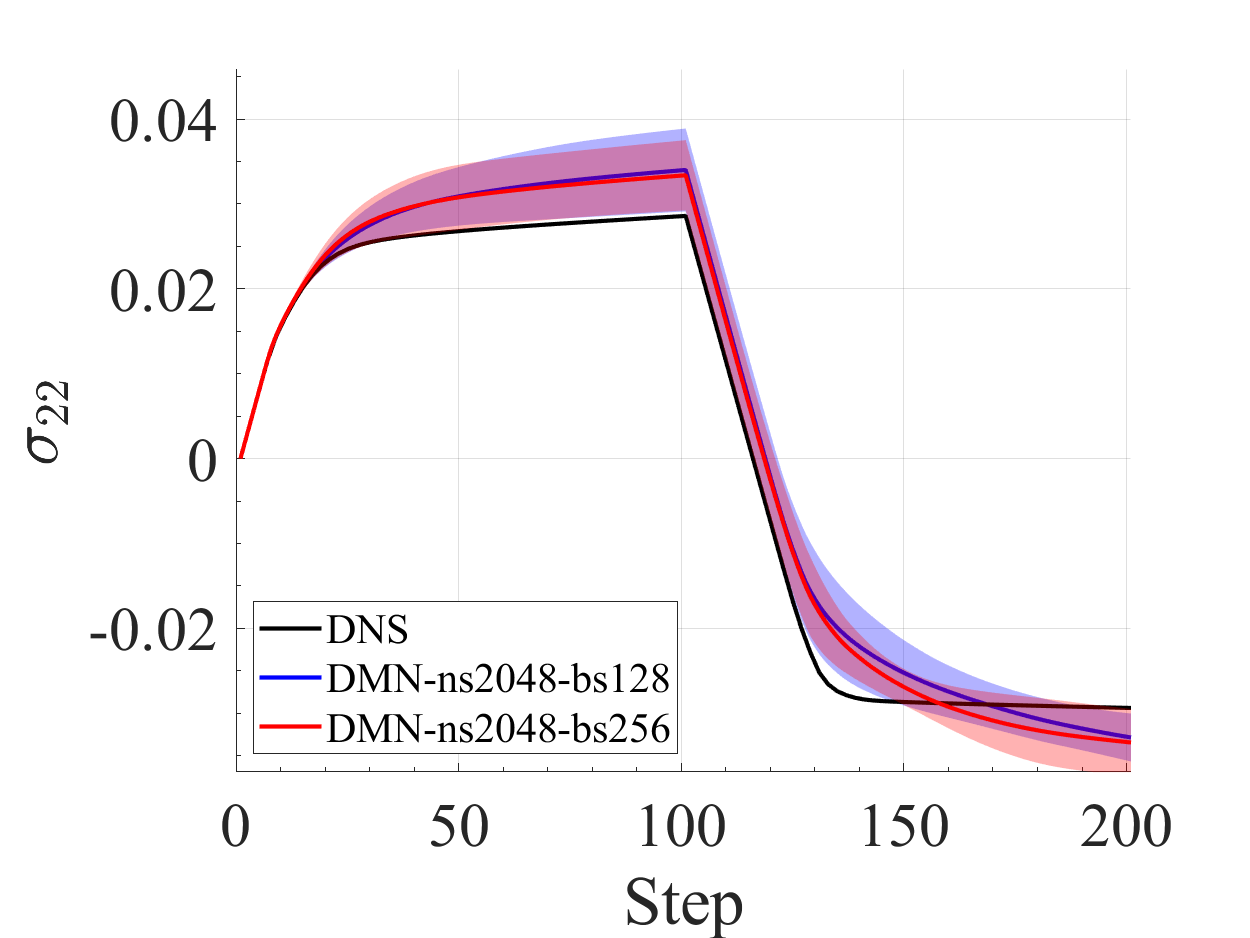}
        \caption{$\sigma_{22}$}
    \end{subfigure}
    \begin{subfigure}{0.325\textwidth}
        \centering
        \includegraphics[width=1\linewidth]{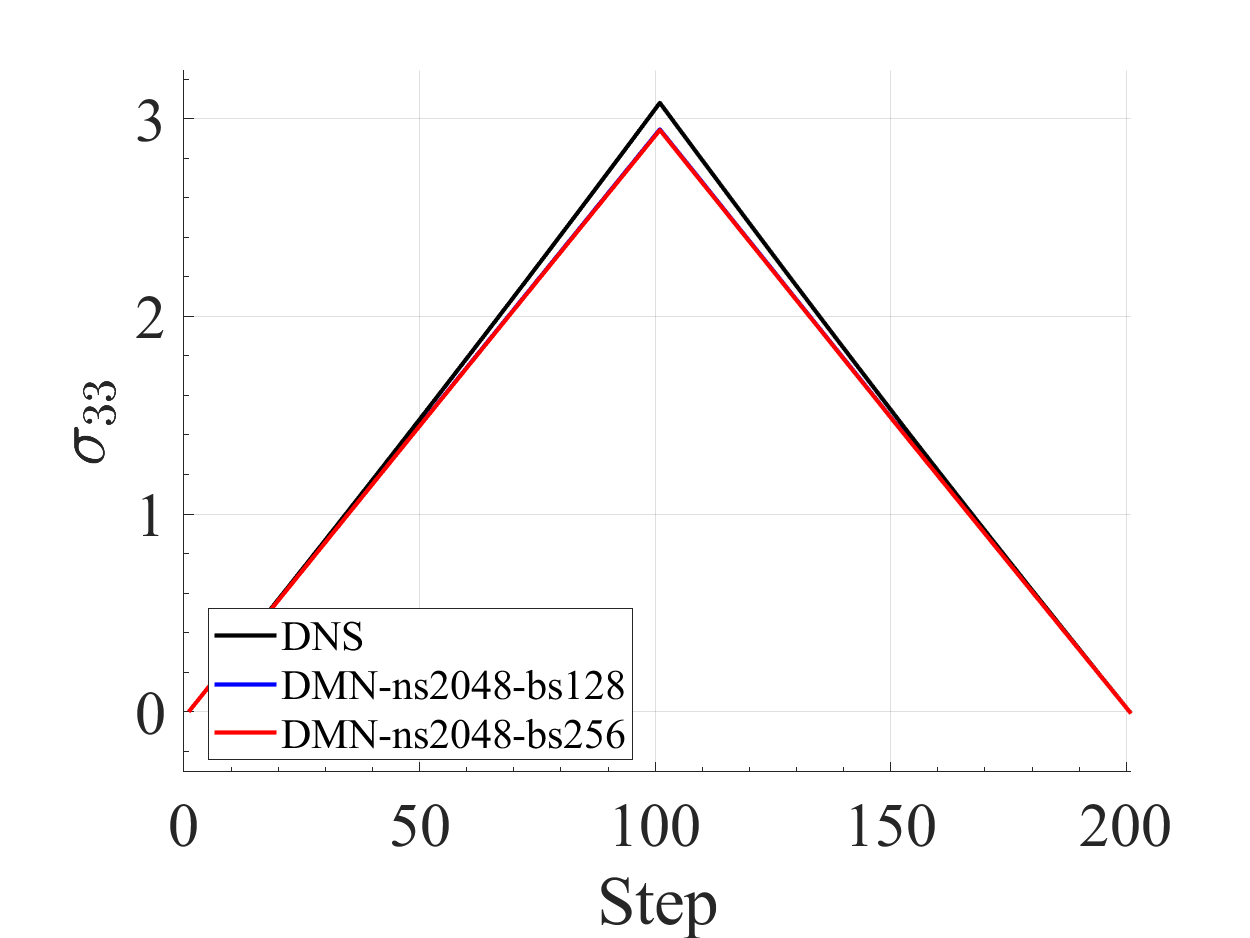}
        \caption{$\sigma_{33}$}
    \end{subfigure}
    \begin{subfigure}{0.325\textwidth}
        \centering
        \includegraphics[width=1\linewidth]{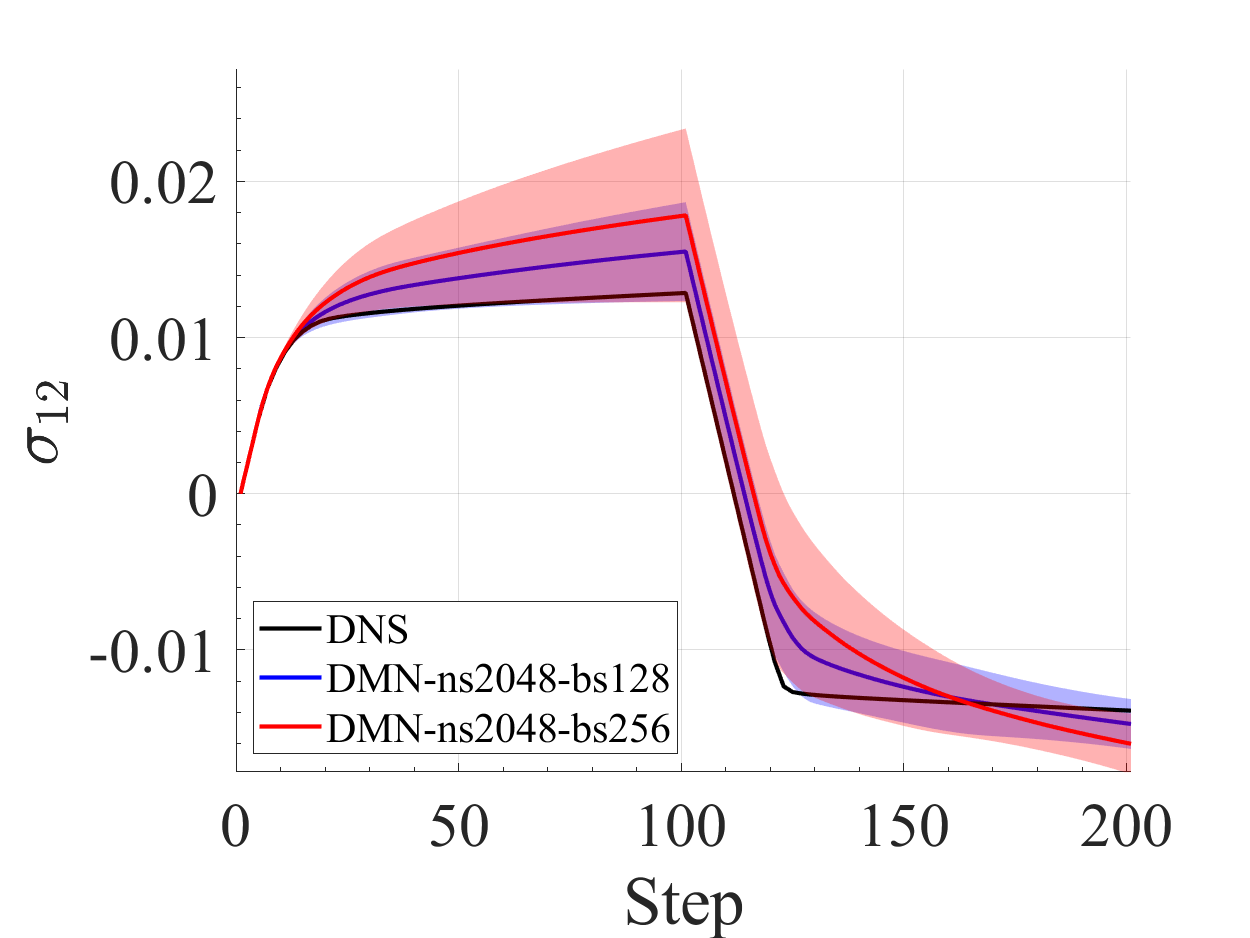}
        \caption{$\sigma_{12}$}
    \end{subfigure}
    \begin{subfigure}{0.325\textwidth}
        \centering
        \includegraphics[width=1\linewidth]{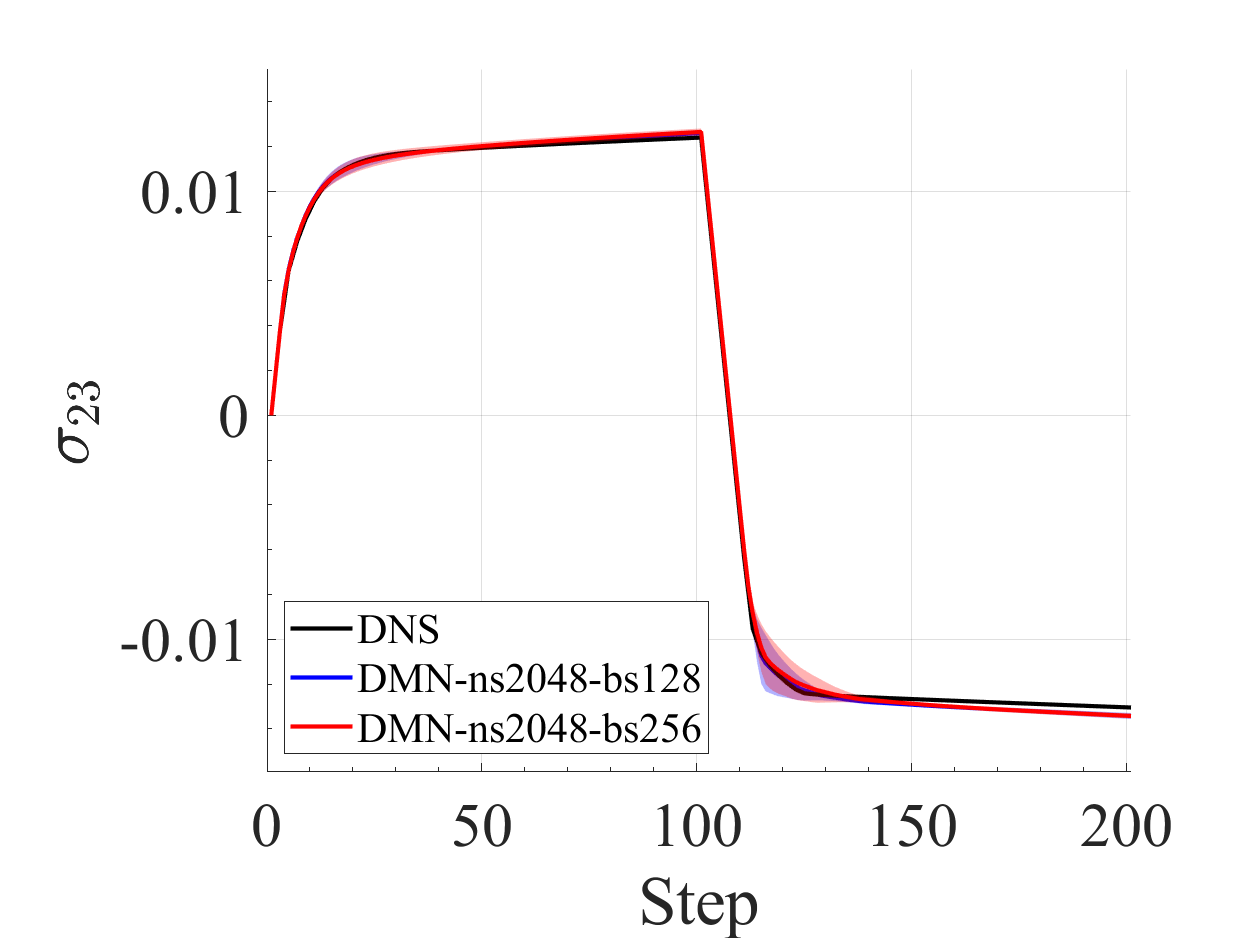}
        \caption{$\sigma_{23}$}
    \end{subfigure}
    \begin{subfigure}{0.325\textwidth}
        \centering
        \includegraphics[width=1\linewidth]{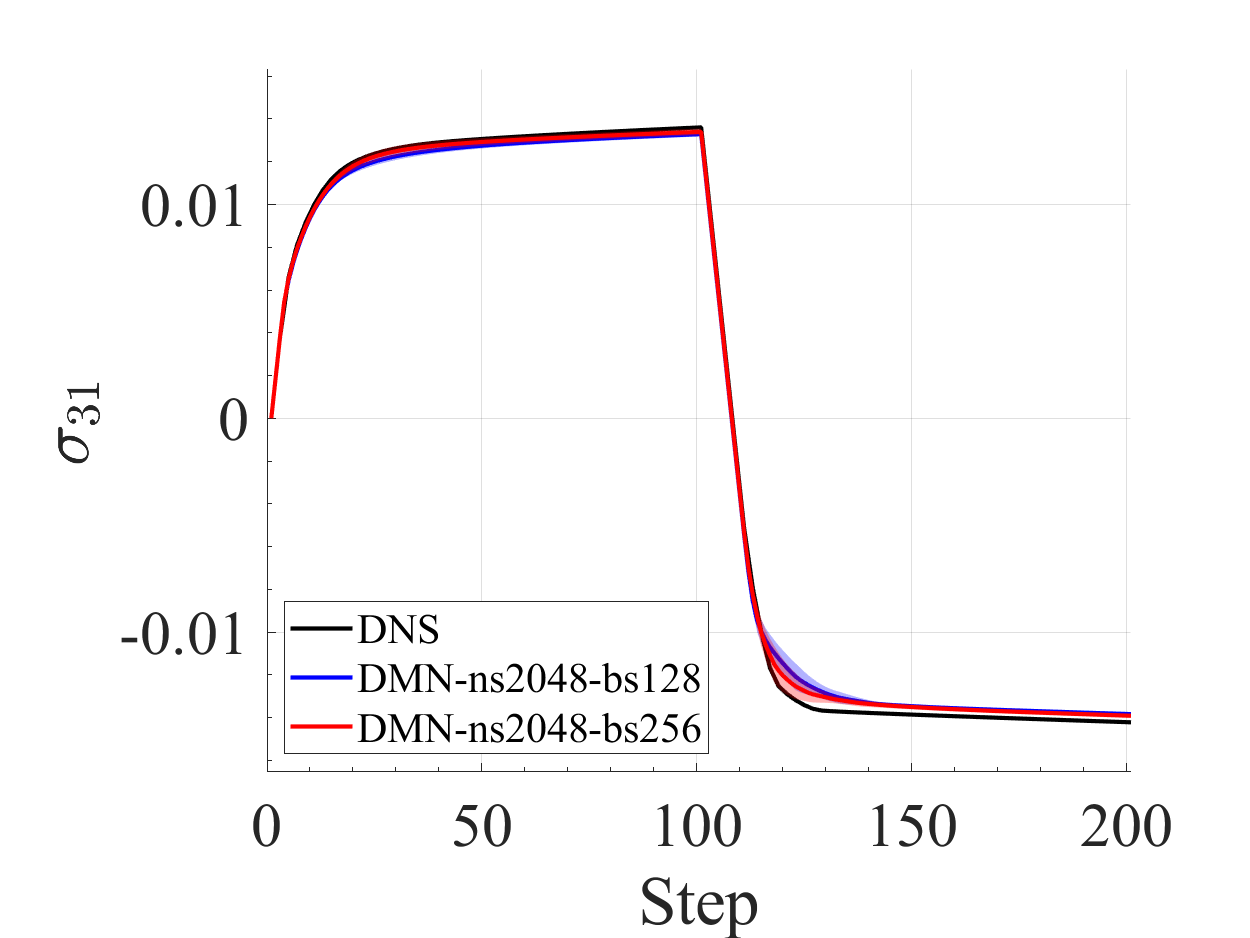}
        \caption{$\sigma_{31}$}
    \end{subfigure}
\caption{Composite 1 - Stress predictions from DMNs trained with 2048 samples using batch sizes of 128 and 256: (a) $\sigma_{11}$; (b) $\sigma_{22}$; (c) $\sigma_{33}$; (d) $\sigma_{12}$; (e) $\sigma_{23}$; (f) $\sigma_{31}$; Solid lines show the mean over 10 runs with random initializations, and the shaded region indicate the corresponding standard deviations.}\label{fig.mat1_ns2048}
\end{figure}

\begin{figure}[htp]
\centering
    \begin{subfigure}{0.325\textwidth}
        \centering
        \includegraphics[width=1\linewidth]{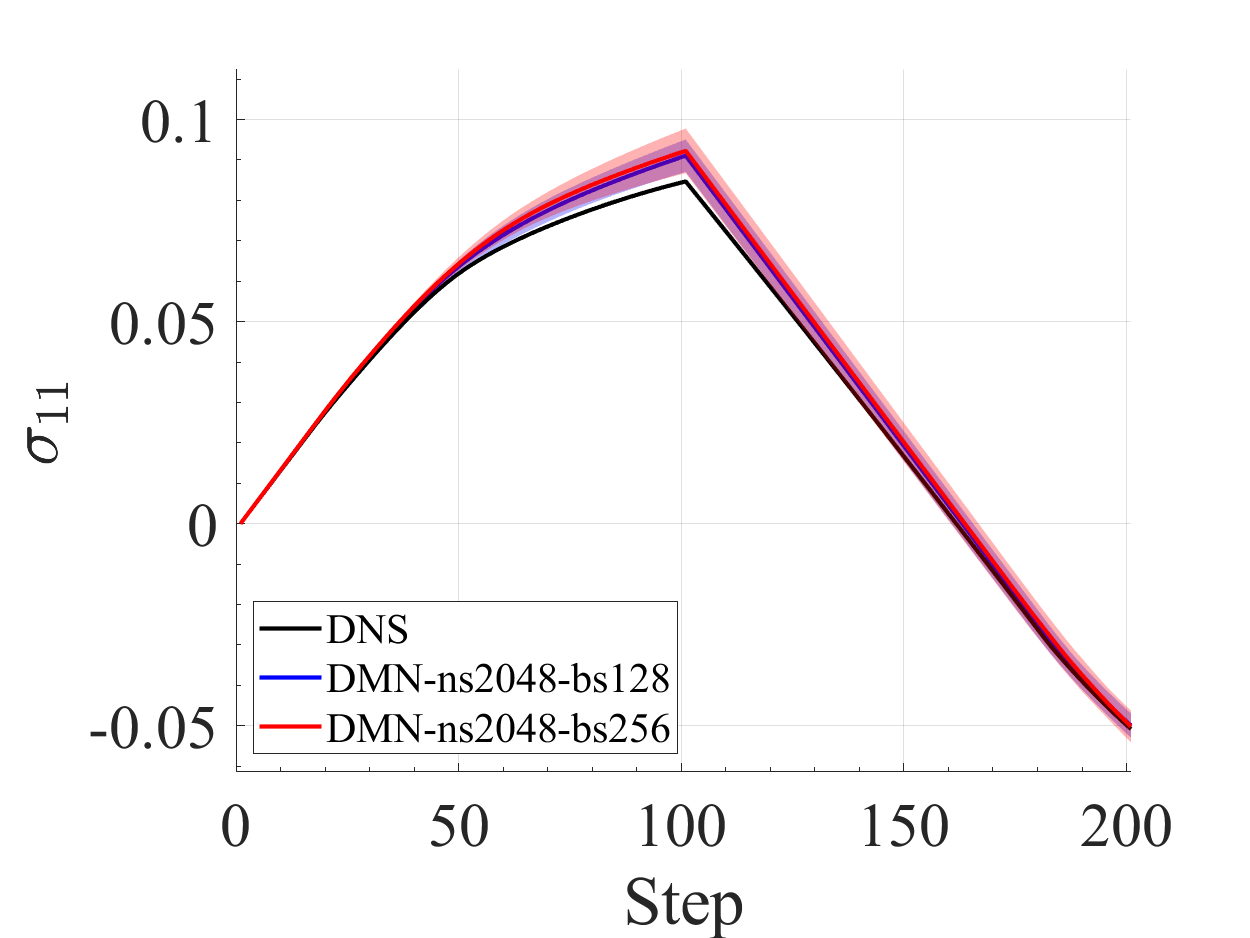}
        \caption{$\sigma_{11}$}
    \end{subfigure}
    \begin{subfigure}{0.325\textwidth}
        \centering
        \includegraphics[width=1\linewidth]{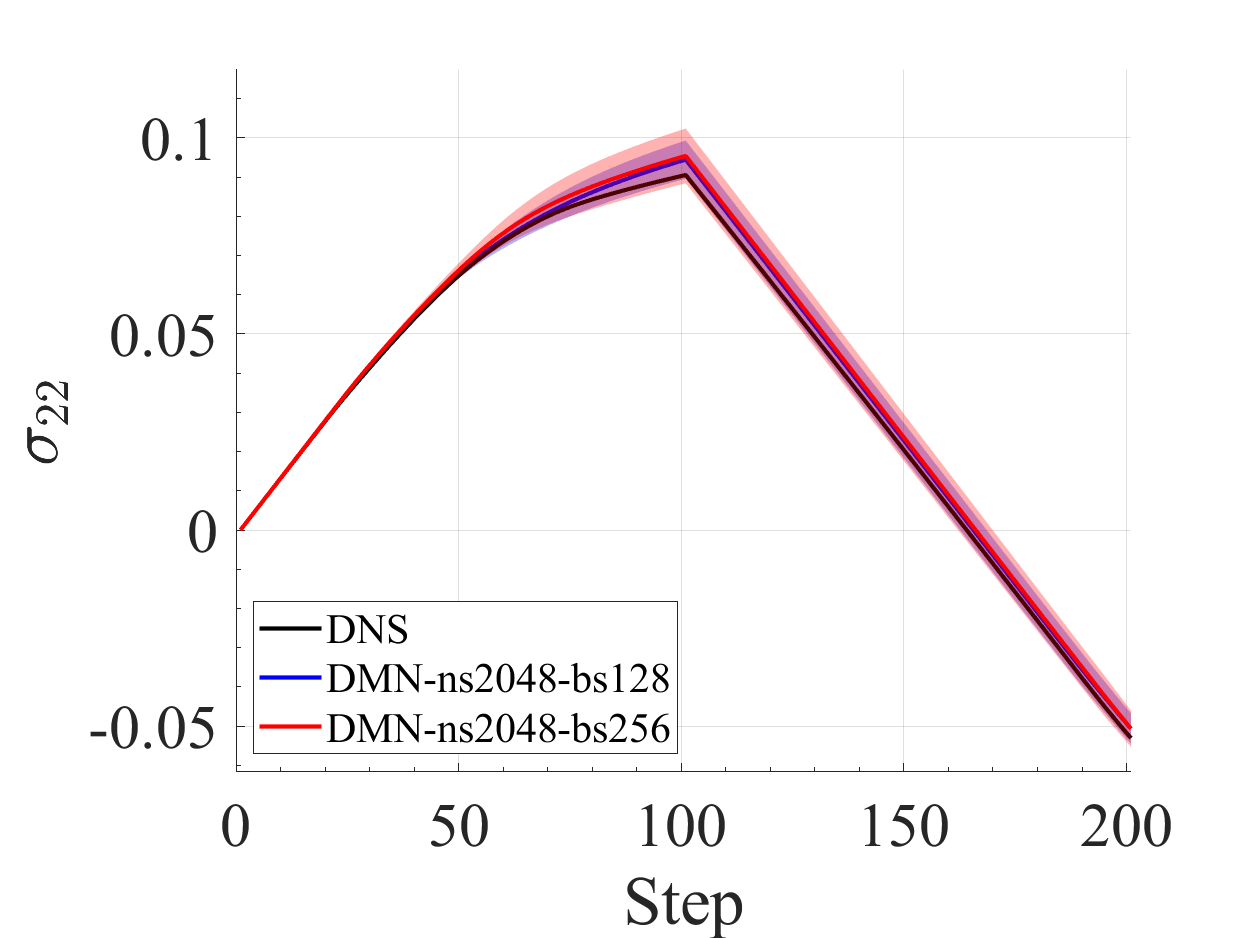}
        \caption{$\sigma_{22}$}
    \end{subfigure}
    \begin{subfigure}{0.325\textwidth}
        \centering
        \includegraphics[width=1\linewidth]{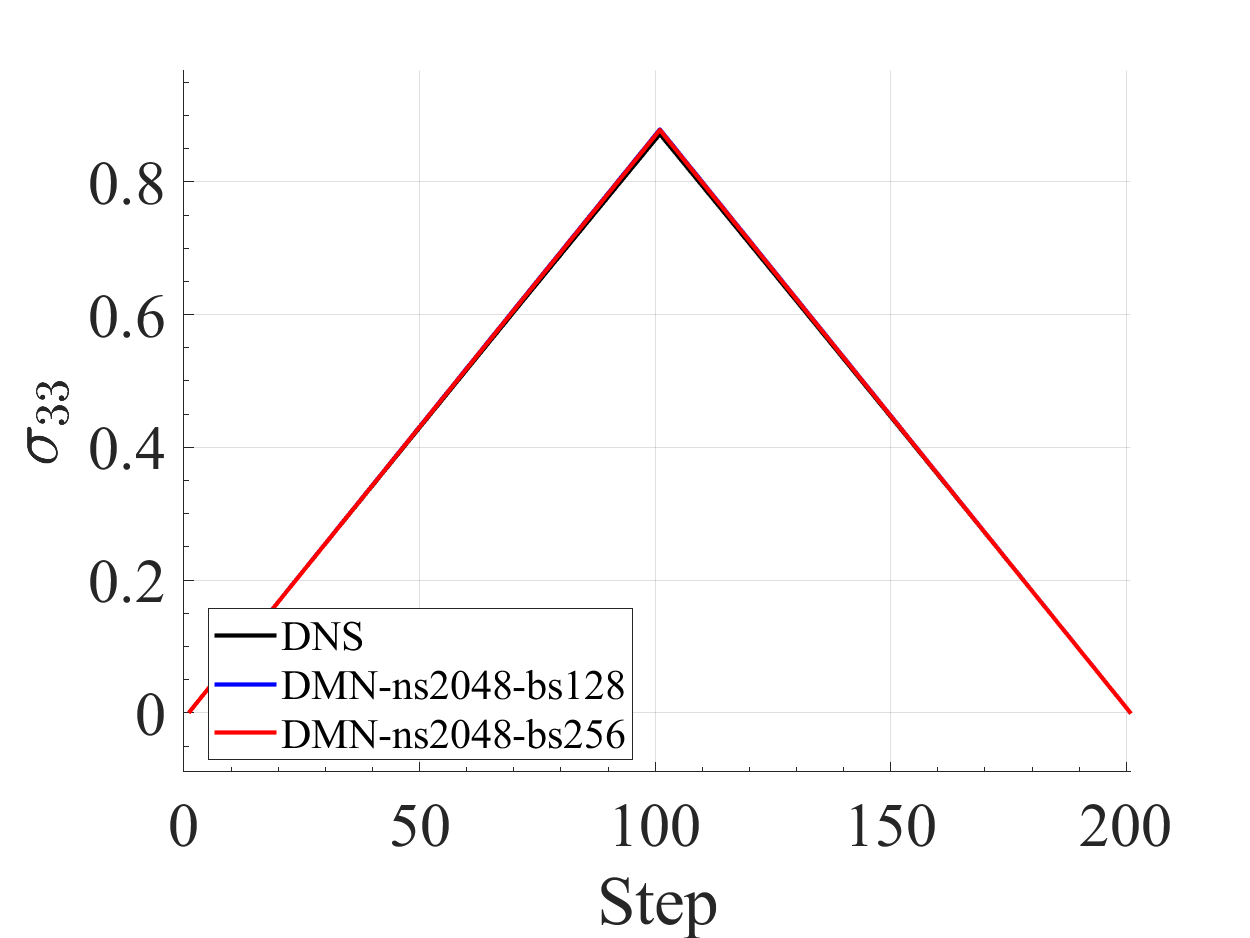}
        \caption{$\sigma_{33}$}
    \end{subfigure}
    \begin{subfigure}{0.325\textwidth}
        \centering
        \includegraphics[width=1\linewidth]{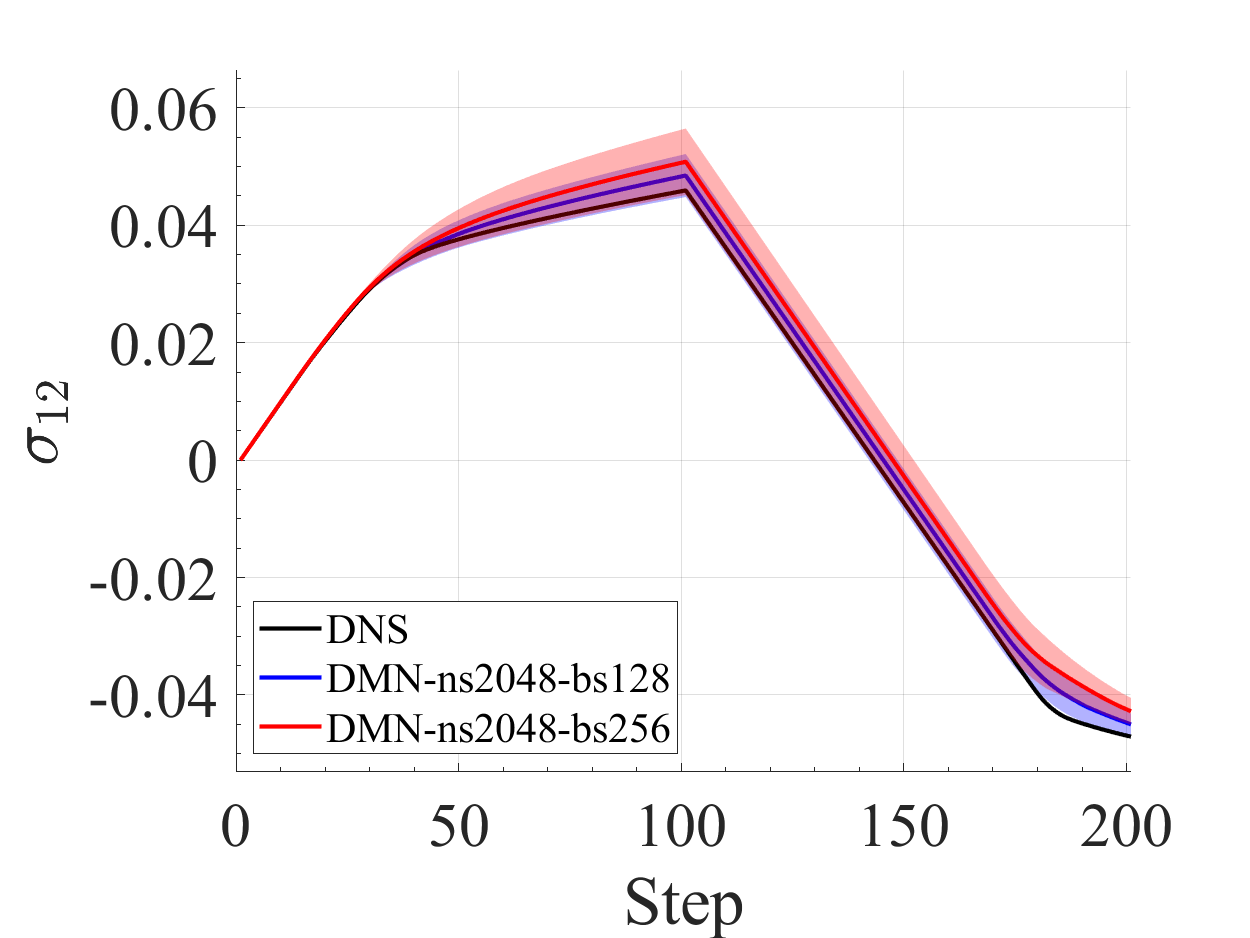}
        \caption{$\sigma_{12}$}
    \end{subfigure}
    \begin{subfigure}{0.325\textwidth}
        \centering
        \includegraphics[width=1\linewidth]{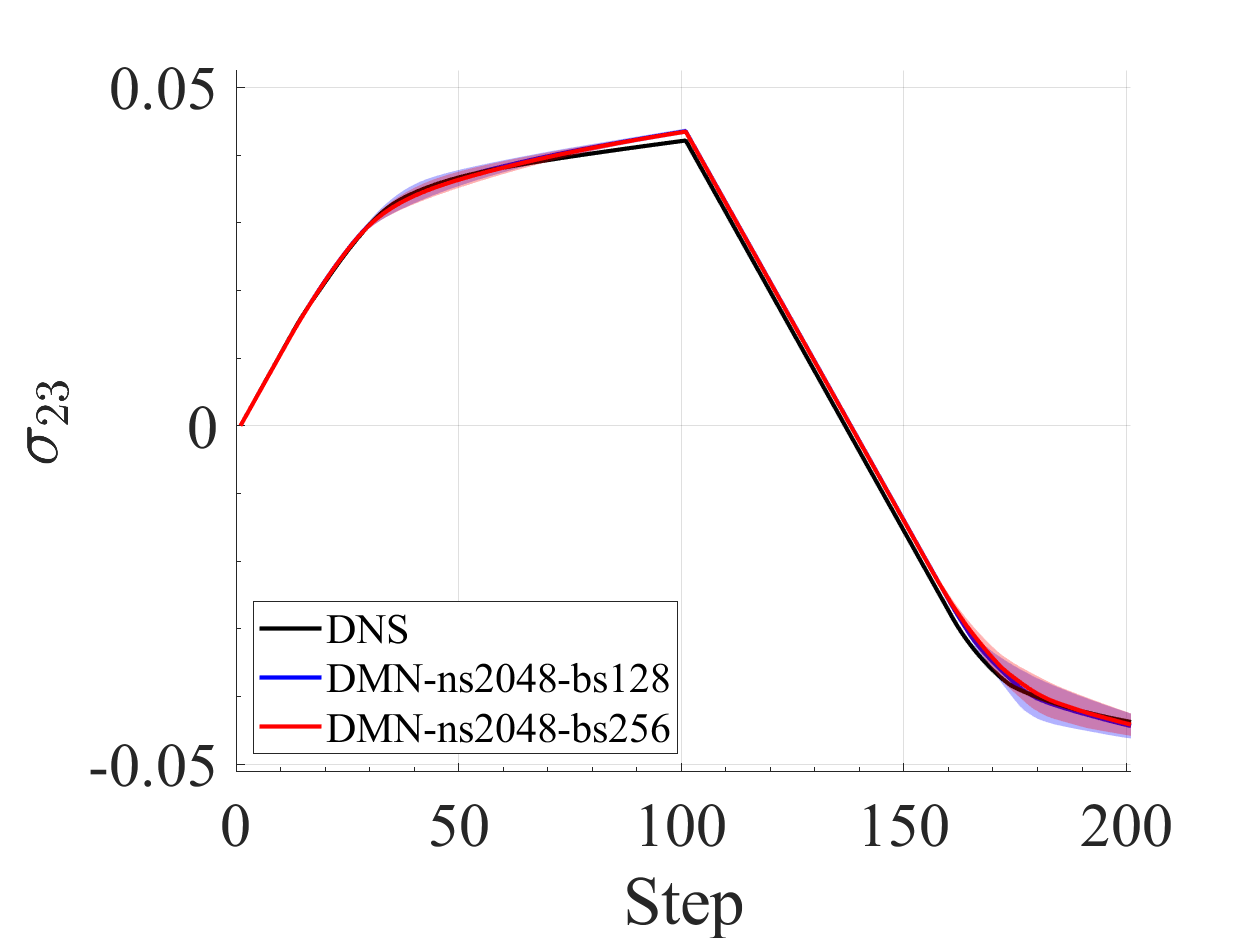}
        \caption{$\sigma_{23}$}
    \end{subfigure}
    \begin{subfigure}{0.325\textwidth}
        \centering
        \includegraphics[width=1\linewidth]{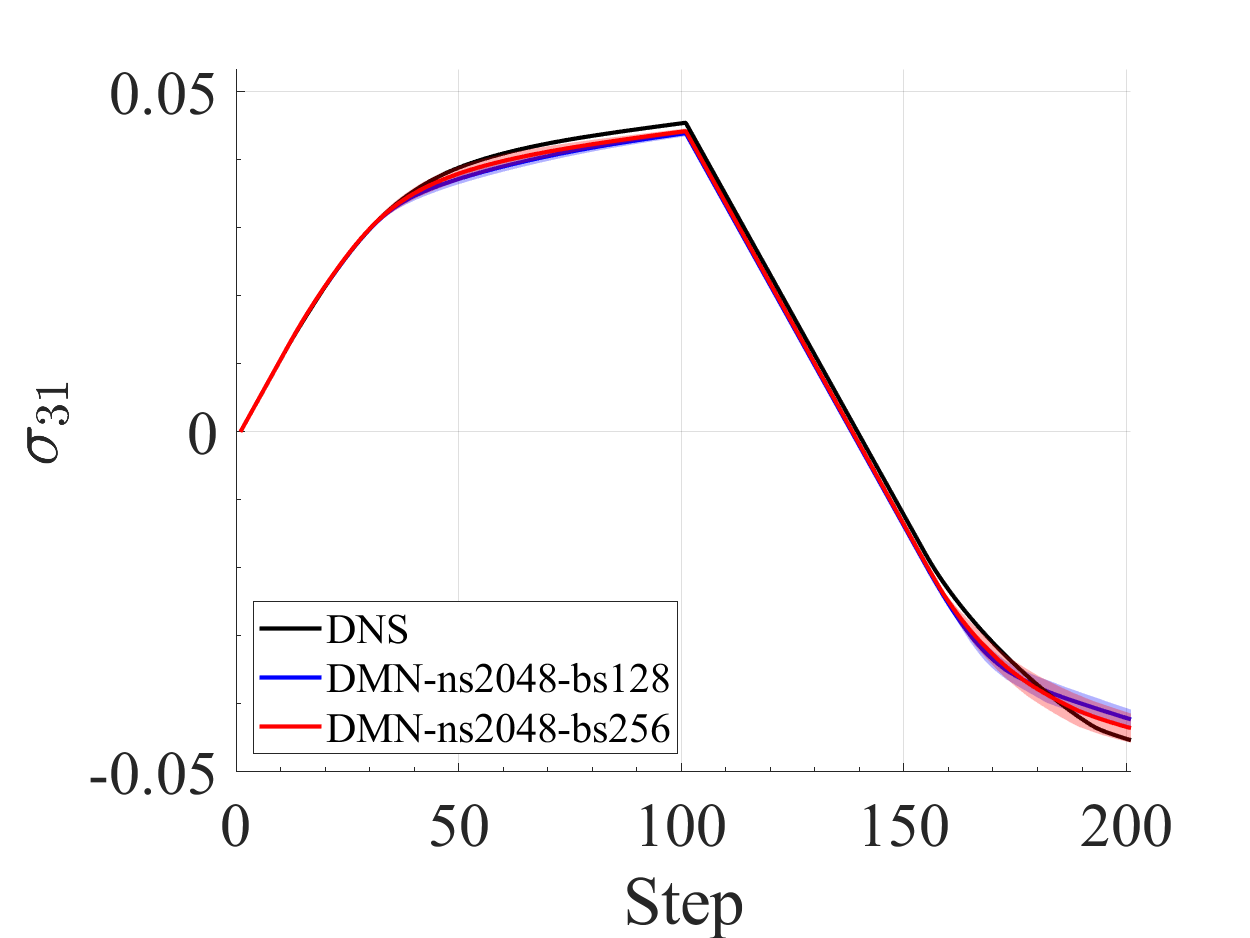}
        \caption{$\sigma_{31}$}
    \end{subfigure}
\caption{Composite 2 - Stress predictions from DMNs trained with 2048 samples using batch sizes of 128 and 256: (a) $\sigma_{11}$; (b) $\sigma_{22}$; (c) $\sigma_{33}$; (d) $\sigma_{12}$; (e) $\sigma_{23}$; (f) $\sigma_{31}$; Solid lines show the mean over 10 runs with random initializations, and the shaded region indicate the corresponding standard deviations.}\label{fig.mat2_ns2048}
\end{figure}

\begin{figure}[htp]
\centering
    \begin{subfigure}{0.325\textwidth}
        \centering
        \includegraphics[width=1\linewidth]{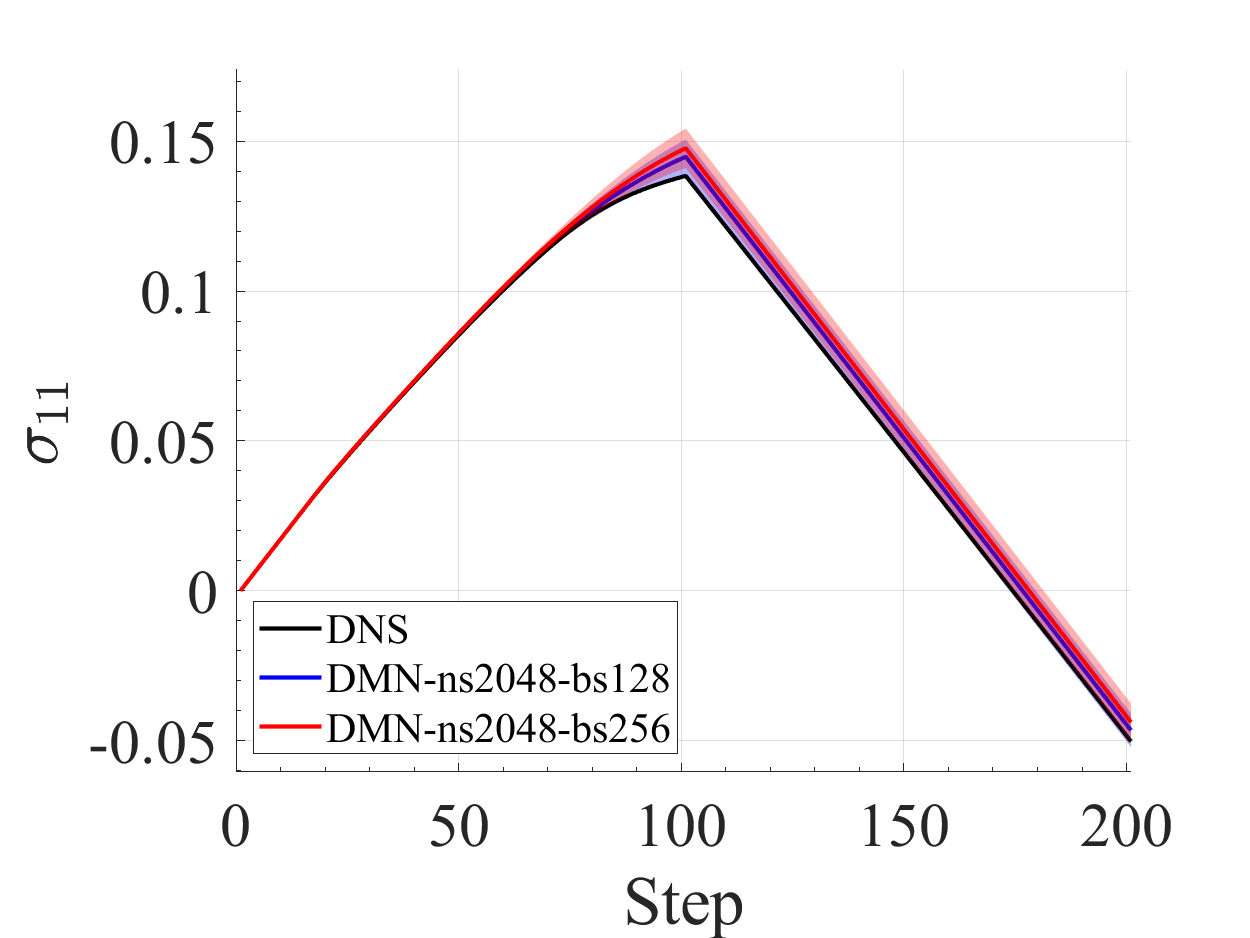}
        \caption{$\sigma_{11}$}
    \end{subfigure}
    \begin{subfigure}{0.325\textwidth}
        \centering
        \includegraphics[width=1\linewidth]{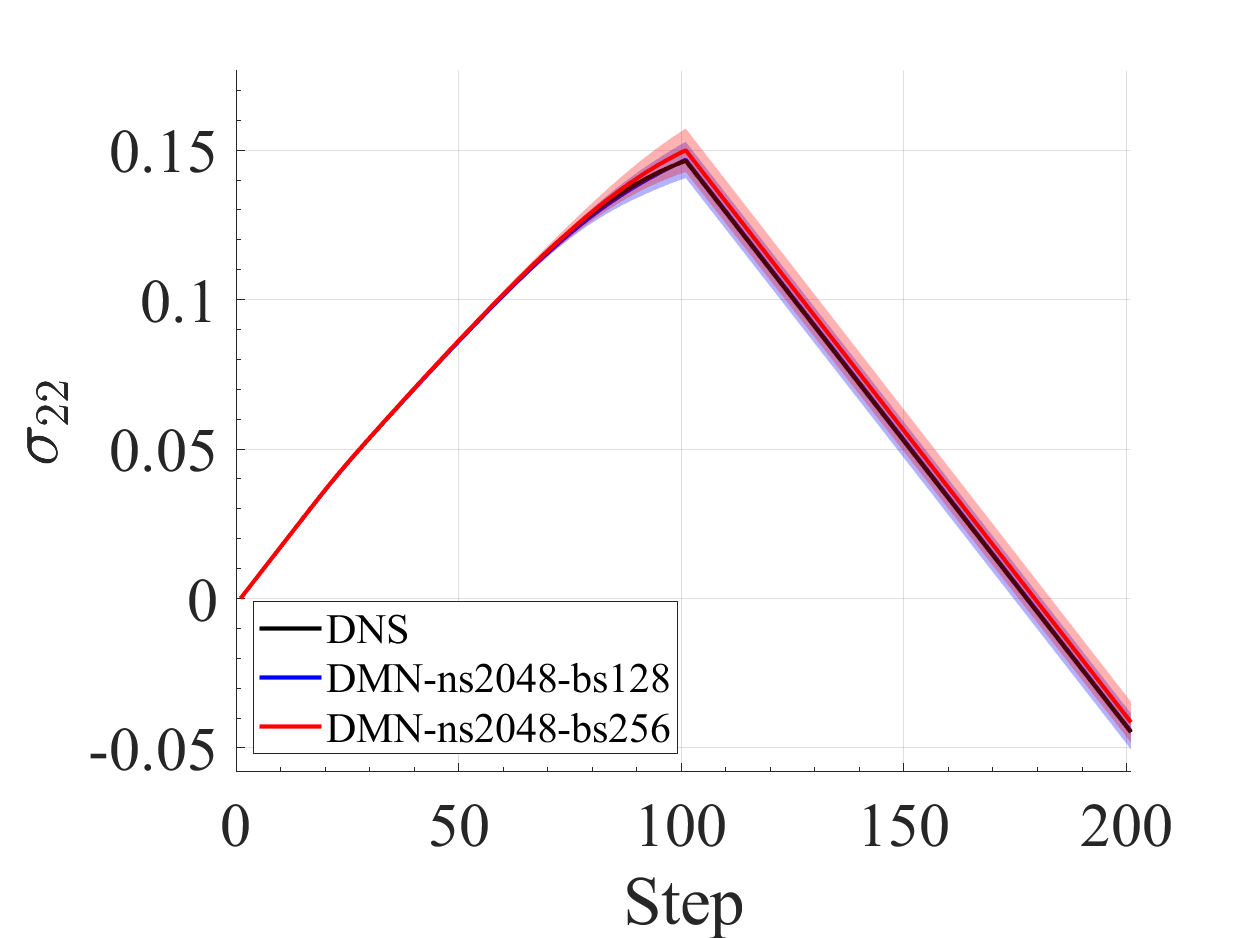}
        \caption{$\sigma_{22}$}
    \end{subfigure}
    \begin{subfigure}{0.325\textwidth}
        \centering
        \includegraphics[width=1\linewidth]{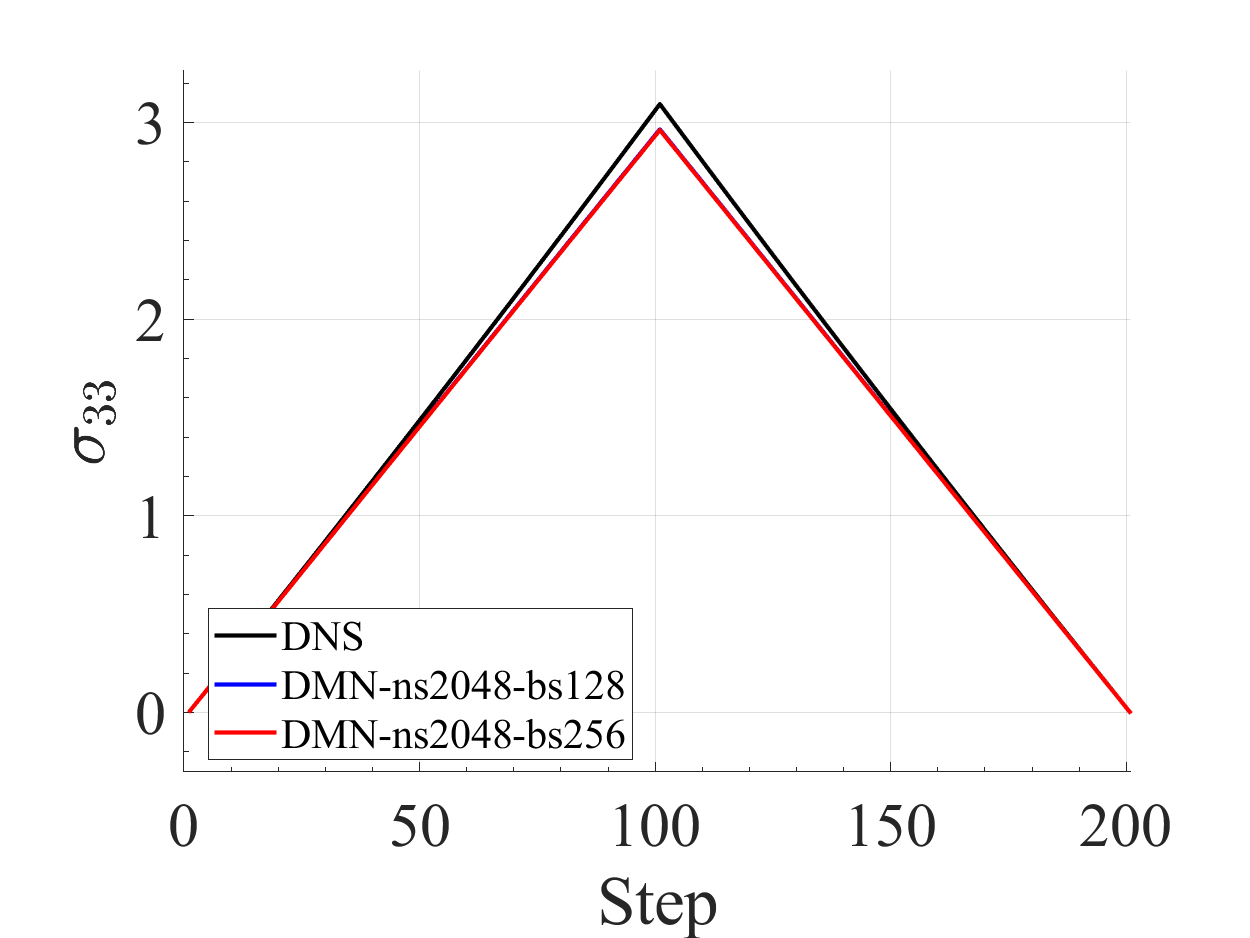}
        \caption{$\sigma_{33}$}
    \end{subfigure}
    \begin{subfigure}{0.325\textwidth}
        \centering
        \includegraphics[width=1\linewidth]{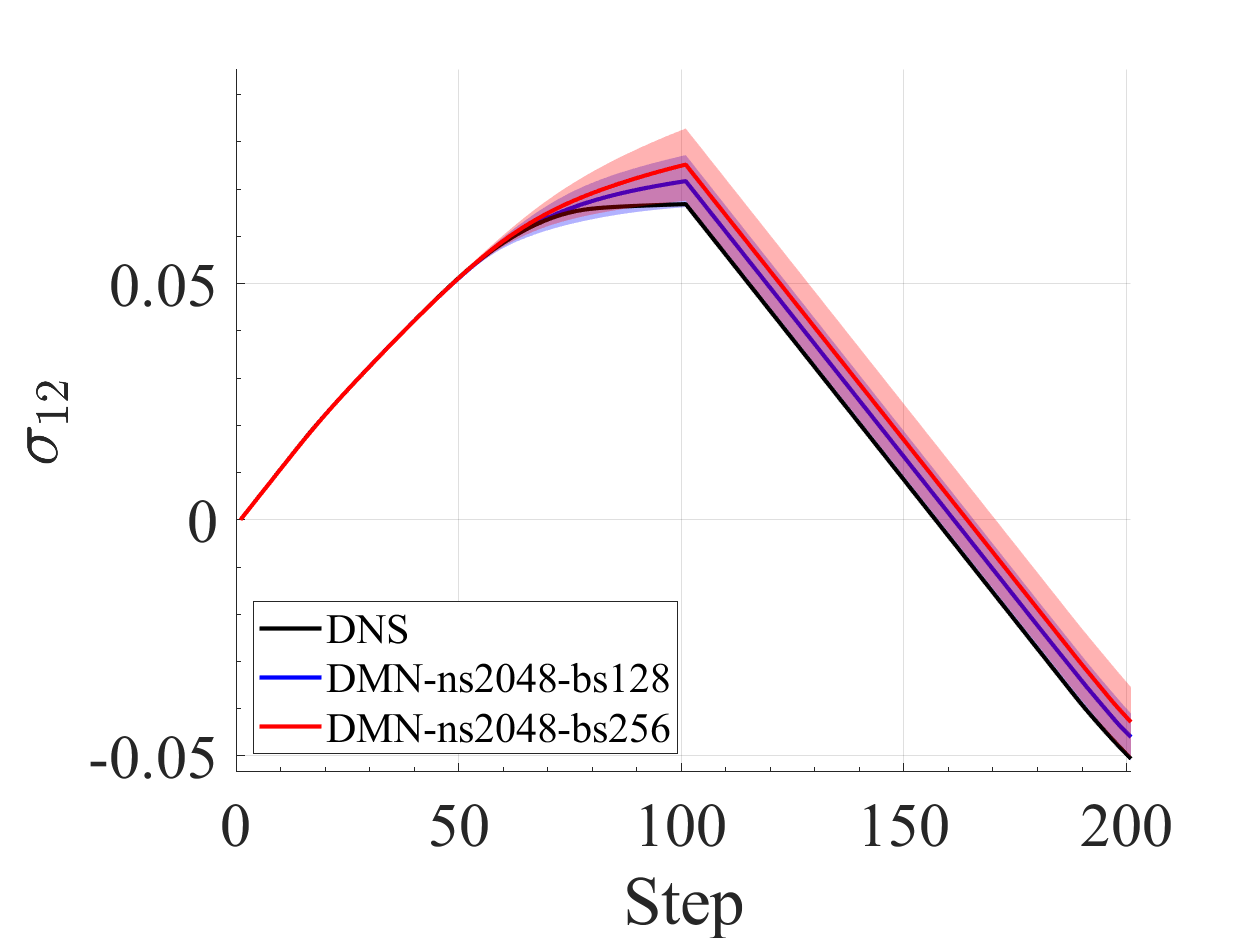}
        \caption{$\sigma_{12}$}
    \end{subfigure}
    \begin{subfigure}{0.325\textwidth}
        \centering
        \includegraphics[width=1\linewidth]{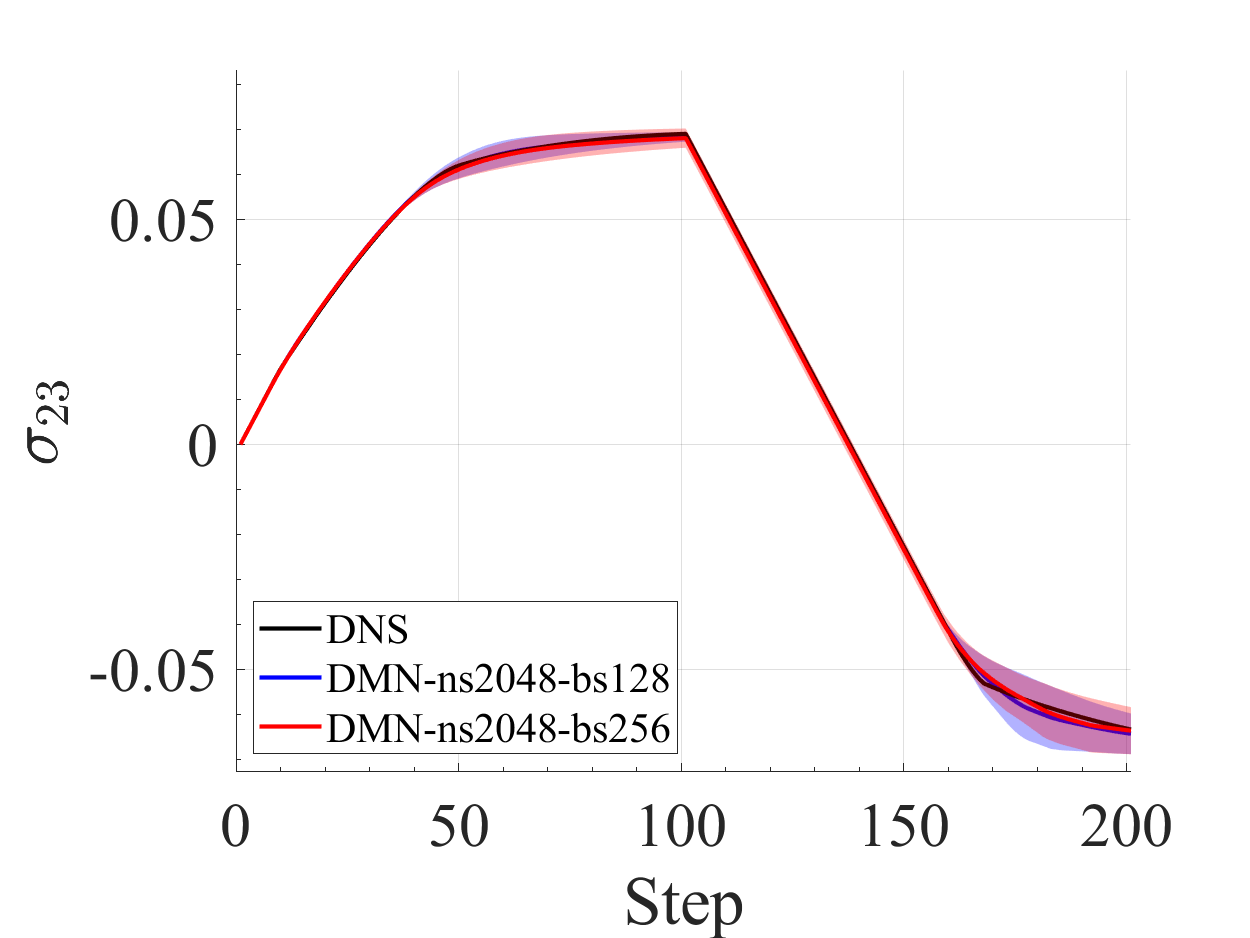}
        \caption{$\sigma_{23}$}
    \end{subfigure}
    \begin{subfigure}{0.325\textwidth}
        \centering
        \includegraphics[width=1\linewidth]{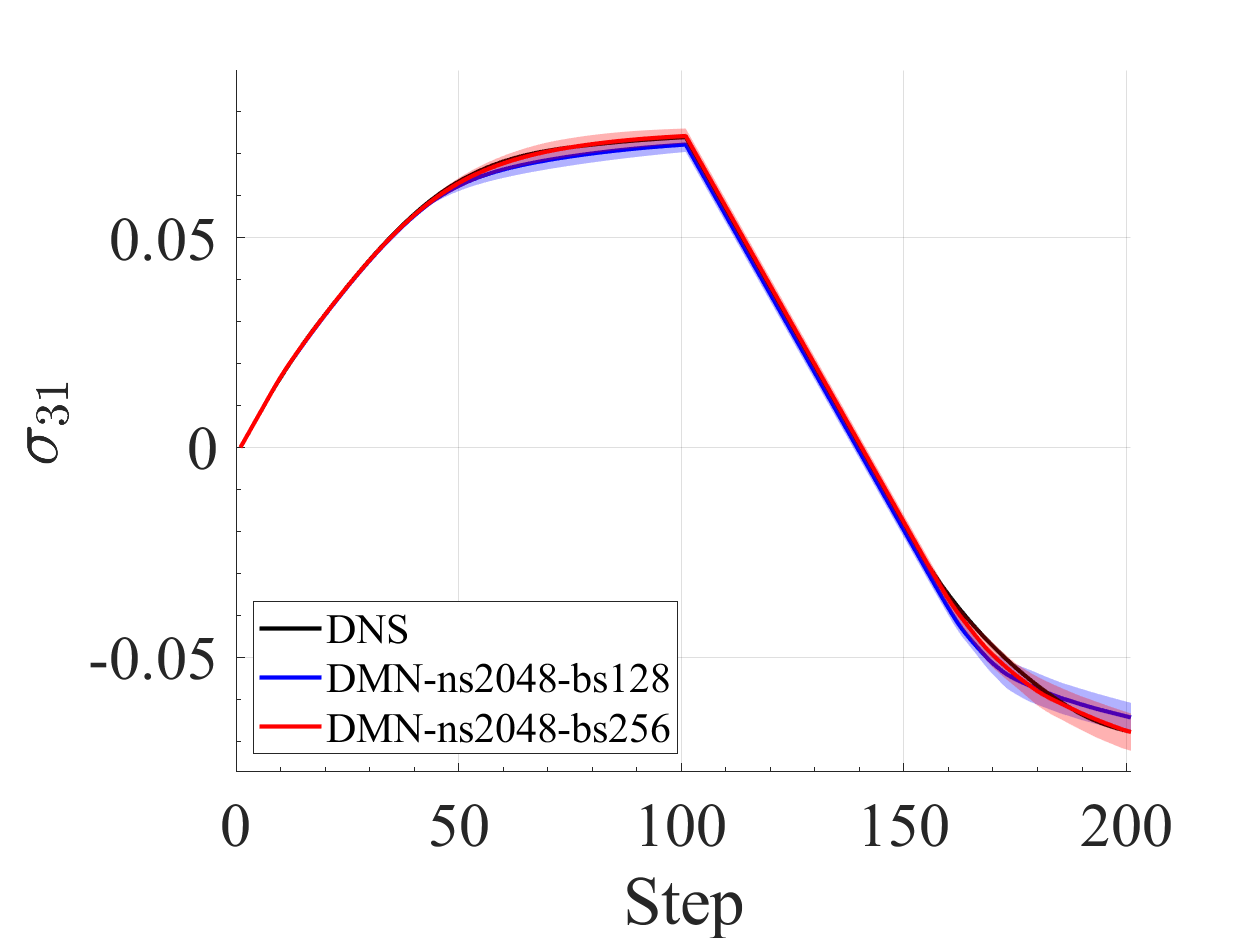}
        \caption{$\sigma_{31}$}
    \end{subfigure}
\caption{Composite 3 - Stress predictions from DMNs trained with 2048 samples using batch sizes of 128 and 256: (a) $\sigma_{11}$; (b) $\sigma_{22}$; (c) $\sigma_{33}$; (d) $\sigma_{12}$; (e) $\sigma_{23}$; (f) $\sigma_{31}$; Solid lines show the mean over 10 runs with random initializations, and the shaded region indicate the corresponding standard deviations.}\label{fig.mat3_ns2048}
\end{figure}

This section presents predictions from DMNs trained on Dataset 3 (2048 samples) with batch sizes of 128 and 256 for three composite materials, as shown in Figs. \ref{fig.mat1_ns2048} - \ref{fig.mat3_ns2048}.
Among the these configurations, a batch size of 128 yields more accurate predictions with lower uncertainty across all tested composite materials, especially in the 11, 22, and 12 directions, consistent with the corresponding training mean relative errors shown in Fig. \ref{fig.dmn_loss_data_size}(c).

\end{document}